\documentclass{article}

\PassOptionsToPackage{numbers, compress}{natbib}

\usepackage[final]{cpal_2025}

\usepackage{amsmath,amsfonts,bm}

\def\eqref#1{equation~\ref{#1}}

\def\ceil#1{\lceil #1 \rceil}
\def\floor#1{\lfloor #1 \rfloor}
\def\1{\bm{1}}

\def\va{{\bm{a}}}
\def\vb{{\bm{b}}}

\def\vh{{\bm{h}}}

\def\vx{{\bm{x}}}

\def\vz{{\bm{z}}}

\DeclareMathAlphabet{\mathsfit}{\encodingdefault}{\sfdefault}{m}{sl}
\SetMathAlphabet{\mathsfit}{bold}{\encodingdefault}{\sfdefault}{bx}{n}

\newcommand{\E}{\mathbb{E}}

\newcommand{\R}{\mathbb{R}}

\newcommand{\softmax}{\mathrm{softmax}}

\usepackage{url}
\usepackage{inconsolata}
\usepackage{highlight}
\usepackage{xcolor}
\usepackage{colortbl}
\usepackage{pgf}
\usepackage{textcomp}
\usepackage{graphicx}
\usepackage{subcaption}
\usepackage{mdframed}
\usepackage{booktabs}
\usepackage{subcaption}
\usepackage{multirow}
\usepackage{wrapfig}
\usepackage{adjustbox}
\graphicspath{{./figures/}}
\usepackage[breakable, theorems, skins]{tcolorbox}
\usepackage{seqsplit}
\usepackage{tikz}

\DeclareRobustCommand{\StartCrate}{%
  \begingroup\normalfont
  \includegraphics[height=\fontcharht\font`\B*4/3]{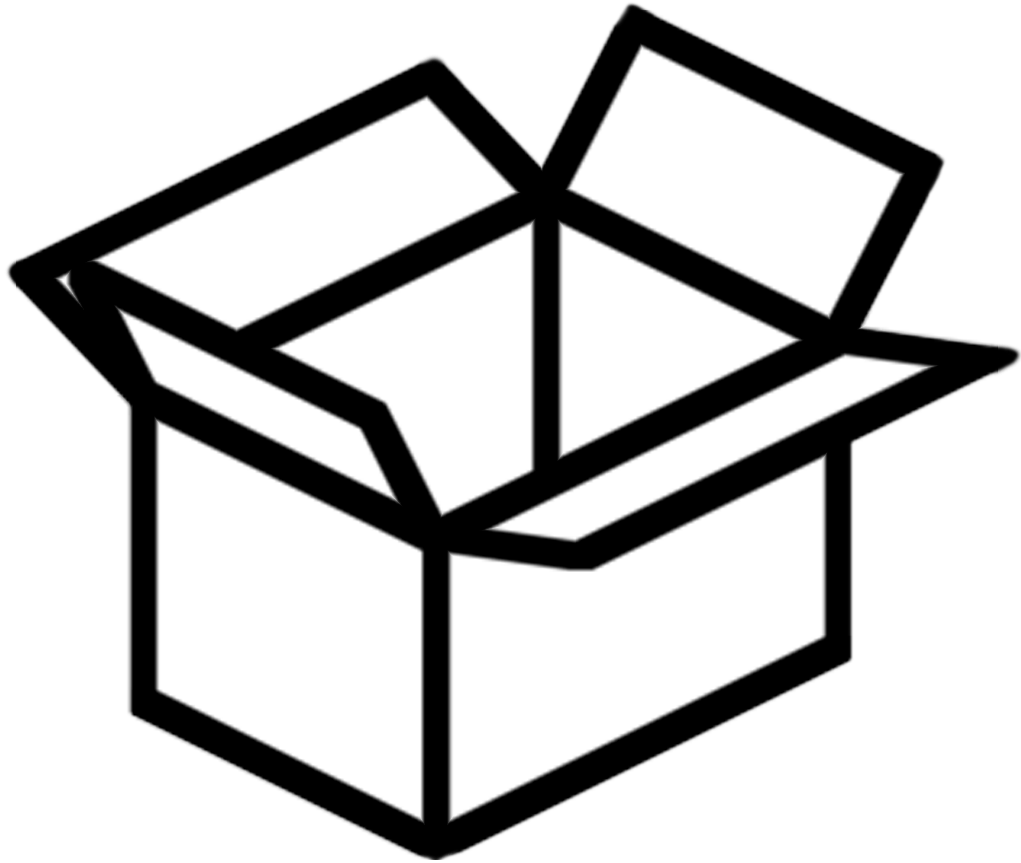}%
  \endgroup
}

\usepackage{algorithmicx}
\usepackage{algpseudocode}

\usepackage{algorithm}

\usepackage{listings}

\definecolor{commentcolor}{RGB}{110,154,155}   
\lstdefinestyle{mystyle}{
    backgroundcolor=\color{white},   
    commentstyle=\color{commentcolor},
    keywordstyle=\color{purple},
    numberstyle=\tiny\color{gray},
    stringstyle=\color{blue},
    basicstyle=\ttfamily\footnotesize,
    breakatwhitespace=false,         
    breaklines=true,
    columns=fullflexible,
    escapeinside=``,
    captionpos=b,                    
    keepspaces=true,                 
    numbers=left,                    
    numbersep=5pt,                  
    showspaces=false,                
    showstringspaces=false,
    showtabs=false,                  
    tabsize=4
}
\NewDocumentCommand{\mat}{m}{\begin{bmatrix}#1\end{bmatrix}}

\newcommand{\vZ}{\mathbb{Z}}

\newcommand{\bR}{\mathbb{R}}

\newcommand{\vX}{\bm{X}}
\newcommand{\X}{\vX}

\newcommand{\D}{\bm{D}}

\renewcommand{\softmax}[1]{\operatorname{softmax}\left(#1\right)}

\newcommand{\head}{\mathrm{head}}
\newcommand{\pre}{\mathrm{pre}}

\newcommand{\pos}{\mathrm{pos}}

\newcommand{\ISTA}[1]{\operatorname{\texttt{ISTA}}(#1)}

\newcommand{\MSSA}[1]{\operatorname{\texttt{MSSA}}(#1)}
\newcommand{\ours}{\textsc{crate}}
\newcommand{\ourscaps}{CRATE}
\newcommand{\ct}[2]{\colorbox[HTML]{#1}{\strut\texttt{#2}}}

\newcommand\blfootnote[1]{%
	\begingroup
	\renewcommand\thefootnote{}\footnote{#1}%
	\addtocounter{footnote}{-1}%
	\endgroup
}

\usepackage{hyperref}
\usepackage{cleveref}

\definecolor{citecolor}{HTML}{0071bc}
\hypersetup{
    colorlinks=true,%
    citecolor=citecolor,%
    filecolor=citecolor,%
    linkcolor=citecolor,%
    urlcolor=citecolor
}

\input{macros.def}

\title{Improving Neuron-level Interpretability with White-box Language Models}

\author{%
  \textbf{Hao Bai}\textsuperscript{1,2} 
  ~\textbf{Yi Ma}\textsuperscript{1,3}
  \\
  \vspace{0.05in}
  \vspace{0.1in}
  \textsuperscript{1}UC Berkeley\quad 
  \textsuperscript{2}UIUC\quad 
  \textsuperscript{3}HKU\\
  \vspace{0.1in}
  \url{https://crate-lm.github.io/}
  \\
}

\newcommand{\jack}[1]{} 
\newcommand{\jackadv}[1]{}

\newcommand{\yaodong}[1]{}

\begin{document}

\maketitle


\begin{abstract}

\jack{The abstract should (related work) briefly introduce the recent advances in neuron activation interpretation, including whether \& how activations can be interpreted, and how to systematically evaluate the interpretability. Then it should introduce (motivation) the white-box architecture \ours{} and why we think it can improve the interpretaion. At last, it should highlight all major contributions and findings of this work, including a brief comparison between \ours{} and the post-hoc sparse-autoencoder method.}

Neurons in auto-regressive language models like GPT-2 can be interpreted by analyzing their activation patterns. Recent studies have shown that techniques such as dictionary learning, a form of post-hoc sparse coding, enhance this neuron-level interpretability.
In our research, we are driven by the goal to fundamentally improve neural network interpretability by embedding sparse coding directly within the model architecture, rather than applying it as an afterthought.
In our study, we introduce a white-box transformer-like architecture named Coding RAte TransformEr (\ours{}), explicitly engineered to capture sparse, low-dimensional structures within data distributions. 
Our comprehensive experiments showcase significant improvements (up to $103$\% relative improvement) in neuron-level interpretability across a variety of evaluation metrics.
Detailed investigations confirm that this enhanced interpretability is steady across different layers irrespective of the model size, underlining \ours{}'s robust performance in enhancing neural network interpretability.
Further analysis shows that \ours{}'s increased interpretability comes from its enhanced ability to consistently and distinctively activate on relevant tokens. These findings point towards a promising direction for creating white-box foundation models that excel in neuron-level interpretation.

\end{abstract}

\blfootnote{\hspace*{-1.5mm}This work was entirely done at UC Berkeley.
}

\vspace{-2mm}
\section{Introduction}
\vspace{-2mm}

\jack{This paragraph gives readers a bottom-top idea about what area we're working in. It should include useful keywords for the area chair for picking reviewers wisely, including representation learning, neuron interpretation, mechanistic interpretation, reversed-engineering language models, etc.}

\textit{Representation learning} aims to learn a continuous mapping, to transform a random vector in a high dimensional space that is sampled from a dataset, to a feature vector in another (typically lower-dimensional) space~\citep{bengio2013representation}. Recently, deep learning has witnessed tremendous empirical success in modeling massive amounts of high-dimensional data, and the predominant practice has been to learn first a task-agnostic representation by pre-training a large neural network, which is commonly known as the \textit{foundation model}~\citep{devlin2018bert, radford2019language}. Among language foundation models, the transformers architecture~\citep{vaswani2017attention} with Generative Pre-Training~\citep{radford2019language} (GPT) has recently demonstrated a strong capability of modeling sequential data and thus predicting subsequent tokens~\citep{brown2020language, ouyang2022training}. 
Such strong capability has emerged significant success in downstream applications~\citep{lewis2020retrieval,yang2023harnessing}, yet the large neural network is known to be \textit{black-box}, where the representations in the model are not independently interpretable, introducing difficulty in designing effective paradigms for major known challenges of (visual) language models like hallucination~\citep{ji2023survey, tong2024eyes}, bias~\citep{garrido2021survey, nadeem2020stereoset}, and catastrophic forgetting~\citep{kemker2018measuring, zhai2024investigating}. 

To interpret the functions of individual modules in the language models, \textit{mechanistic interpretation} was proposed to reverse-engineer such models, through identifying meaningful patterns in the data representations and computational mechanisms of the model components~\citep{olah2022mech, meng2022locating}. Recent studies on auto-regressive models like GPT-2 have delved into \textit{neuron-level interpretation}, where the focus is on understanding the activations within the model's MLP layers~\citep{yun2021transformer}. This approach helps to reveal the specific roles of individual neurons, which is crucial for precise model editing and control~\citep{meng2022locating, meng2022mass}. Recent research has also proposed that \textit{sparse auto-encoder} (SAE)-based dictionary learning effectively promotes mono-semanticity of neurons, thus enhancing neuron-level interpretability~\citep{bricken2023towards}. 
However, as a post-hoc method, sparse auto-encoders introduce non-negative reconstruction loss, causing noise and reducing fidelity when interpreting neurons~\citep{bricken2023towards}. Recent studies also highlight their scalability challenges, as neuron decomposition becomes difficult in larger language models due to the vast number of directions~\citep{kissane2024attention, templeton2024scaling, rajamanoharan2024jumping}.

\textbf{\textit{Can we instead build sparsity directly into the language model?}} In this paper, we develop the \ours{} language model, a GPT-2-size language model that builds sparse coding into the model with a mathematically principled way. \ours{} handles the problems introduced by sparse encoders at scale: it \textbf{(i)} escapes the loss introduced in reconstructing the language model representations, enabling loss-free steering, and \textbf{(ii)} escapes the unsteady process of training a sparse auto-encoder. To avoid adding inconsistency into the language model, we develop on top of a mathematically principled white-box model framework, named \ours{}~\citep{yu2023white}.\footnote{In the remaining parts of the paper, we use ``\ours{}" or ``\ours{} language model" to refer to our language model architecture, while ``original \ours{}" denotes the architecture framework described in the literature.} After encoding the text tokens into numbers, we apply language-domain-specific modifications to the original \ours{} architecture and obtain the token representations. The final representations are then used to predict the next token, while the intermediate representations gets interpreted.

\jack{This paragraph is a condensed version of the related works. It should briefly mention (1) neuron activation interpretation: what it is, why it's important, and what are potential applications of the interpretation, and (2) the automatic evaluation metrics of the interpretability.}

\begin{figure}[t!]
     \centering
      \begin{subfigure}[b]{1.0\textwidth}
         \centering
    \includegraphics[width=\textwidth]{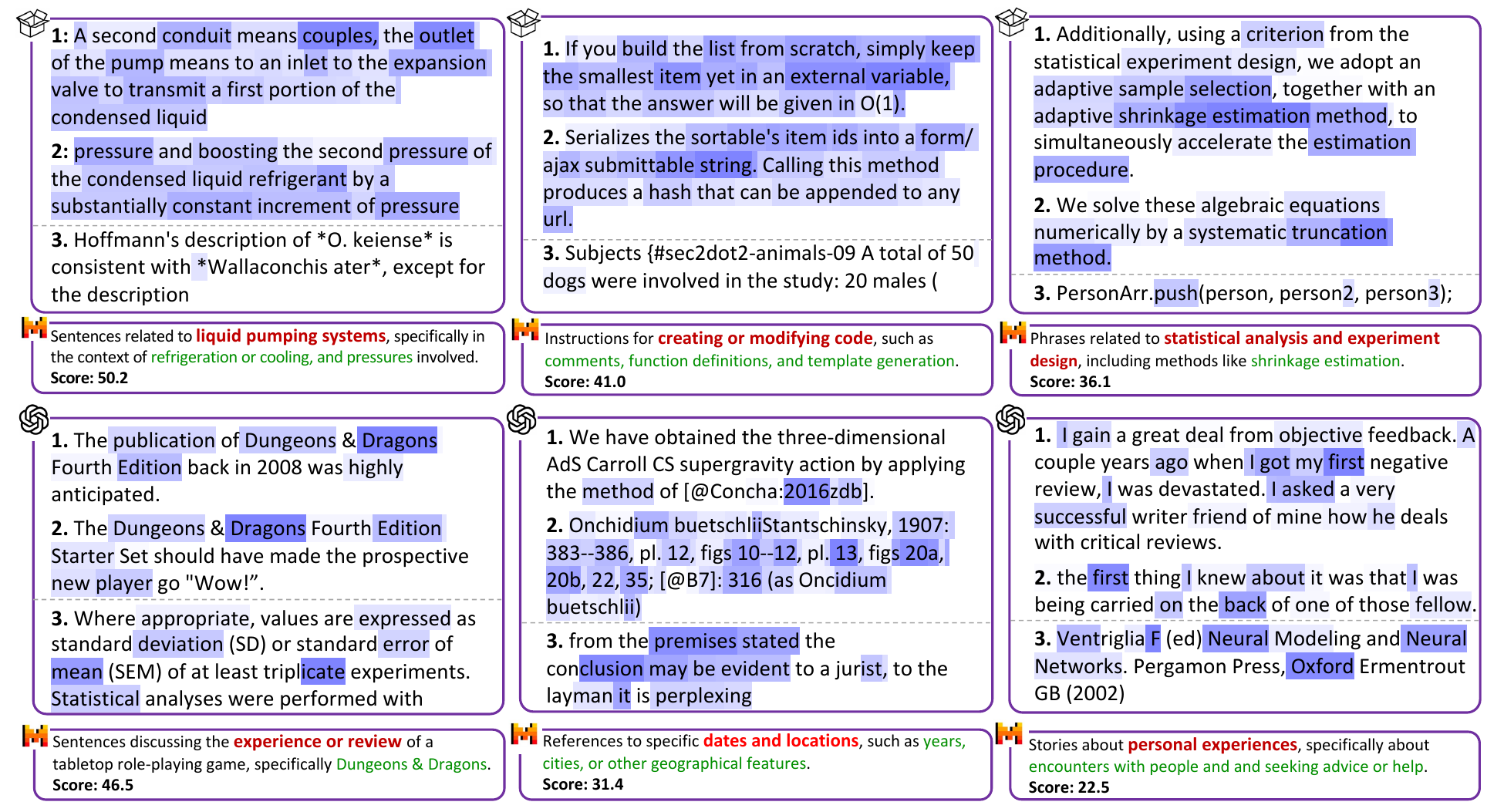}
     \end{subfigure}
        \caption{Instances are systematically identified where the interpretability of \StartCrate{} \ours{} (ours, \textit{row 1}) outperforms GPT-2 (\textit{row 2}). For each neuron (\textit{rounded box}), we show two top activated text excerpts (\textit{excerpt \textbf{1} and \textbf{2}}) and one randomly activated excerpt (\textit{excerpt \textbf{3}}). Results show that \ours{} consistently activates \textit{on and only on} semantically relevant text excerpts (first two excerpts), leading to more precise explanations predicted by agents like Mistral.}
        \label{fig:teaser}
\vspace{-4mm}
\end{figure}

To this end, the main contribution of this work is to propose a causal language model architecture based on the \ours{} model framework that builds sparsity inherently, that achieves significantly better neuron-level interpretability (up to \textbf{103\% relative increase}) than language models with the GPT-2 architecture under a similar configuration. \ours{} forms a family of models from single-layer model up to a 12-layer configuration. Comparative qualitative analysis of neuron activations between \ours{} and GPT-2 is provided in~\Cref{fig:teaser}, alongside extensive quantitative evaluations demonstrating that \textit{by explicitly integrating sparse coding into the language model, \ours{} achieves markedly improved interpretability across layers compared to GPT-2, applicable across a wide range of model sizes under a variety of evaluation metrics}.

\vspace{-2mm}
\section{Related Work}
\vspace{-2mm}

\textbf{Neuron-level interpretation.} Recent studies have provided insights into how auto-regressive models like GPT-2 work at the level of individual neurons, named \textit{neuron-level interpretation}. These studies focus on analyzing \textit{activations}, which are the outputs from the activation functions within the model's multi-layer perceptron (MLP) blocks~\citep{yun2021transformer}. Neuron-level interpretation is crucial to understanding the mechanisms in a model, including what concepts are learned in the neurons of the network, whether specific neurons are learning particular concepts, and how localized/distributed and redundantly the knowledge is preserved within neurons of the network. 

A higher neuron-level interpretability indicates that more neurons are interpretable or neurons are more interpretable~\citep{sajjad2022neuron}. As interpretations of the neurons can help localize the knowledge obtained in a neural network, neuron-level interpretation can be used for editing the knowledge in models~\citep{meng2022locating, meng2022mass}, model pruning and distillation~\citep{belinkov2020linguistic}, adapting the model to different domains and steering the output~\citep{erhan2009visualizing, rimsky2023steering}, and debugging model errors~\citep{hernandez2021natural}. Improved neuron-level interpretability increases reliability and performance in the applications above.

\textbf{Sparse auto-encoders.} To enhance interpretability, post-hoc sparse coding methods like dictionary learning~\citep{kreutz2003dictionary} are used, but these techniques result in imperfect reconstructions and thus always introduces loss when steering the model~\citep{arthur2023saearticle, bricken2023towards}. Literature also indicates that sparse-autoencoders are hard to scale, i.e., a dramatic drop in the number of interpretable features can be observed when the model being interpreted becomes deeper~\citep{kissane2024attention}. Additionally, tuning SAE models for larger $L$ values involves extensive hyperparameter tuning and time-consuming training, requiring multiple metrics (reconstruction rate, L1 loss, number of dead neurons) for reliable judgment, which can't be easily optimized with automatic engineering tricks~\citep{bricken2023towards}.

\textbf{Evaluation of neuron-level interpretability.} Metrics now exist to evaluate neuron-level interpretability in language models, examing if neurons trigger on relevant tokens in given contexts~\citep{bills2023language, bricken2023towards}. Recent works have demonstrated that a small number of circuits in language models are interpretable~\citep{wang2022interpretability, chughtai2023toy}, but comprehending each neuron, out of millions, is vital for thorough model safety audits. Given the prohibitive cost of human evaluation on such a scale, OpenAI introduced an automated metric using large language models for interpretability assessment~\citep{bills2023language}, which Anthropic later refined for sparse activations~\citep{bricken2023towards}. These methods align closely with human judgment and have gained broad acceptance within the research community~\citep{conmy2024towards, liu2023trustworthy, burns2023weak, lieberum2023does}. These metrics show that neuron-level interpretability in auto-regressive models is limited~\citep{sajjad2022neuron}, where the popular hypothesis is that neurons are superpositions of simpler semantics, which makes them \textit{fire} (produce a high activation) at multiple semantically distinct sets of tokens~\citep{elhage2022toy}. Metrics mentioned above are what we use in this work.

\jack{This paragraph continues the previous paragraph on the related work, and focuses on (1) the proposed \ours{} architecture: previous work in designing the architecture, (2) the main contribution of this work. It should stress on the motivation for us to interpret the activations in \ours{}, that is, to validate why we think the activations might be more interpretable in \ours{}. It should mention previous work in finding meaningful representations, why we think it will work better in the language interpretation setting.}

\textbf{White-box models and structured representation learning.} 
In the domain of structured representation learning, \textit{white-box} models stand out for their ability to generate explicit, structured data representations that adhere to specific, desirable configurations such as sparsity and piece-wise linearity, as discussed by~\citet{gregor2010learning} and~\citet{chan2021redunet}. Within this framework,~\citet{yu2023white} introduced an innovative approach to constructing deep networks based on unrolled optimization. Specifically,~\citet{yu2023white} proposed the \ours{} foundation model framework, utilizing an information-theoretic objective aimed at promoting the \textit{compression and sparsity} of data towards a predefined statistical structure.
Recently, empirical experiments suggest that the white-box design of \ours{} inherently develops segmentation capabilities from the data representations at both holistic and component levels with supervised training in the vision domain~\citep{yu2023emergence}, which directly motivates us to further explore the data representations within such architecture for language models. Recent work has also shown that the \ours{} framework is scalable: it can be effectively scaled up to comparable performance as Vision Transformer (ViT) with careful engineering~\citep{yang2024scaling}. Furthermore, the fine-tuning performance of the pretrained \ours{} model is also proven to be comparable in both the language domain~\citep{yu2023white} and vision domain~\citep{yang2024scaling}. 

\section{Preliminaries on the \ourscaps{} Architecture}

This section introduces the original \ours{} architecture introduced in~\citet{yu2023white}.

\textbf{Notations.} In this paper, we denote the one-hot input tokens by $\vX=[\vx_1, \dots, \vx_N]\in \R^{V\times N}$, where $\vx_i\in\R^{V\times 1}$ represents the $i$-th one-hot token, $N$ is the total number of input tokens, and $V$ is the vocabulary size. 
We use $f\in\mathcal F: \R^{V\times N}\rightarrow \R^{d\times N}$ to denote the mapping induced by the model, which is a composition of $L+1$ operators (layers) $f=f^L\circ \cdots f^\ell \circ \cdots f^1\circ f^\pre$, where $f^\ell:\R^{d\times N}\rightarrow \R^{d\times N} (1\le \ell \le L)$ represents the mapping of the $\ell$-th operator, and $f^\pre: \vX\in\R^{V\times N}\rightarrow \vZ^1\in\R^{d\times N}$ represents the pre-processing layer that transforms the one-hot token representations $\vX=[\vx_1, \dots, \vx_N]$ to semantic embeddings $\vZ^1=[\vz_1^1, \dots, \vz_N^1]$. 
We let $\vZ^\ell$ denote the input token representations of the $\ell$-th operator $f^\ell$ for $1\le \ell \le L$, so that $\vz_i^\ell\in\R^d$ denotes the representation of the $i$-th token $\vx_i$ before the $\ell$-th layer. We denote $\vZ=\vZ^{L+1}$ as the output token representations of the last ($L$-th) layer.

\textbf{Framework, objective, and optimization.} The transformation of input data into \textit{parsimonious} (piecewise linearized and compact) representations is accomplished by adopting a local signal model for the marginal distribution of the tokens \(\vz_{i}\). This statement suggests that the tokens can be approximately considered to occupy a union of several (identified as $K$) low-dimensional spaces, each with a dimension \(p \ll d\). These spaces are characterized by orthonormal bases, represented as \(\vU_{[K]} = (\vU_{k})_{k = 1}^{K}\) where \(\vU_{k} \in \bR^{d \times p}\).
Within the framework of this local signal model, \ours{} aims to optimize the \textit{sparse rate raduction} objective:

\vspace{-2mm}
\begin{equation}\label{eq:sparse-rate-reduction}
 \max_{f \in \mathcal{F}} \mathbb{E}_{\vZ}\big[ \Delta R(\vZ\mid \vU_{[K]}) - \lambda \|\bm{Z}\|_0 \big] = \max_{f \in \mathcal{F}} \mathbb{E}_{\vZ}\big[ R(\vZ) - \lambda \|\bm{Z}\|_0 - {R}^c(\vZ; \vU_{[K]}) \big].
\end{equation}

where $\lambda$ is the sparsification regularizer and $\vZ=f(\X)$. The coding rate \(R(\vZ)\) serves as a close estimate (following~\citet{ma2007segmentation}) for the average amount of bits necessary for encoding the tokens \(\vz_{i}\) to a  precision level \(\varepsilon\) using a Gaussian codebook. Additionally, \(R^{c}(\vZ \mid \vU_{[K]})\) represents the theoretical maximum average amount of bits needed to encode the projection of the tokens onto each low dimensional subspace defined in the local signal model, specifically  \(\vU_{k}^{*}\vz_{i}\), to the same precision level \(\varepsilon\) utilizing a Gaussian codebook, as outlined by~\citet{yu2023white}. If the subspaces are adequately incoherent from each other, the solutions that minimize the object function, viz. \Cref{eq:sparse-rate-reduction}, in terms of $\vZ$, are associated with subspace configurations that are both incoherent and aligned with the axes, as pointed out by~\citet{OriginalMCR2}.



A network aimed at optimizing the sparse coding rate reduction objective through unrolled optimization gradually shifts the distribution of $\X$ towards the intended canonical forms, where each iteration of the unrolled optimization process acts as a layer:

\vspace{-2mm}
\begin{equation}
f\colon \vX
\xrightarrow{\hspace{1mm} f^{\pre} \hspace{1mm}} \vZ^1 \rightarrow \cdots \rightarrow \vZ^\ell \xrightarrow{\hspace{1mm} f^\ell \hspace{1mm}} \vZ^{\ell+1} \rightarrow  \cdots \to \vZ^{L+1} = \vZ.
\label{eq:incremental}
\end{equation}

The iterative optimization framework incorporates multiple design choices, among which is a two-step alternating minimization approach grounded in robust theoretical principles~\citep{yu2023white}. This approach delineates two distinct blocks: the \texttt{MSSA} and the \texttt{ISTA} block, collectively defining a single \ours{} layer:

\vspace{-2mm}
\begin{equation}
\vZ^{\ell+1/2} \doteq \vZ^{\ell} + \texttt{MSSA}(\vZ^{\ell} \mid \vU_{[K]}^{\ell}), 
\qquad 
f^{\ell}(\vZ^{\ell}) = \vZ^{\ell+1}\doteq \texttt{ISTA}(\vZ^{\ell+1/2} \mid \D^\ell).
\end{equation}


\textbf{Compression operator: Multi-Head Subspace Self-Attention (MSSA).} Given local models $\vU_{[K]}^\ell$, to form the incremental transformation
$f^\ell$ optimizing \Cref{eq:sparse-rate-reduction} at layer $\ell$, \ours{} first compresses the token set $\vZ^{\ell}$ against the subspaces by
minimizing the coding rate $R^c(\,\cdot \mid \vU_{[K]}^\ell)$.
As \citet{yu2023white} show, doing this minimization locally by performing a step of gradient descent on \(R^{c}(\,\cdot \mid \vU_{[K]}^{\ell})\) leads to the so-called multi-head subspace self-attention (\texttt{MSSA}) operation, defined as

\begin{equation}
    \MSSA{\vZ \mid \vU_{[K]}} 
    \doteq \frac{p}{(N + 1)\varepsilon^{2}} \mat{\vU_{1}, \dots, \vU_{K}}\mat{(\vU_{1}\adj \vZ)\softmax{(\vU_{1}\adj\vZ)\adj(\vU_{1}\adj\vZ)} \\ \vdots \\ (\vU_{K}\adj \vZ)\softmax{(\vU_{K}\adj\vZ)\adj(\vU_{K}\adj\vZ)}},\label{eq:mssa-original}
\end{equation} 

In practice, the calculation of the intermediate representations $\vZ^{\ell+1/2}$ with the output from the \texttt{MSSA} block is calculated in a weighted form:

\vspace{-3mm}
\begin{equation}
    \vZ^{\ell + 1/2} \approx \left(1-\kappa\cdot\frac{p}{(N + 1)\varepsilon^{2}}\right) \vZ^{\ell} + \kappa
    \cdot \frac{p}{(N + 1)\varepsilon^{2}}\cdot \MSSA{\vZ^{\ell} \mid \vU_{[K]}},  \label{eq:gd-mcr-parts} 
\end{equation}
\vspace{-3mm}

where $\kappa > 0$ is a learning rate hyperparameter. This block resembles to the multi-head self-attention block of the GPT-2 model, but the query, key, and value projection matrices within a single head are all identical in the \texttt{MSSA} block.

\textbf{Sparsification operator: Iterative Shrinkage-Thresholding Algorithm (ISTA).} The remaining term to optimize in \Cref{eq:sparse-rate-reduction} is the
difference of the global coding rate $R(\vZ)$ and the $\ell^0$ norm of the
tokens, which together encourage the representations to be both sparse and
non-collapsed. \citet{yu2023white} show that local minimization of this
objective in a neighborhood of the intermediate representations $\vZ^{\ell +
1/2}$ is approximately achieved by a LASSO problem with respect to a
sparsifying orthogonal dictionary $\vD^{\ell}\in \R^{d\times h}$. Taking an iterative step towards solving this LASSO problem gives the iterative shrinkage-thresholding algorithm (\texttt{ISTA}) block~\citep{Wright-Ma-2022,yu2023white}. The ReLU non-linearity appearing in this block arises from an additional non-negativity constraint on the representations in the LASSO program, motivated by the goal of better separating distinct modes of variability in the token distribution:

\vspace{-2mm}
\begin{equation}\label{eq:ista-block-orig}
    \vZ^{\ell+1} = f^{\ell}(\vZ^{\ell}) = \operatorname{ReLU}(\vZ^{\ell+1/2} + \eta
    \D^{\ell*}(\vZ^{\ell+1/2} - \vD^{\ell}\vZ^{\ell+1/2}) - \eta \lambda \bm{1})
    \doteq \ISTA{\vZ^{\ell+1/2} \mid \vD^{\ell}}. 
\end{equation}

\section{The \ourscaps{} Language Model} \label{sec:method}
\vspace{-2mm}

This section introduces the difference between our work and the original \ours{} paper~\Citep{yu2023white}, thus introducing what changes we made to the \ours{} architecture. 

First, we note that the task in this work is different: we apply the \ours{} architecture to the \textit{next-token prediction task} in the language domain, while the original \ours{} paper applies the architecture to the image classification task in the vision domain. This difference immediately leads to differences when designing the model architecture: \textbf{(i)} we apply a causal mask to the original \ours{} model to avoid the model seeing tokens after the current token, and \textbf{(ii)} we change the embedding layer and heads of the original \ours{} model to fit the language vocabulary. 

Second, we're interested in interpreting the neurons within \ours{} on the next-token prediction task and making \textit{direct comparisons} to the GPT-2 model architecture. As neuron-level interpretation is commonly evaluated on the hidden states in the MLP block of the GPT-2 model~\citep{bills2023language, bricken2023towards}, \textbf{(iii)} we \textit{over-parameterize} (increase the hidden dimension of) the \texttt{ISTA} block of the original \ours{} model to align with the hidden dimension of the MLP block of the GPT-2 model.

\begin{figure}[t!]
     \centering
      \begin{subfigure}[b]{0.80\textwidth}
         \centering
    \includegraphics[width=\textwidth]{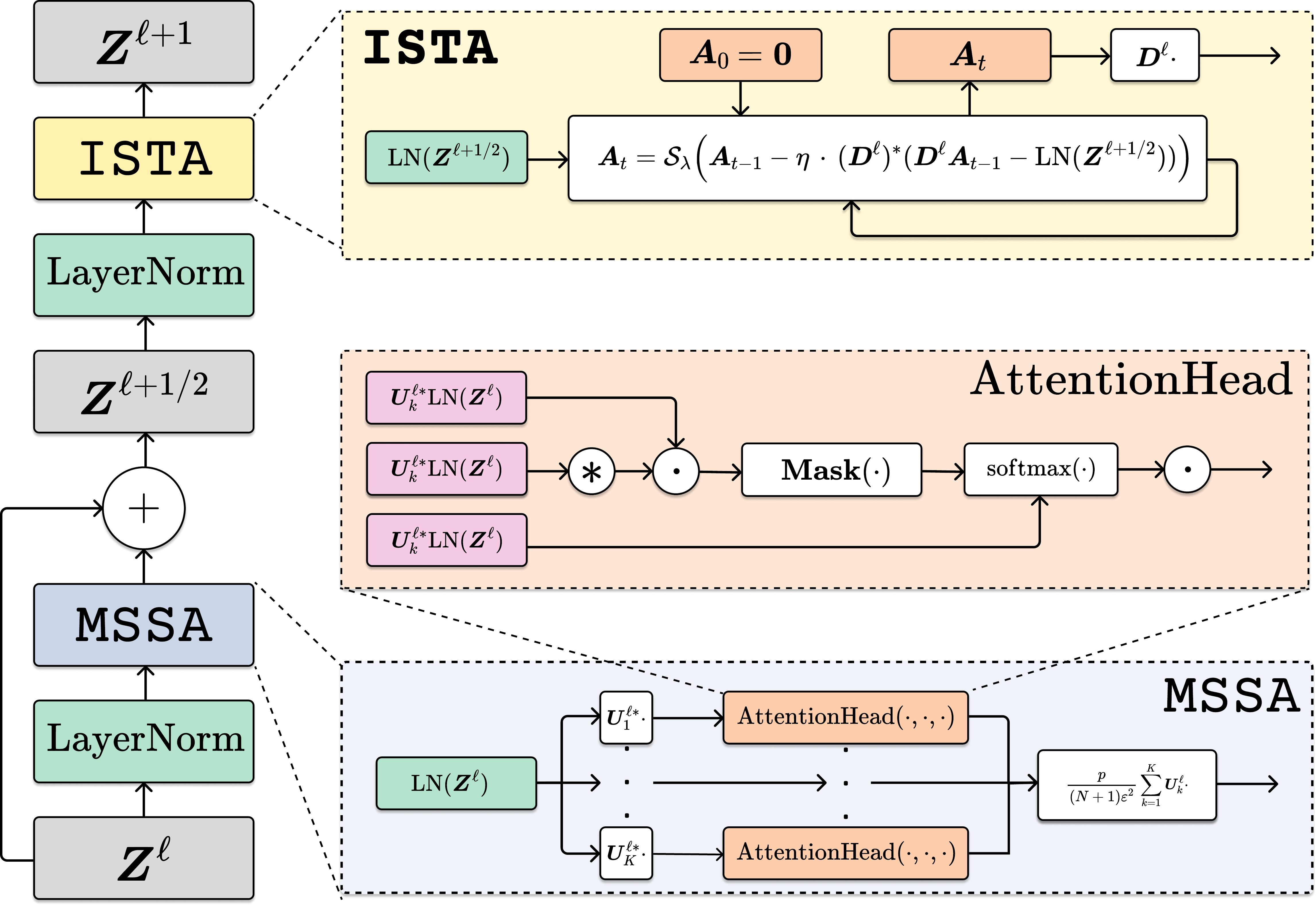}
     \end{subfigure}
        \caption{Block architecture for the \ours{} language model, where $S_{\lambda}(x) = \operatorname{ReLU}(x - \eta\cdot\lambda\cdot 1)$. Differences from the original architecture mentioned in~\citet{yu2023white} are marked \textbf{bold}: we \textbf{(i)} add a causal mask $\text{Mask}(\cdot)$ and \textbf{(ii)} over-parameterize the \texttt{ISTA} block.}
        \label{fig:crate-arch-overparam}
        \vspace{-2mm}
\end{figure}

Below we show specific definitions of the modifications we made. We illustrate the architecture in~\Cref{fig:crate-arch-overparam}, and show implementation details in~\Cref{app:impl-crate-blocks}.

\textbf{Embedding and Head.} In order to apply the \ours{} architecture to the language domain, we define the pre-processing layer $f^{\pre}$ that transforms language tokens into position-aware semantic embeddings, and define post-processing head $f^{\head}(\vZ)$ that maps the final representations to output token distributions in the vocabulary space:

\vspace{-6mm}
\begin{align}\label{eq:pre-head-def}
    f^{\pre}(\vX) = \vE_{\mathrm{sem}}(\vX) + \vE_{\pos}, 
    \qquad 
    f^{\head}(\vZ) = \vW^{\head}\vZ,
\end{align}
\vspace{-5mm}

where $\vE_{\mathrm{sem}}$ is a semantic embedding matrix that maps input tokens \(\vx_{i}\) to embedding vectors in \(\bR^{d}\), $\vE_{\pos}\in\mathbb{R}^{d\times N}$ is a positional embedding matrix, and \(\vW^{\head} \in \bR^{V \times d}\) maps the (contextualized) token representations $\vZ^{L+1}$ to the distribution of the next token. All parameters mentioned above are learnable.

\textbf{MSSA Block.} To align with the next word prediction task used in GPT-2~\citep{radford2019language}, we replace the attention matrix in \texttt{MSSA} (\Cref{eq:mssa-original}) with a \textit{causally masked} self-attention, defined as


\vspace{-4mm}
\begin{align}
\begin{split}
    \softmax{(\vU_{k}\adj\vZ)\adj(\vU_{k}\adj\vZ)} 
    &\to \softmax{\texttt{CausalMask}((\vU_{k}\adj\vZ)\adj(\vU_{k}\adj\vZ))}, \\
    \text{where}\ \texttt{CausalMask}(\vM)_{ij} 
    &= \begin{cases}\vM_{ij}, & i \leq j \\ -\infty, & i > j.\end{cases} \label{eq:causal-mssa}
\end{split}
\end{align}

\textbf{ISTA Block.} To investigate the neuron interpretability of the activation matrix $\vA\in \R^{h\times N}$, we design an \textit{overcomplete} version of the original \texttt{ISTA} block (\Cref{eq:ista-block-orig}) with $\vD^\ell\in\R^{d\times h}$ where $h=nd$, and $n=4$ to keep a fair comparison to the GPT-2~\citep{radford2019language} and standard transformer architecture~\citet{vaswani2017attention}):

\vspace{-6mm}
\begin{align}\label{eq:ista-block}
\begin{split}
    &\vA_{t} \doteq \ISTA{\vZ^{\ell+1/2} \mid \vD^\ell}, \\ 
    &\vA_{k} = \operatorname{ReLU}(\vA_{k-1} - \eta (\D^\ell)\adj(\vD^\ell\vA_{k-1} - \vZ) - \eta\cdot\lambda\cdot \bm{1}),\quad
    \vA_{0} = \boldsymbol{0},\,\, k\in[t], \\
    &\vZ^{\ell+1}=\vD^\ell\vA_t,
\end{split}
\end{align}
\vspace{-5mm}

where $\eta > 0$ is the step size, $\lambda > 0$ is the sparsification regularizer, and $t$ is the number of \texttt{ISTA} iterations. In practice, we set $t=2$ to keep computation efficient.
The \texttt{ISTA} block resembles the MLP block in the GPT-2 model, but with a relocated skip connection.

\begin{figure}[t!]
     \centering
     \begin{subfigure}[b]{1.0\textwidth}
         \centering
    \includegraphics[width=\textwidth]{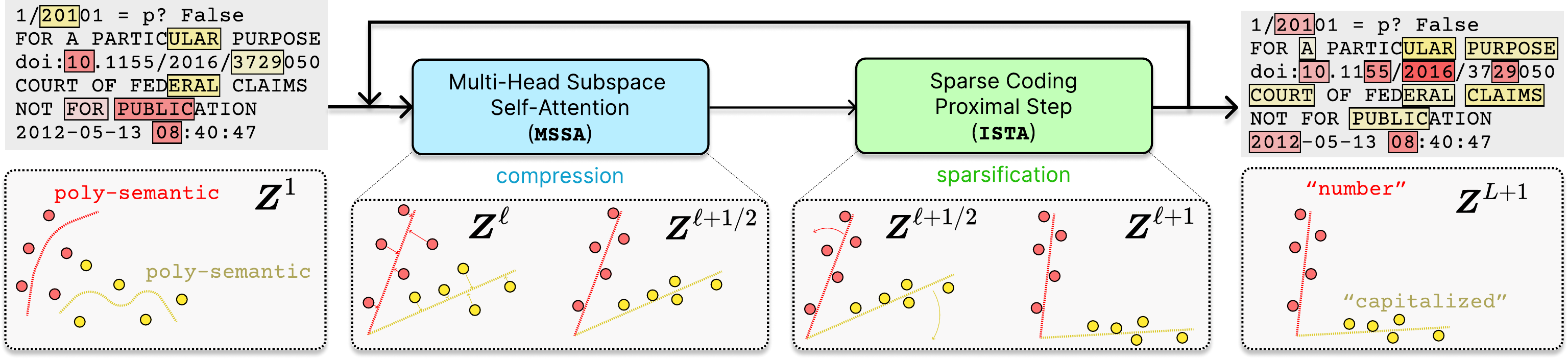}
     \end{subfigure}
        \caption{\ours{} iteratively compresses (\texttt{MSSA} block) and sparsifies (\texttt{ISTA} block) the token representations (\textit{colored points}) across its layers from $1$ to $L$, transforming them into parsimonious representations aligned on axes (\textit{colored lines}) with distinct semantic meanings.}
        \label{fig:learning-process}
        \vspace{-4mm}
\end{figure}

\textbf{Token Processing.} The desired token processing procedure is illustrated in~\Cref{fig:learning-process}. The process starts with random token representations ($\vZ^1$). Through successive layers, the representations ($\vZ^\ell$) are \textit{compressed} to align with the axis via the \texttt{MSSA} block, forming $\vZ^{\ell+1/2}$ that are semantically more consistent among relevant tokens. This is then refined by \textit{sparse coding} (the \texttt{ISTA} block) to produce the representations $\vZ^{\ell+1}$ that align on incoherent axes, leading to semantically more specified token representations. Repeated across layers, this culminates in distinct token representations $Z^{L+1}$ aligned on unique semantic axes. More detailed explanation of this optimization process can be found in~\Cref{app:learning-process}.

\vspace{-1mm}
\section{Empirical Experiments} \label{sec:experiments}
\vspace{-1mm}

\jack{This paragraph gives a brief introduction to the whole experiments section, on what messages we want to deliver in each subsection.}

We examine the next token prediction performance and neuron interpretability of \ours{} in this section. We detail the architecture, size, pre-training recipe (\Cref{sec:exp-setup}), performance (\Cref{sec:performance-at-scale}), and neuron-level interpretability (\Cref{sec:interpretability}) of \ours{} compared to the standard transformer architecture. In this section, we denote $K$ as the number of attention heads, $d$ as the dimension of the residual stream in the model, and $h$ as the hidden (inner) dimension of the \texttt{ISTA} or MLP module.

\jack{This paragraph defines the model architecture and calculate their size. It should deliver two core messages: (1) \ours{} is more lightweight than GPT, and (2) all comparisons in the experiments we conduct is apple-to-apple in terms of model configuration.}

\vspace{-2mm}
\subsection{Setup} \label{sec:exp-setup}
\vspace{-2mm}

\textbf{Model architecture and size.} The \ours{} model is designed with various sizes $L\in\{1,2,3,6,12\}$, where each size matches the GPT-2 configurations for direct comparisons, as shown in~\Cref{tab:configs-crate-text}. Configurations for $L\in\{1,2,3\}$ adhere to GPT-2 models as per~\citet{bricken2023towards}, while $L\in\{6,12\}$ follow configurations from~\citet{sanh2019distilbert} and~\citet{radford2019language}, respectively. Notably, \ours{} maintains approximately $2/3$ the size of GPT-2 at scale. Both models utilize the Byte-level BPE tokenizer with a vocabulary size of 50,257, following~\citet{radford2019language}. We explain the parameter size difference between \ours{} and GPT-2 in~\Cref{app:param-size-comparison}.

\begin{table}[H]
    \centering
    \label{tab:configs-crate-text}
    \footnotesize
    \setlength{\tabcolsep}{9.6pt}
        \begin{tabular}{lcccccc}
            \toprule
            {Model Config} & $d$ & $K$ & $L$ & $h$ & \ourscaps{} & GPT-2
            \\
            \midrule
            \textbf{1L} & 128 & 4 & 1 & 512 & 6.54M & 6.64M \\
            \textbf{2L} & 128 & 4 & 2 & 512 & 6.64M & 6.83M \\
            \textbf{3L} & 128 & 4 & 3 & 512 & 6.74M & 7.03M \\
            \textbf{S}(mall) & 768 & 12 & 6 & 3,072 & 55.9M & 81.1M \\
            \textbf{B}(ase) & 768 & 12 & 12 & 3,072 & 81.2M & 123.6M
            \\
            \bottomrule
        \end{tabular}%
        \vspace{2mm}
        \caption{\small Model configuration of \ours{} and model size comparison to GPT-2.}
        \vspace{-6mm}
\end{table}

\jack{This paragraph aims to give specific information on what datasets we're training on and abstract reproduction details. It should deliver the message that the training practice is grounded by existing literature and well defined.}

\textbf{Datasets and optimization.} We pre-train both models on the next token prediction task using the uncopyrighted \textit{Pile} dataset~\citep{gao2020pile} and the Adam optimizer~\citep{kingma2014adam}. Following~\citet{bricken2023towards}, we pre-train both \ours{} and GPT-2 of smaller sizes ($L\in\{1,2,3\}$) on 100 billion tokens with a context window of 1,024 tokens. 
Following the pre-training setup in~\citet{nanogpt} and scaling law in~\citet{touvron2023llama}, we pre-train using 100 billion tokens for the Small models, and 160 billion tokens for the Base models.\footnote{Practicaly, we train with a batch size of 768 for 125,000 iterations for $L\in\{1,2,3,6\}$, and a batch size of 256 for 600,000 iterations for $L=12$.} It takes about 4 days to pre-train \ours{}-Base on 160 billion tokens with 32 A5000 GPUs.


\vspace{-2mm}
\subsection{Performance} \label{sec:performance-at-scale}
\vspace{-2mm}

This section demonstrates that \ours{}, despite not outperforming GPT-2, still generates reasonable predictions, as evidenced through quantitative and qualitative comparisons.

\jack{This paragraph delivers the message that \ours{} effectively learns from the dataset and has a good performance in the end. By good performance, it should (1) have good training loss curve, (2) have nearly-identical validation loss as training loss in the end, (3) have good generalizability on other datasets, and (4) have comparable validation loss with the same-tier GPT-2 models.}

\begin{figure}[!t]
     \centering
     \begin{subfigure}[b]{0.44\textwidth}
         \centering
    \includegraphics[width=\textwidth]{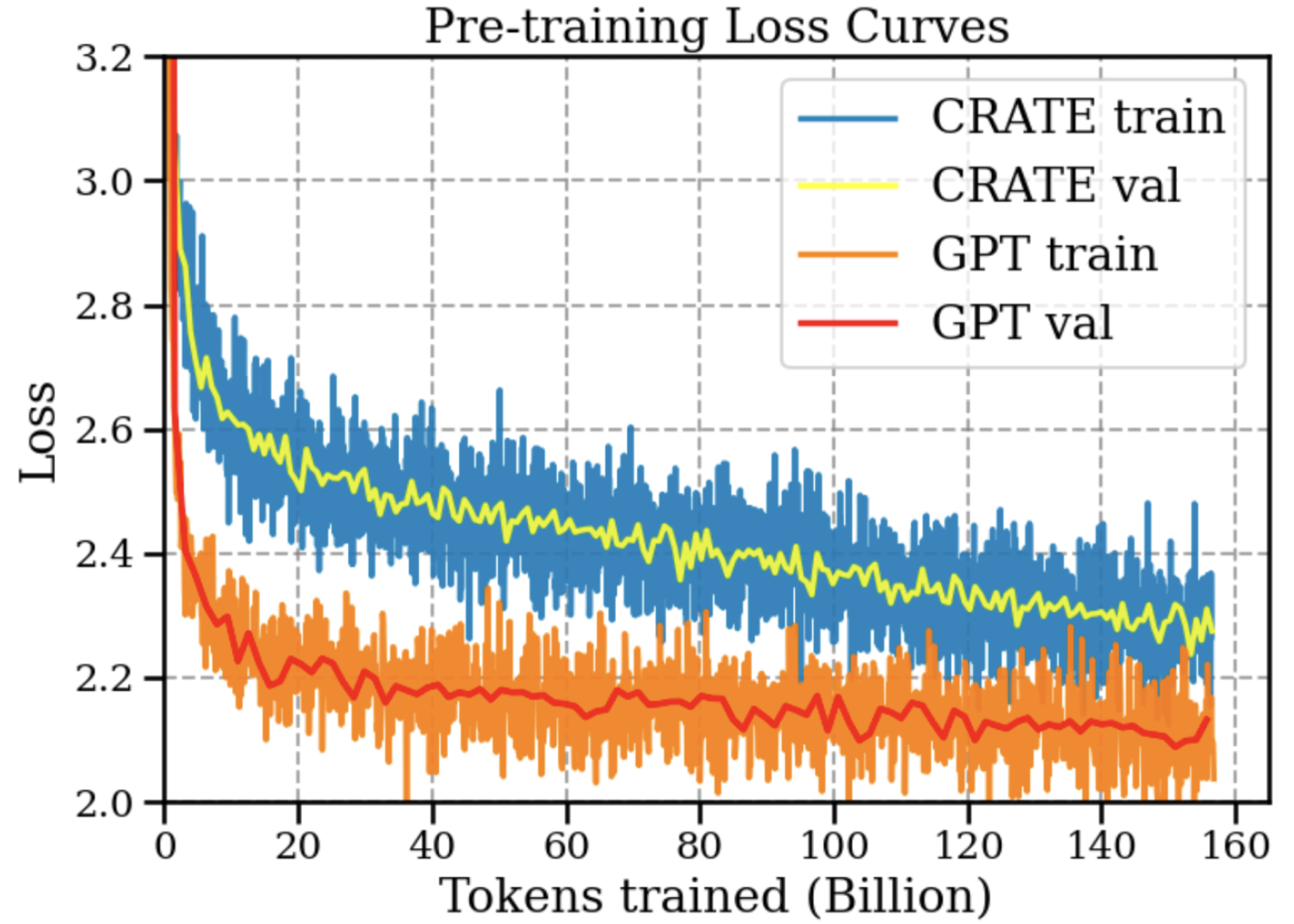}
     \end{subfigure}
    \begin{subfigure}[b]{0.44\textwidth}
         \centering
    \includegraphics[width=\textwidth]{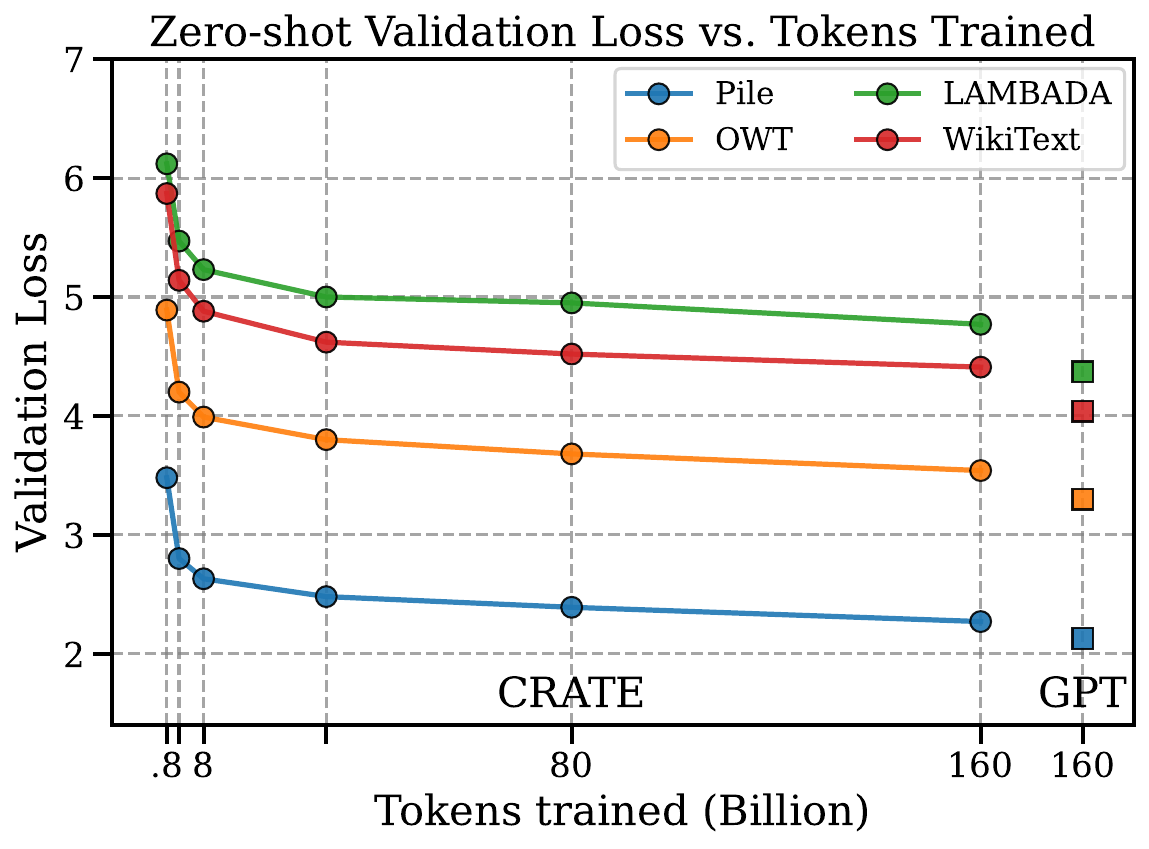}
     \end{subfigure}
        \caption{\small \textit{Left:} loss curve when pre-training \ours{}-Base and GPT2-Base on the Pile dataset.
        \textit{Right:} zero-shot validation loss of \ours{} evaluated on a variety of datasets (Pile, LAMBADA, OpenWebText and WikiText).}
        \label{fig:gpt-crate-loss}
\end{figure}

We observe that both training and validation loss curve of \ours{}-Base on the Pile dataset \textit{converges} well, as presented in~\Cref{fig:gpt-crate-loss} (\textit{left}). Although the convergence is slower than GPT-2, the loss curve of \ours{} keeps decreasing after training on 160 billion tokens, while GPT-2 already tends to converge. 

\begin{figure}[!t]
     \centering
    \begin{subfigure}[c]{0.40\textwidth}
         \centering
    \includegraphics[width=\textwidth]{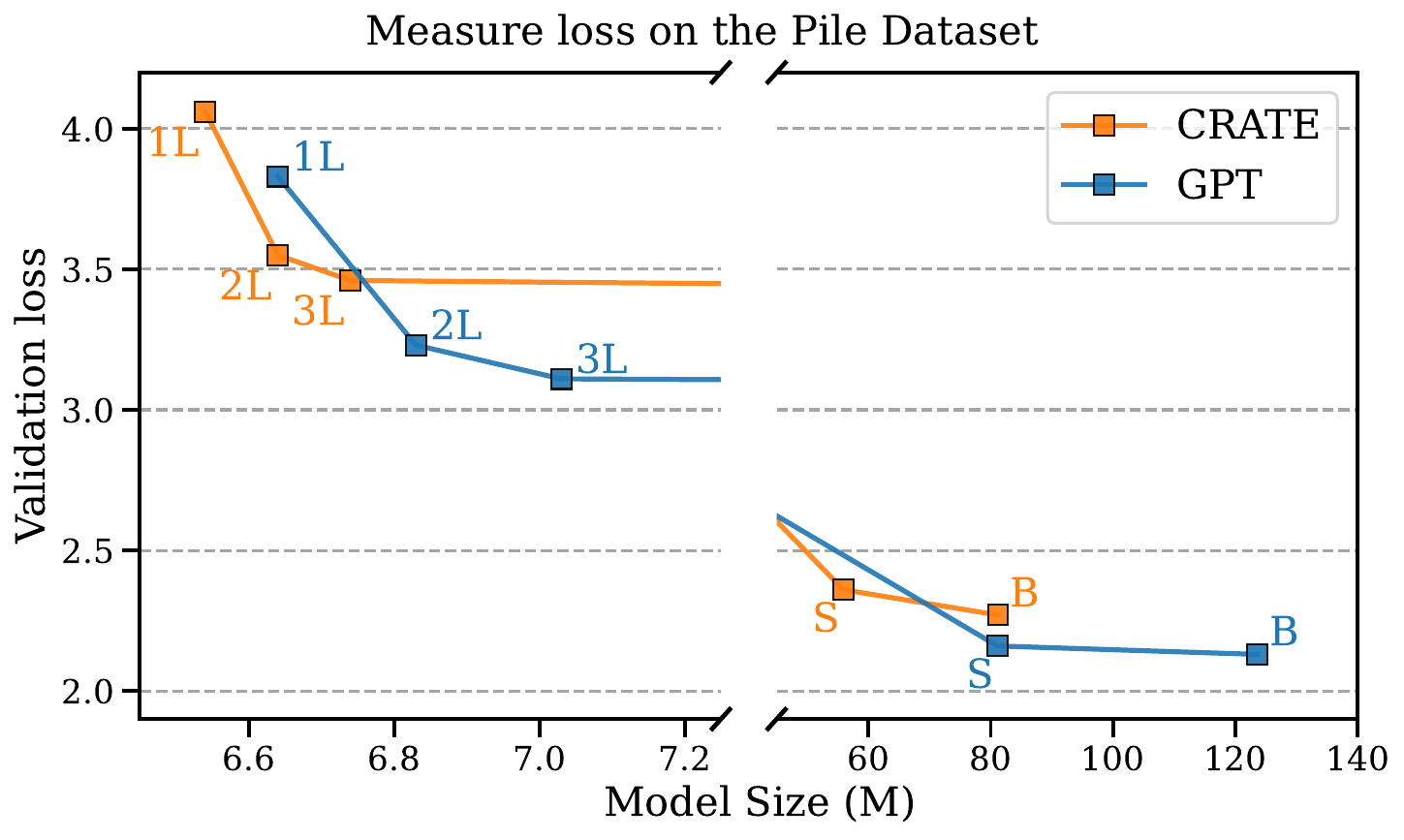}
     \end{subfigure}
     \begin{subfigure}[c]{0.58\textwidth}
         \centering
    \includegraphics[width=\textwidth]{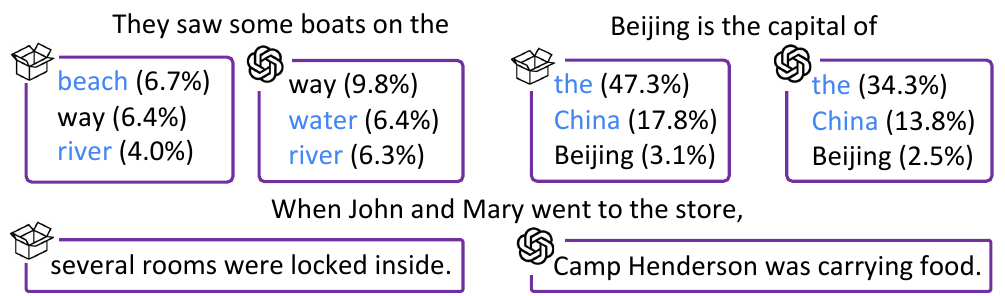}
     \end{subfigure}
        \caption{\small \textit{Left:} Validation loss of \ours{} compared to GPT-2 on the Pile dataset, with respect to the model size. \textit{Right:} Qualitative examples of predictions made by \ours{} and GPT-2. The tokens \textcolor{blue}{in blue} are considered good. We compare \StartCrate{} \ours{}-Base to GPT2-Base on the next word prediction task.}
        \label{fig:eval-gpt-quan-and-qual}
\vspace{-2mm}
\end{figure}

We also demonstrate the zero-shot validation loss curve of \ours{} evaluated on OpenWebText as well as other datasets~\citep{radford2019language} in~\Cref{fig:gpt-crate-loss} (\textit{right}). Results show that \ours{} effectively learns transferable representations across a number of datasets, and achieves comparable performance to GPT-2 after full training on the 160 billion tokens. 
We also demonstrate the \textit{scalability} of the \ours{} architecture by comparing the validation loss of \ours{} and GPT-2 with respect to the model size in~\Cref{fig:eval-gpt-quan-and-qual} (\textit{left}). Results show that the performance of \ours{} is close to GPT-2 across all model sizes. However, we do recognize that forcing sparsification in a model potentially leads to a higher compute cost on the next-token-prediction objective, which aligns with observations in~\citet{bricken2023towards} that \textit{enabling monosemanticity might hurt model performance}. Analysis on the sparsity of \ours{} and GPT-2 can be found in~\Cref{app:analysis-sparsity}.

Qualitative examples from \ours{} and GPT-2 are demonstrated in~\Cref{fig:eval-gpt-quan-and-qual} (\textit{right}). We conclude that \ours{} can make reasonable predictions, encouraging us to further look into its neuron-level interpretability.

\vspace{-2mm}
\subsection{Interpretability} \label{sec:interpretability}
\vspace{-2mm}

In order to quantitatively evaluate the interpretability of the neuron activations, we adopt the large language model-based algorithm introduced in~\citet{bills2023language} and~\citet{bricken2023towards}, as demonstrated in~\Cref{algo:interp-scoring}. We retrieve the sparse code $\vA_t$ (activations after the $\operatorname{ReLU}$ unlinearlity in the \texttt{ISTA} block) of \ours{} for interpretation, and compare to activations from the MLP block of GPT-2.

\begin{algorithm}
\caption{Interpretability Evaluation Algorithm~\citep{bills2023language}} \label{algo:interp-scoring}
\begin{algorithmic}[1]

\State \textbf{Inputs:} Input token set $\vS$ (in text form) and its activation matrix $\vA\in \R^{h\times T\times B}$ at $\ell$-th layer, where $T$ is the length of a single text excerpt, and $B$ is the number of text excerpts in the corpus.
\State \textbf{Models:} Explanation model $\mathcal F_1$
 , simulation model $\mathcal F_2$.

\For{$i\ \in [d]$}
    \State $\vS'\sim \vS, {\vA}'\in \R^{h\times T \times b}\sim \vA:$ \textit{Retrieve} $b$ text excerpts of $T$ tokens, together with the corresponding activation matrices.
    \State $k_i=\mathcal{F}_1(\vS', {\vA}'_{i,*}):$ \textit{Explain} common patterns in the retrieved activations of $i$-th neuron.
    \State $\tilde {\vA}_{i,*}'=\mathcal F_2(k_i, \vS'):$ Use the explanation to \textit{simulate} scores given only the tokens, not including true activations.
    \State $\rho_i = \rho(\vA_{i,*}', \tilde{\vA}_{i,*}'):$ Calculate \textit{correlation} between the accurate and simulated activations.
\EndFor
\State \textbf{Output:} Averaged interpretation score over all neurons $\rho=\E_{i\in [d]} (\rho_i)$.

\end{algorithmic}
\end{algorithm}

In practice, we adopt three evaluation metrics: two from OpenAI (\textit{top-and-random} and \textit{random-only})~\citep{bills2023language} and one from Anthropic~\citep{bricken2023towards}. We adopt the official implementation from~\citet{wu2023automated}, where details on the implementation are elaborated in~\Cref{app:impl-interp-metric}. 
Note that the Anthropic metric has much shorter text excerpts than the OpenAI metrics, so it is biased to sparse activations. For all evaluations, we discard the last layer of \ours{} and GPT-2, according to the empirical observation that the last layer of \ours{} is biased to the pre-training task~\citep{yu2023emergence}.

\jack{This paragraph delivers the message that \ours{} has much better interpretability than GPT across all three metrics (also mention the scalability).}

\begin{figure}[!t]
     \centering
    \begin{subfigure}[b]{0.32\textwidth}
     \centering
    \includegraphics[width=\textwidth]{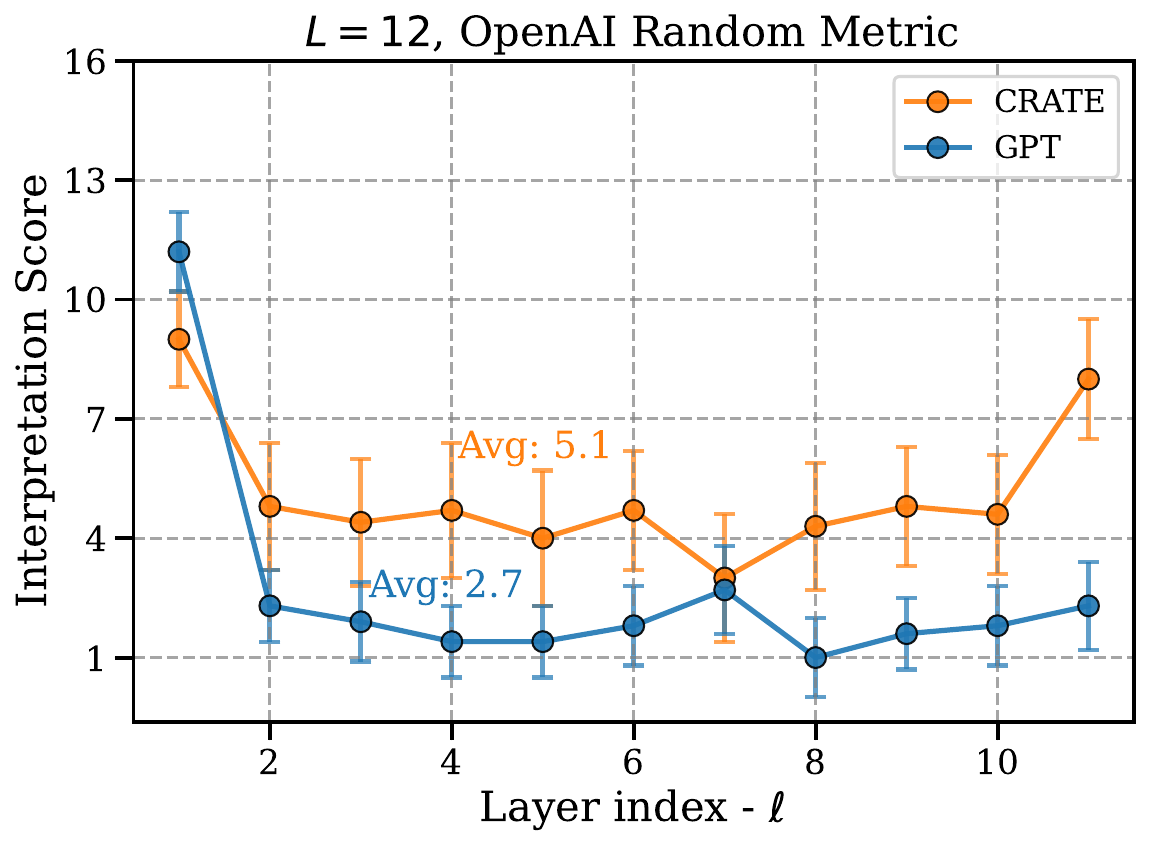}
     \end{subfigure}
     \begin{subfigure}[b]{0.32\textwidth}
     \centering
    \includegraphics[width=\textwidth]{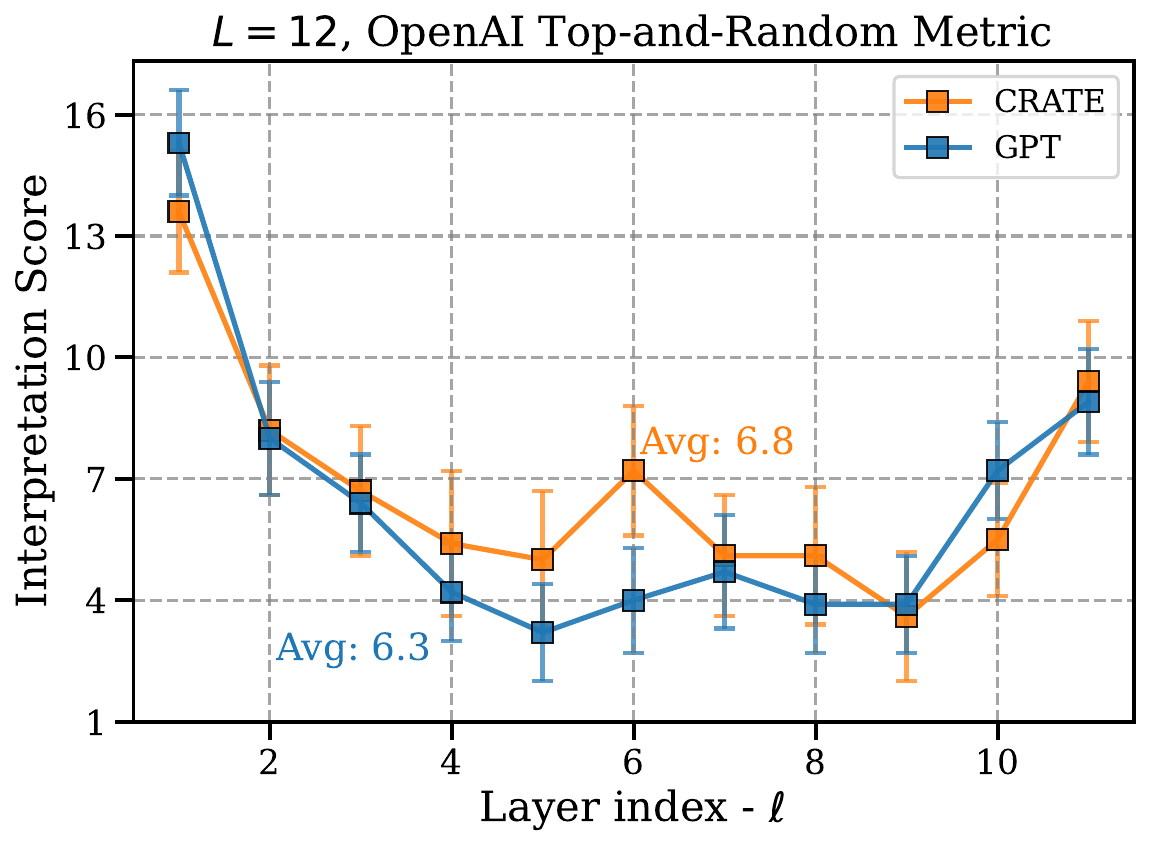}
     \end{subfigure}
     \begin{subfigure}[b]{0.32\textwidth}
     \centering
    \includegraphics[width=\textwidth]{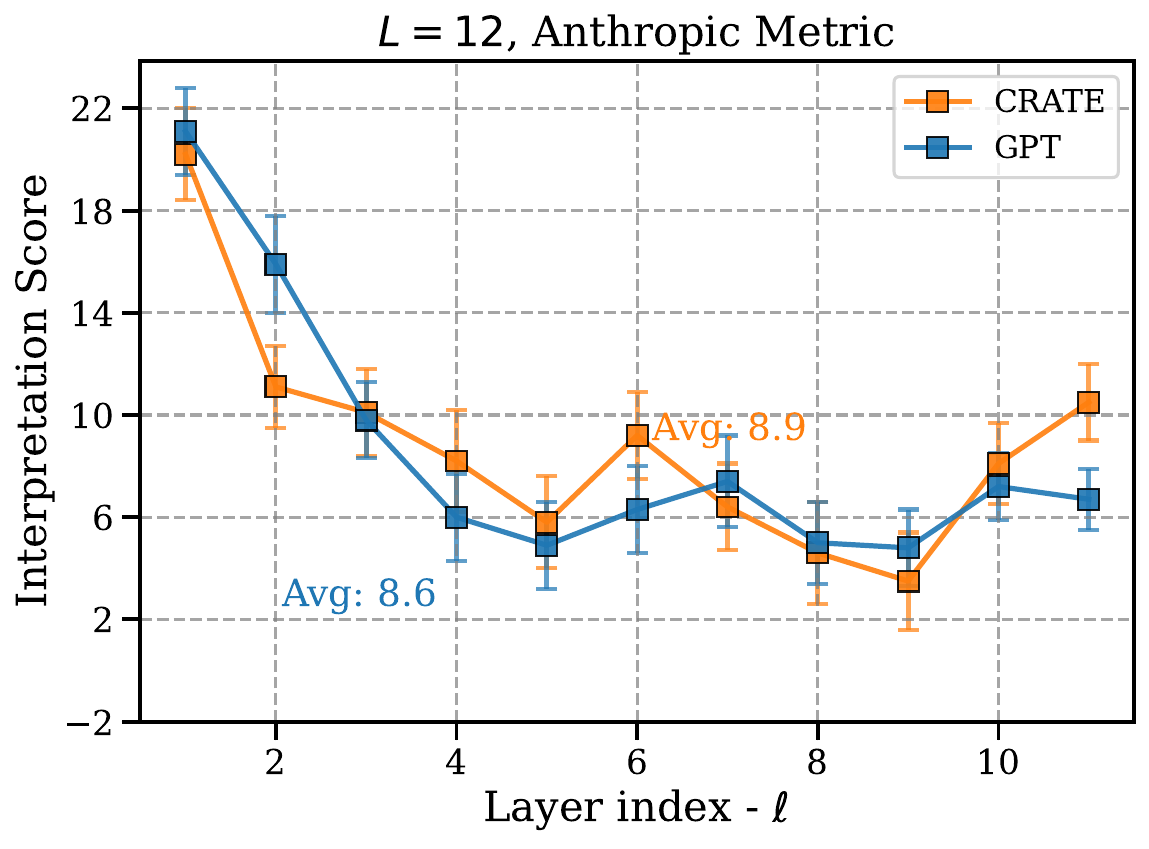}
     \end{subfigure}
    \begin{subfigure}[b]{0.32\textwidth}
     \centering
    \includegraphics[width=\textwidth]{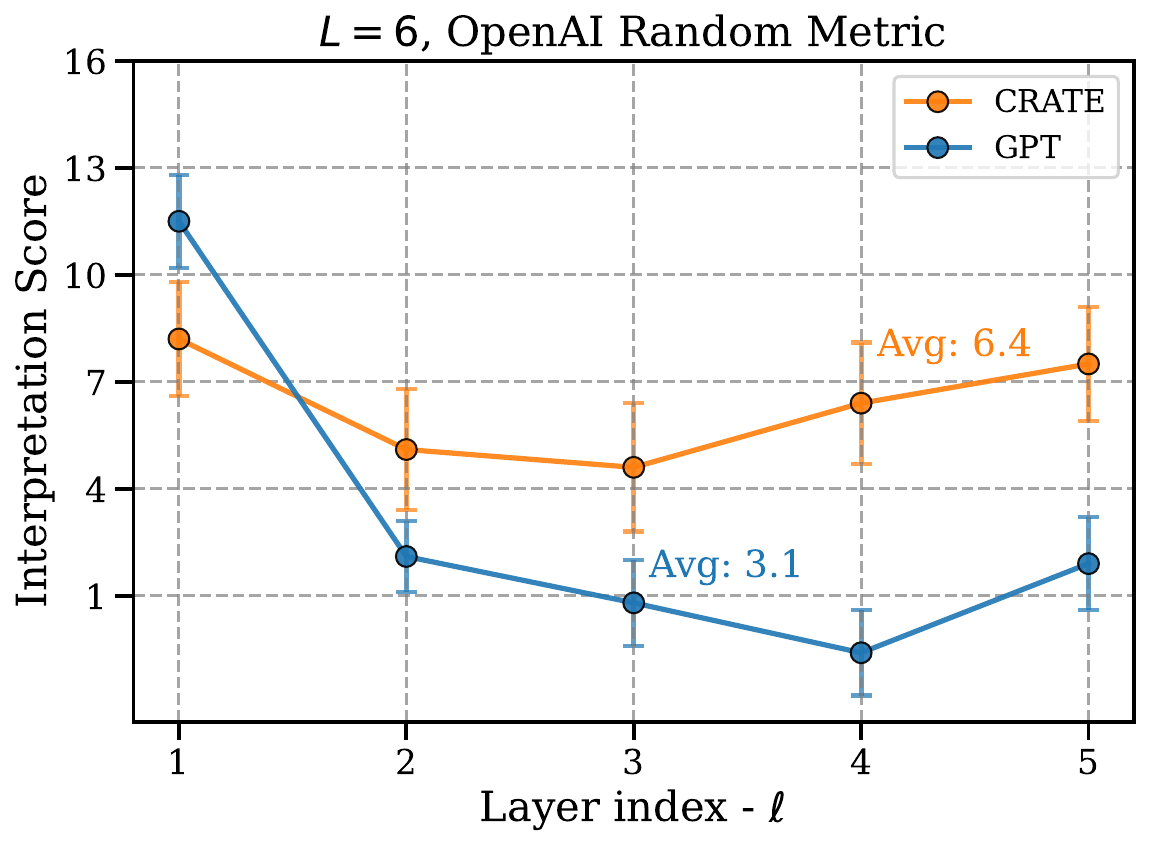}
     \end{subfigure}
     \begin{subfigure}[b]{0.32\textwidth}
     \centering
    \includegraphics[width=\textwidth]{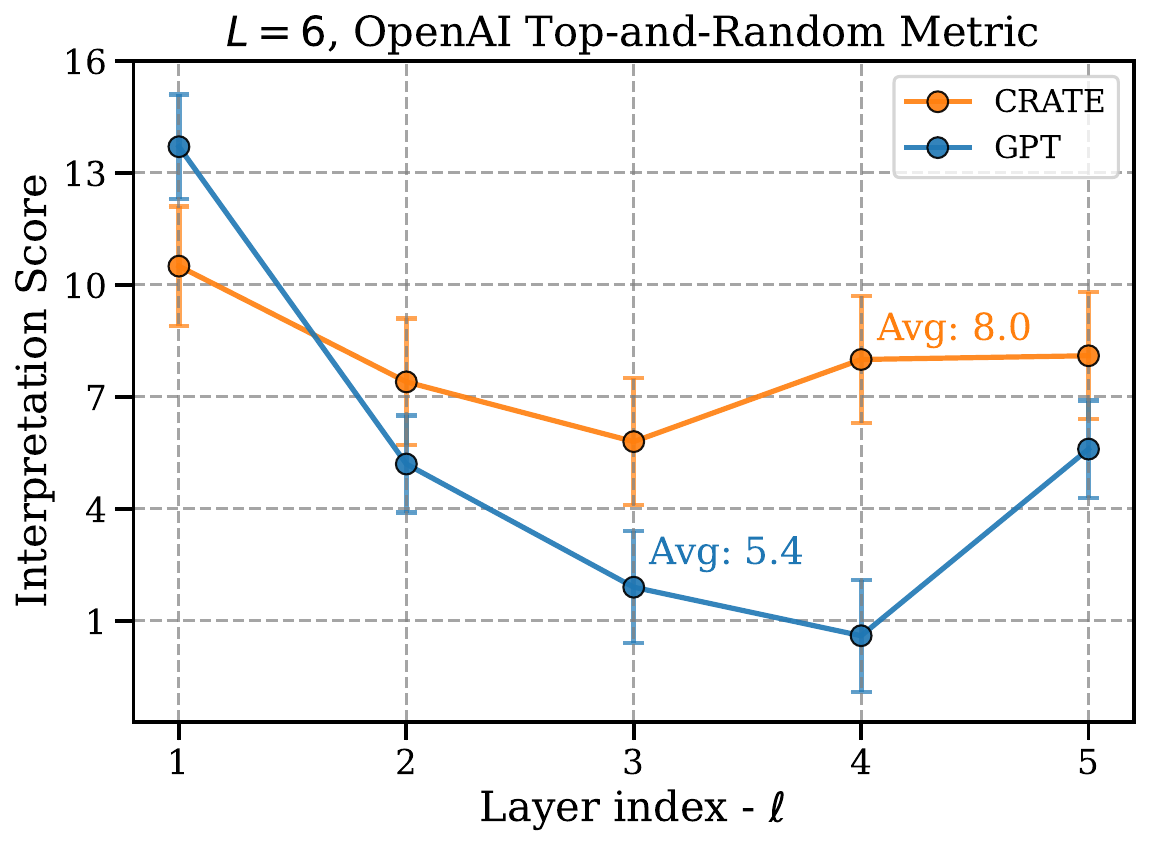}
     \end{subfigure}
     \begin{subfigure}[b]{0.32\textwidth}
     \centering
    \includegraphics[width=\textwidth]{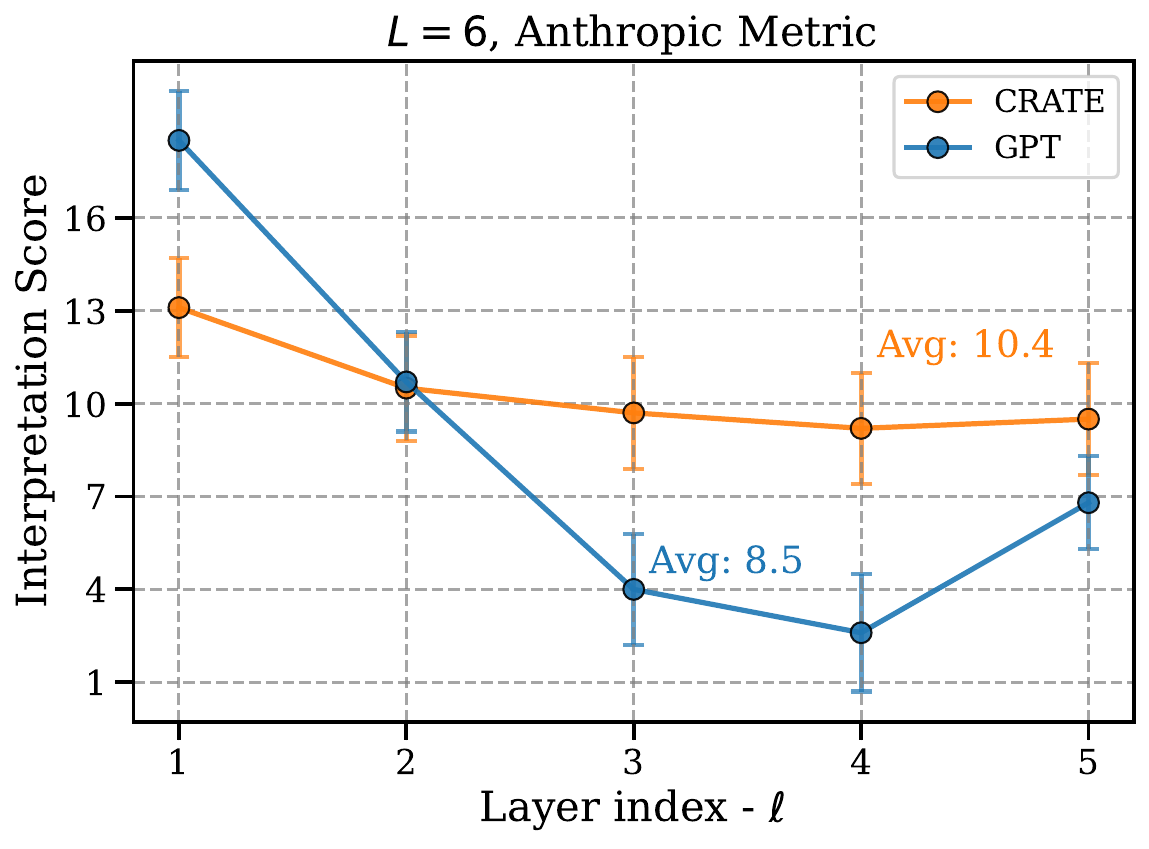}
     \end{subfigure}
     \begin{subfigure}[b]{0.32\textwidth}
         \centering
    \includegraphics[width=\textwidth]{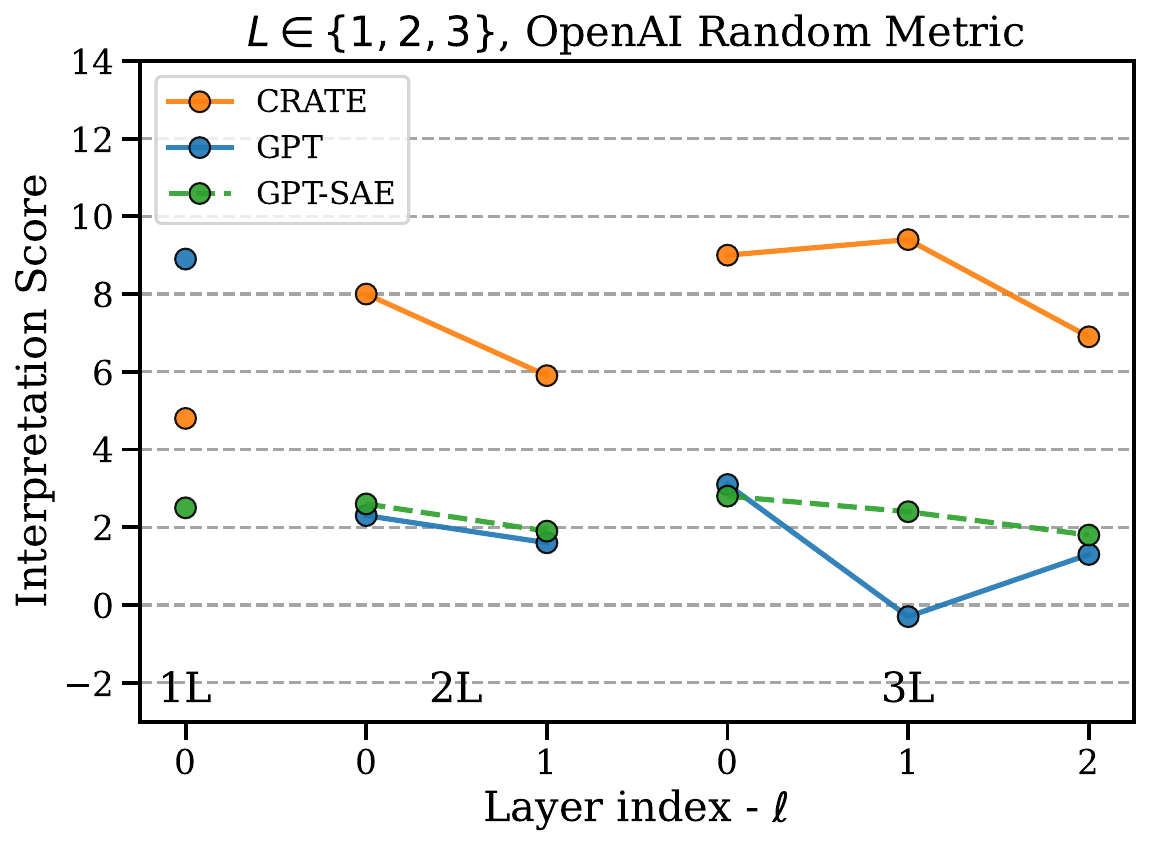}
     \end{subfigure}
     \begin{subfigure}[b]{0.32\textwidth}
         \centering
    \includegraphics[width=\textwidth]{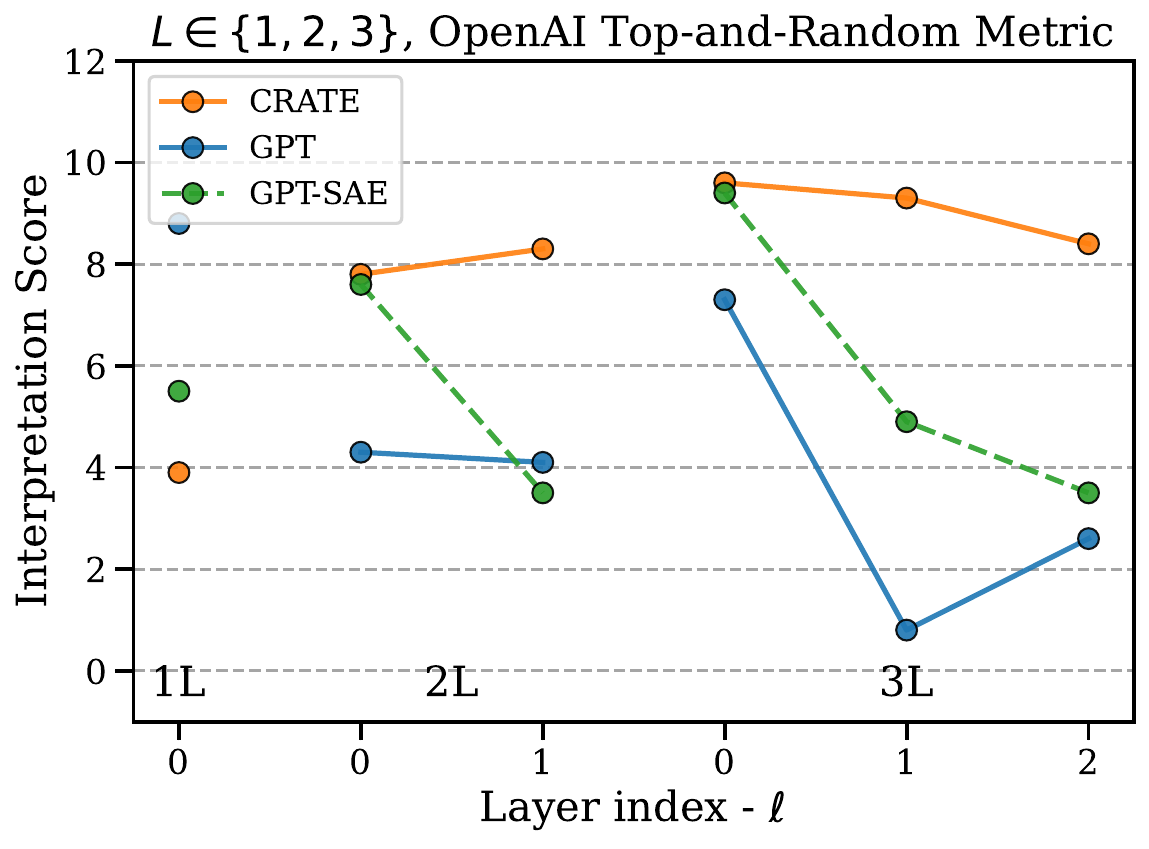}
     \end{subfigure}
     \begin{subfigure}[b]{0.32\textwidth}
         \centering
    \includegraphics[width=\textwidth]{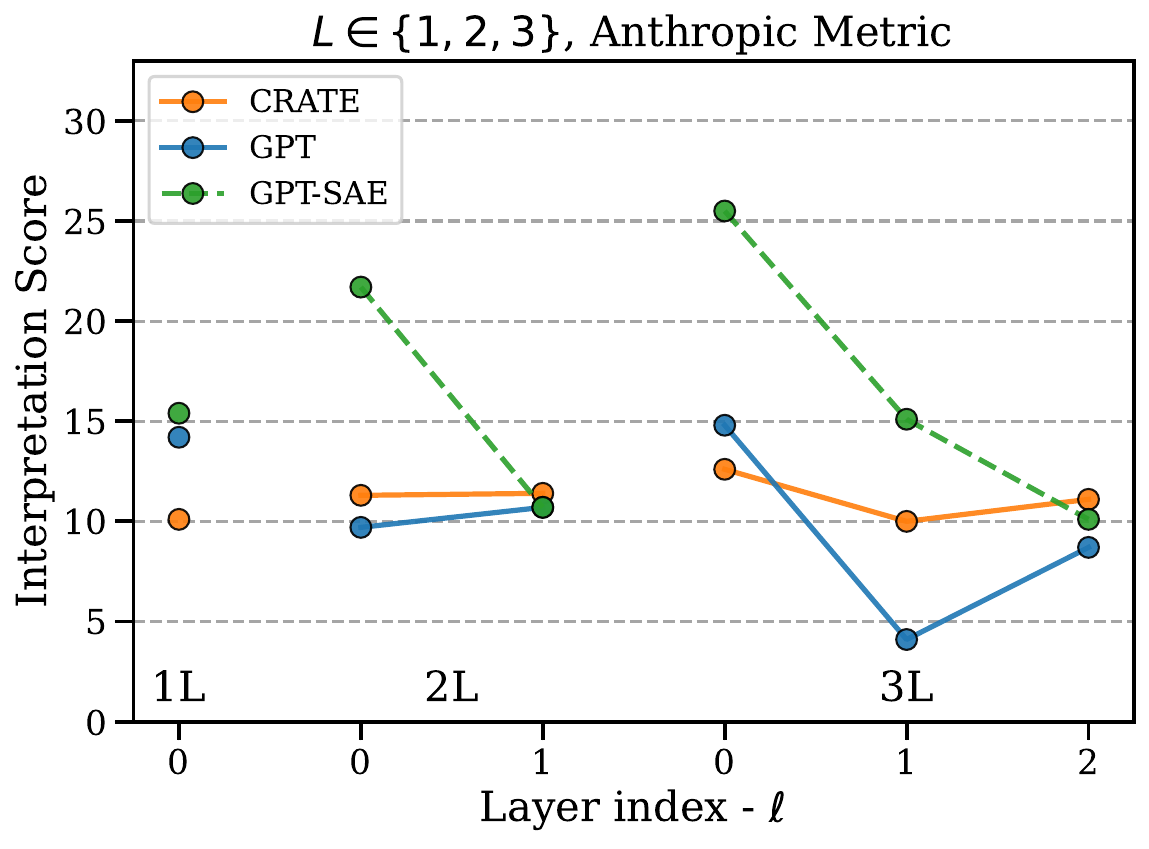}
     \end{subfigure}
        \caption{\small Interpretation scores evaluated using the OpenAI Random-only metric, Top-and-Random metric, and Anthropic metric, respectively. \textit{Top:} interpretation scores of \ours{} and GPT-2 for $L=12$. \textit{Middle:} interpretation scores of \ours{} and GPT-2 for $L=6$. \textit{Bottom:} interpretation scores of \ours{}, GPT-2, and GPT-2 with sparse auto-encoder for $L\in\{1,2,3\}$. Variance bars are normalized to $1/10$ of its original size.}
        \label{fig:gpt-crate-interp-comp}
        \vspace{-4mm}
\end{figure}

\textbf{\ourscaps{} achieves markedly improved and more steady neuron-level interpretability across layers compared to GPT-2, applicable across a wide range of model sizes.} We show evaluation results of the interpretability of \ours{} and GPT-2 averaged across layers in~\Cref{tab:interp-mean-and-var} (\textit{left}). 
We observe that the interpretability of \ours{} comprehensively outperforms GPT-2 on all metrics for $L\in\{2,3,6,12\}$. When averaging the mean interpretability across all metrics, \ours{} outperforms GPT-2 up to strikingly $103$\% relative improvement under the OpenAI Random-only metric when $L=6$. 
We also present the layer-wise interpretation scores in~\Cref{fig:gpt-crate-interp-comp}, which demonstrates that \ours{} achieves higher interpretability than GPT-2 on almost all layers. For detailed distributions of the layer-wise scores of \ours{}-Base compared to GPT2-Base on different metrics, refer to~\Cref{app:interp-distribution}.

\begin{table}[!t]
\def\arraystretch{1.1}
\scriptsize
\centering
\begin{tabular}{c|cc|cc|cc|cc|cc|cc}
\hline
& \multicolumn{6}{c|}{\textbf{Mean} ($\uparrow$, darker green means more interpretable)} & \multicolumn{6}{c}{\textbf{Variance} ($\downarrow$, darker red means less steady)} \\
\hline
& \multicolumn{2}{c|}{Top-and-Random} & \multicolumn{2}{c|}{Random-only} & \multicolumn{2}{c|}{Anthropic} & \multicolumn{2}{c|}{Top-and-Random} & \multicolumn{2}{c|}{Random-only} & \multicolumn{2}{c}{Anthropic} \\
& \ours & GPT-2 & \ours & GPT-2 & \ours & GPT-2 & \ours & GPT-2 & \ours & GPT-2 & \ours & GPT-2 \\
\hline
1L & \gcp{3.9} & \gcp{8.8} & \gcq{4.8} & \gcq{8.9} & \gcr{10.1} & \gcr{14.2} & \gcs{0.0} & \gcs{0.0} & \gct{0.0} & \gct{0.0} & \gcu{0.0} & \gcu{0.0}  \\ 
2L & \gcp{8.05} & \gcp{4.2} & \gcq{6.95} & \gcq{1.95} & \gcr{11.35} & \gcr{10.2} & \gcs{0.06} & \gcs{0.01} & \gct{1.1} & \gct{0.12} & \gcu{0.0} & \gcu{0.25}  \\ 
3L & \gcp{9.1} & \gcp{3.57} & \gcq{8.43} & \gcq{1.37} & \gcr{11.23} & \gcr{9.2} & \gcs{0.26} & \gcs{7.51} & \gct{1.2} & \gct{1.93} & \gcu{1.14} & \gcu{19.21}  \\ 
6L & \gcp{7.96} & \gcp{5.4} & \gcq{6.36} & \gcq{3.14} & \gcr{10.4} & \gcr{8.52} & \gcs{2.29} & \gcs{20.85} & \gct{1.87} & \gct{18.39} & \gcu{2.01} & \gcu{32.56}  \\ 
12L & \gcp{6.8} & \gcp{6.34} & \gcq{5.12} & \gcq{2.67} & \gcr{8.88} & \gcr{8.65} & \gcs{7.09} & \gcs{11.35} & \gct{2.83} & \gct{7.48} & \gcu{18.3} & \gcu{24.65}  \\ 
\hline
\end{tabular}
\vspace{2mm}
\caption{\small Mean and variance of the average interpretability across layers for different model sizes.}
\label{tab:interp-mean-and-var}
\end{table}

\jack{This paragraph delivers the message that \ours{} has better consistency of interpretability than GPT.}

The variances of the average interpretation scores of \ours{} and GPT-2 across layers are shown in~\Cref{tab:interp-mean-and-var} (\textit{right}). From the results we draw a solid conclusion that the interpretability of \ours{} is much more steady than GPT-2 across all model sizes.~\Cref{fig:gpt-crate-interp-comp} further demonstrates a clear pattern that, for all model sizes, \ours{} maintains a higher interpretability than GPT-2 among almost all layers.

\jack{This paragraph explains why \ours{} is more interpretable than GPT: \ours{} is more contextualized than GPT, in the sense that it activates on tokens mentioned in consecutive tokens and don't activate on irrelevant tokens.}

\textbf{The built-in sparse coding approach introduces consistent and specific neuron-level behaviors.} The strong  interpretability of \ours{} on the OpenAI Top-and-Random metric and the Anthropic metric, as shown in~\Cref{fig:gpt-crate-interp-comp}, indicates its consistent behavior on relevant tokens. These two methods contain a large portion of top-activated text excerpts, so they are valid for measuring whether the activations are consistent with the summarized explanation~\citep{bills2023language, bricken2023towards}. 
Additionally, the larger interpretability gap of \ours{} and GPT-2 on the OpenAI Random-only metric versus the Top-and-Random metric highlights the specificity of \ours{} in avoiding firing on irrelevant tokens. The random-only metric usually includes highly irrelevant text excerpts, so it effectively measures the capability of the language model to avoid activating on semantically irrelevant tokens~\citep{bills2023language}.

Qualitatively, we refer back to the interpretation examples shown in~\Cref{fig:teaser}. In this figure, we list three neurons from \ours{} (row 1) and GPT-2 (row 2), respectively. For each neuron, we show two top-activated text excerpts and one random excerpt. 
Results show that \ours{} is able to consistently activate on sementically similar tokens within the most relevant text excerpts, and does not activate on random tokens that are semantically distinguished from the top tokens. This promotes a more precise explanation given by the explanation model (\texttt{Mistral} in the figure).
On the other hand, GPT-2 is much worse at distinguishing tokens from different contexts, because it also has high activations on random text excerpts where the semantic meanings deviate far from the top activations. 

\jack{These paragraphs compares the interpretability and performance of \ours{} to GPT with sparse auto-encoders. The message is that although we have less monosemanticity than GPT-SAE (see Anthropic metric), we achieve better interpretability than GPT-SAE (see OpenAI metrics). Also mention that GPT-SAE takes lots of efforts to train and is lossy.}

\textbf{Comparing \ourscaps{} to GPT-2 with post-hoc sparse auto-encoders.} We follow~\citet{bricken2023towards} and train SAEs for models with layers $L\in\{1,2,3\}$, using output activations from GPT-2 on the Pile dataset's training split, leading to the GPT2-SAE model. Details on the SAEs’ architecture and training are in \Cref{app:sae}.

The interpretability scores of GPT2-SAE compared to \ours{} and GPT-2, as depicted in \Cref{fig:gpt-crate-interp-comp}, reveal that under the long-context OpenAI metrics, GPT2-SAE matches GPT-2 but falls short of \ours{}. This is attributed to its neuron activations becoming nearly 99\% sparse after sparse auto-encoding, diminishing interpretability in long contexts. Conversely, under the Anthropic metric, GPT2-SAE surpasses both GPT-2 and \ours{} in interpretability, corroborating findings in~\citet{bricken2023towards} that post-hoc approaches enhance short-context interpretability, often a sign of mono-semanticity. However, the interpretability of GPT2-SAE on the Anthropic metric decreases significantly when $\ell$ increases, while \ours{} remains steady, introducing concern in scalability. Further qualitative comparisons are can be found in \Cref{app:qual-neuron-interps}.


Besides its good performance on the Anthropic metric, the post-hoc dictionary learning approach requires considerable \textit{manual effort}. To get a taste, training a sparse auto-encoder for a single GPT-2 layer takes 4 hours when $h=512$ and a day when $h=3072$ on an A100 GPU. 

\begin{table}[H]
    \centering
    \begin{tabular}{@{}lcc@{}}
        \hline
         & \textbf{OpenAI TaR} & \textbf{Anthropic} \\ \midrule
        $\rho($\ours{}\textsc{-sae}$)$ - $\rho($\ours{}$)$ & \textbf{-10.2} & \textbf{+34.8} \\
        $\rho($\textsc{gpt{\tiny 2}-sae}$)$ - $\rho($\textsc{gpt-{\tiny 2}}$)$ & +6.5 & +38.1 \\ 
        \hline
    \end{tabular}
       \caption{\small \textbf{Interpretability improvement of \ours{} and GPT-2 after applying SAE.} Results are obtained by subtracting the interpretation scores of the language model and the SAE model trained on that language model.}
        \label{tab:interp-comparison-gpt-crate-sae}
\end{table}

\textbf{Does \ours{} have more optimal representation than GPT-2 in terms of interpretability?} Alternatively, we train SAE models upon the \ours{} model, denoted \ours\textsc{-sae}, and compare the interpretability improvement of \ours\textsc{-sae} over \ours{} to the interpretability improvement of GPT2-SAE over GPT-2. As shown in~\Cref{tab:interp-comparison-gpt-crate-sae}, the improvement of interpretability of \ours\textsc{-sae} over \ours{} is smaller than GPT2-SAE over GPT-2 under both OpenAI and Anthropic metrics. This suggests that \ours{} has more optimal representations than GPT-2 in terms of interpretability. Experimental details can be found in~\Cref{app:interp-crate-sae}.

\begin{figure}[!t]
     \centering
     \begin{subfigure}[b]{0.98\textwidth}
         \centering
    \includegraphics[width=\textwidth]{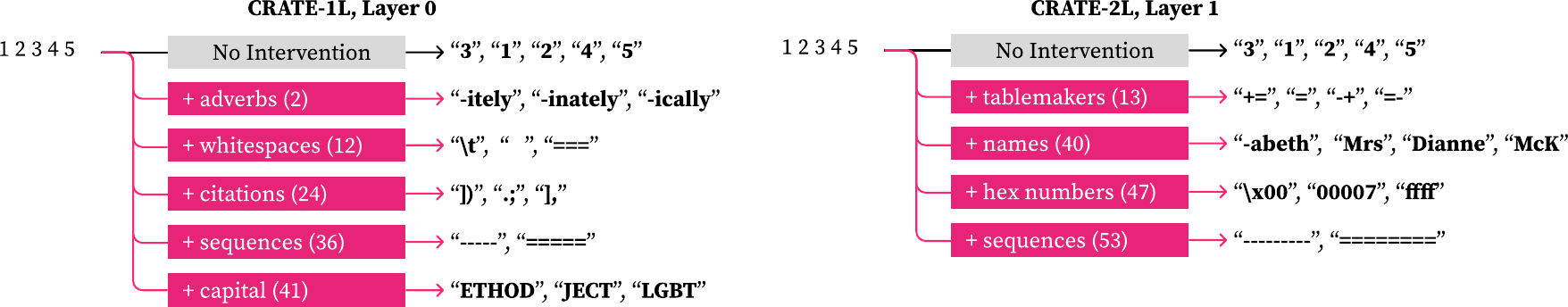}
     \end{subfigure}
        \caption{\small \textbf{Qualitative examples on logit effects of manually activating feature (i) in \ours{}.} Text shown on the right side are the most positive changes in token prediction probability. The logit effects align with feature interpretations.}
        \label{fig:crate-logit-effects}
\end{figure}

\textbf{Steering the \ourscaps{} model.} Following~\citet{bricken2023towards}, we manually activate some neurons and observe the \textit{logit effects} (changes of the token probability of the language model head). Some qualitative examples are shown in~\Cref{fig:crate-logit-effects}. Compared to the lossy steering of the SAE models, \ours{} can be steered without loss. Discussions on the lossy steering process can be found in~\Cref{sec:steer-lm-or-sae}.

\subsection{Ablation Study: Disentangling Interpretability from Other Factors} \label{sec:ablations}

Despite the superior interpretability of \ours{}, we're interested in whether this interpretability comes from some side factors. Specifically, we're interested in \textbf{(i)} whether it's due to the worse performance of \ours{} and \textbf{(ii)} whether it's because \ours{} has less parameters.

\textbf{Is the improved interpretability due to performance gap?} We first investigate whether the interpretability improvement is due to worse performance by comparing the interpretability of two different checkpoints of \ours{} (intermediate checkpoint and full checkpoint). Results in~\Cref{tab:interp-diff-ckpts} show that the interpretability of the intermediate checkpoint is lower than the full checkpoint, which suggests that “sacrificing” performance does \textit{not} necessarily introduce better interpretation scores.

\begin{table}[!htp]
    \centering
    \small
    \begin{tabular}{@{}lccccccccccccc@{}}
        \toprule
        \textbf{Checkpoint} & \textbf{Loss} & \textbf{0} & \textbf{1} & \textbf{2} & \textbf{3} & \textbf{4} & \textbf{5} & \textbf{6} & \textbf{7} & \textbf{8} & \textbf{9} & \textbf{10} & \textbf{Avg} \\ \midrule
        79B Tokens & 2.38 & 13.9 & 8.3 & 6.5 & 6.0 & 4.3 & 6.7 & 5.1 & 4.6 & 3.7 & 5.2 & 8.0 & 6.6 \\
        158B Tokens & 2.29 & 13.6 & 8.2 & 6.7 & 5.4 & 5.0 & 7.2 & 5.1 & 5.1 & 3.6 & 5.5 & 9.4 & 6.8 \\ \bottomrule
    \end{tabular}
    \vspace{2mm}
    \caption{\small \textbf{Validation loss and interpretability of the \ours{} model at different checkpoints.} The interpretation scores are measured under the OpenAI TaR metric.}
    \vspace{-2mm}
    \label{tab:interp-diff-ckpts}
\end{table}

Another piece of evidence is that \ours{}-2L has higher interpretability scores than \ours{}-1L. As shown in~\Cref{fig:eval-gpt-quan-and-qual} (\textit{left}), \ours{}-2L has much better performance than \ours{}-1L on the next-token prediction task. On the other hand, as shown in~\Cref{tab:interp-mean-and-var}, the interpretability of \ours{}-2L is also much higher than \ours{}-1L. Thus, a lower performance does not necessarily introduce higher interpretability.

\textbf{Is the interpretability gap due to number of parameters?} We observe that \ours{}-Base (81.2M) has a similar number of parameters as GPT2-Small (81.1M). However, results in~\Cref{tab:interp-mean-and-var} indicate that the interpretability of \ours{}-Base is higher than GPT2-Small on all metrics, and their layer-wise interpretation scores are also different in~\Cref{fig:gpt-crate-interp-comp}. This evidence suggests that two models with similar number of parameters does not necessarily have similar interpretability.

Another piece of evidence is that the interpretability of \ours{}/GPT2-1L all the way up to \ours{}/GPT2-12L does not have a consistent trend of increasing/decreasing interpretability, but their number of parameters both monotonously increases. This indicates that a model with larger number of parameters does not necessarily has better/worse interpretability.

\vspace{-1mm}
\section{Conclusion, Limitation, and Future Work} \label{sec:conc-and-future-work}
\vspace{-1mm}

In this paper, we demonstrated that replacing the standard transformer architecture with the white-box model \ours{} as a foundational architecture significantly improves the interpretability. Our empirical findings on the capability of \ours{} to be consistent and distinctive on the neuron-level activations underscore the importance of the white-box design in developing better language foundation models, fostering optimism that the introduction of built-in sparse coding approaches will catalyze further advancements in neuron-level interpretations.

Despite these findings, we acknowledge that the performance of \ours{} is not as good as GPT-2 on the next-token prediction task, which is potentially due to the introduction of the \texttt{ISTA} operator that introduces sparsity. This aligns with previous work suggesting that the performance might drop when explicitly introducing sparsity~\citep{bricken2023towards}. 
Future work should investigate towards a better trade-off between performance and interpretability of language models with built-in sparsity. It would also be meaningful to research on more qualitative mechanisms in the white-box language model and how to use these mechanisms for downstream edits.

\newpage







\bibliography{cpal_2025}
\bibliographystyle{plainnat}

\newpage

\appendix

\section{Details on the \ourscaps{} Architecture} \label{app:impl-crate-blocks}

\subsection{Parameter Size of \ours{} and GPT-2} \label{app:param-size-comparison}

\ours{} is smaller than GPT-2 because of the architecture difference. The vanilla GPT-2 architecture has two main parameterized blocks: Attention block and MLP block.

\textbf{Parameter size of the MSSA Block.} In \ours{}, the MSSA block resembles the Attention block, but instead of K, Q, V matrices, we only have one matrix. Therefore, compared to standard transformers, \ours{} uses $1/3$ of the parameters for the multi-head attention part.

\textbf{Parameter size of the ISTA block.} The MLP block in vanilla GPT-2 has one parametric matrix that transforms the input representations to the inner space (usually 4x larger), and another parametric matrix that transforms the inner representations back to the output space (as large as the input space). In \ours{}, the MLP block is replaced by the overcomplete ISTA block, which transforms the input representation to the overcomplete basis ($4\times$ larger) and transforms back with the same parametric matrix. Therefore, compared to standard transformers, \ours{} uses $1/2$ of the parameters for the MLP part.

\subsection{Causal MSSA Block} \label{app:impl-mssa-block}

This process can be implemented by PyTorch-like code shown in~\Cref{alg:torchcode-mssa}.

\begin{algorithm}[H]
    \caption{PyTorch-Like Code for Causal MSSA Forward Pass}
    \begin{lstlisting}[language=python]
class CausalMSSA(nn.Module):
    def __init__(self, config):
        super().__init__()
        assert config.n_embd % config.n_head == 0
        self.c_attn = nn.Linear(config.n_embd, config.n_embd, bias=False)
        self.c_proj = nn.Linear(config.n_embd, config.n_embd, bias=True)
        self.attn_dropout = nn.Dropout(config.dropout)
        self.resid_dropout = nn.Dropout(config.dropout)
        self.n_head = config.n_head
        self.n_embd = config.n_embd
        self.dropout = config.dropout
        self.register_buffer("bias", torch.tril(torch.ones(config.block_size, config.block_size)).view(1, 1, config.block_size, config.block_size)) # causal mask

    def forward(self, x, enhanced_feature_id=None):
        B, T, C = x.size()
        qkv = qkv.view(B, T, self.n_head, C // self.n_head).transpose(1, 2)
        att = (qkv @ qkv.transpose(-2, -1)) * (1.0 / math.sqrt(qkv.size(-1)))
        att = att.masked_fill(self.bias[:,:,:T,:T] == 0, float('-inf'))
        att = F.softmax(att, dim=-1)
        att = self.attn_dropout(att)
        y = att @ qkv
        y = y.transpose(1, 2).contiguous().view(B, T, C)
        y = self.resid_dropout(self.c_proj(y))
        return y\end{lstlisting}
        \label{alg:torchcode-mssa}
\end{algorithm}

\subsection{Overcomplete ISTA Block} \label{app:impl-ista-block}

To give a better idea of how~\Cref{eq:ista-block} works, we expand the two-iteration process ($t=2$). Given $\vD^\ell\in\R^{d\times h}$, we expand the first ISTA step to

\vspace{-2mm}
\begin{align}
\begin{split}
    \vA_0&=\mathbf{0}, \\
    \vA_1&=\mathcal S_\lambda\left(\vA_0-\eta\cdot(\vD^\ell)^*(\vD^\ell\vA_0-\mathrm{LN}(\vZ^{\ell+1/2}))\right) \\
    &= \mathrm{ReLU}\left(\eta\cdot(\vD^\ell)^*\mathrm{LN}(\vZ^{\ell+1/2})-\eta\lambda\right).
\end{split}
\end{align}

The second ISTA step continues the process from the initialized sparse code $\vA_1$:

\vspace{-4mm}
\begin{align}
\begin{split}
    \vA_2&=\mathcal S_\lambda\left(\vA_1-\eta\cdot(\vD^\ell)^*(\vD^\ell\vA_1-\mathrm{LN}(\vZ^{\ell+1/2}))\right)\\
    &= \mathrm{ReLU}\left(\vA_1-\eta\cdot(\vD^\ell)^*(\vD^\ell\vA_1-\mathrm{LN}(\vZ^{\ell+1/2}))-\eta\lambda\right),
\end{split}
\end{align}

which can be decomposed to:

\vspace{-4mm}
\begin{align}
\begin{split}
    \vG_1&=(\vD^\ell)^* \vD^\ell \vA_1 \\
    \vG_2&=(\vD^\ell)^* \cdot\mathrm{LN}(\vZ^{\ell+1/2}) \\
    \vG&=\eta\cdot(\vG_2-\vG_1)-\eta\cdot\lambda \\
    \vA_2&=\mathrm{ReLU}(\vA_1+\vG)
\end{split}
\end{align}

where $\vA_2$ is the output sparse code. At last, we convert the output sparse code from the coding rate space back to the original representation space:

\vspace{-4mm}
\begin{align}
    \vZ^{\ell+1}=\vD^\ell \vA_2
\end{align}

This process can be implemented by PyTorch-like code shown in~\Cref{alg:torchcode-ista}.

\begin{algorithm}[H]
    \caption{PyTorch-Like Code for Over-complete ISTA Forward Pass}
    \begin{lstlisting}[language=python]
class ISTA(nn.Module):
    def __init__(self, config):
        super().__init__()
        self.weight = nn.Parameter(torch.Tensor(4 * config.n_embd, config.n_embd)) # h*d
        with torch.no_grad():
            init.kaiming_uniform_(self.weight)
        self.step_size = 0.1
        self.lambd = 0.1

    def forward(self, x, enhanced_feature_id=None):
        z_init = F.relu(self.step_size * F.linear(x, self.weight, bias=None) - self.step_size * self.lambd) # A1
        x1 = F.linear(z_init, self.weight.t(), bias=None)
        grad_1 = F.linear(x1, self.weight, bias=None)
        grad_2 = F.linear(x, self.weight, bias=None)
        grad_update = self.step_size * (grad_2 - grad_1) - self.step_size * self.lambd
        output_sparse_code = F.relu(z_init + grad_update) # A2
        output = F.linear(output_sparse_code, self.weight.t(), bias=None)
        return output\end{lstlisting}
        \label{alg:torchcode-ista}
\end{algorithm}

\subsection{Details of Processing Tokens} \label{app:learning-process}

This section discusses the interpretation of~\Cref{fig:learning-process}.

\textbf{In~\Cref{fig:learning-process}, what is the space the points are drawn in?} The space is the representation space (of layer $\ell$). Because the model is pretrained with next-token prediction in the language domain, the space is specifically a semantic space. Thus, each point (token) has a semantic representation in this high-dimensional space.

\begin{figure}[H]
     \centering
     \begin{subfigure}[b]{0.95\textwidth}
         \centering
    \includegraphics[width=\textwidth]{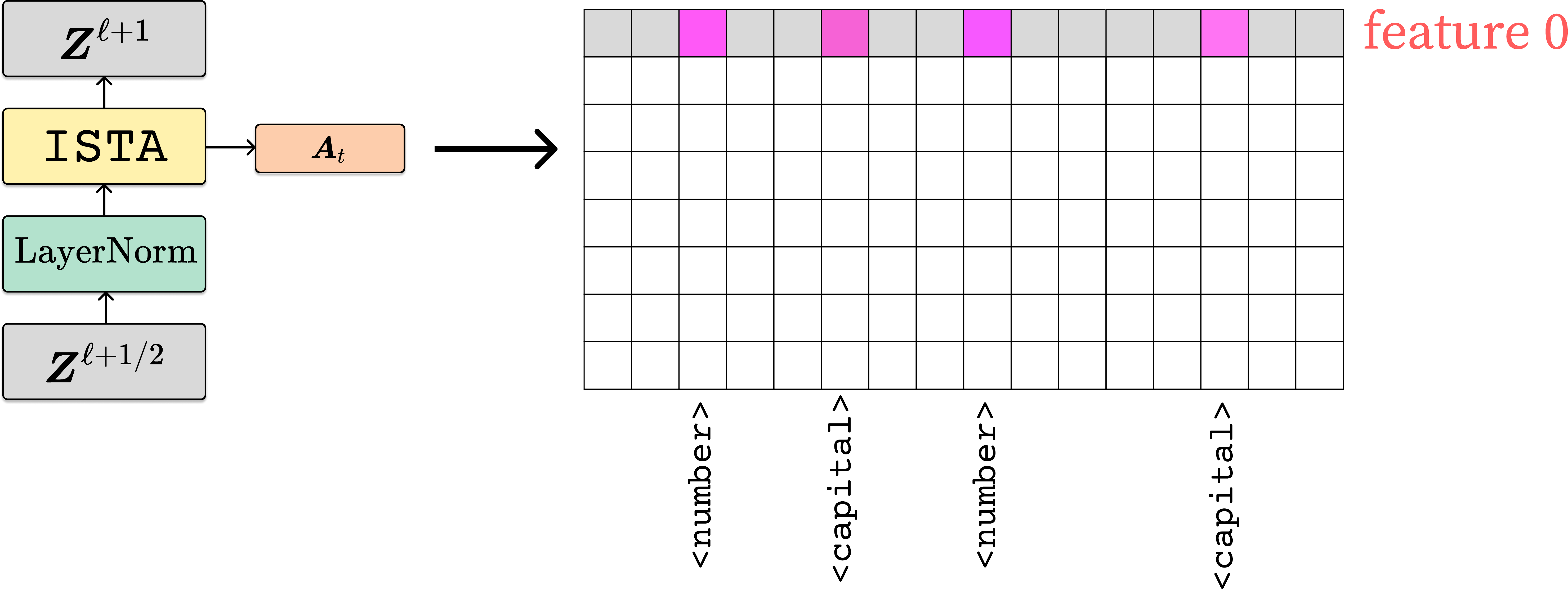}
     \end{subfigure}
        \caption{Illustration of the concepts of the activation matrix and poly-semanticity.}
        \label{fig:extract-tokens}
        \vspace{-4mm}
\end{figure}

\textbf{How do the layouts of the points suggest mono or poly-semanticity?} First, each axis (red/yellow) in the figure represents a neuron/feature visualized in the semantic space. We visualized the activation matrix $\boldsymbol{A}_t$ in~\Cref{fig:extract-tokens}. For example, when $L\in\{1,2,3\}$, the model dimension is $128$, which means that the overcomplete basis of ISTA will have a dimension of $512$, introducing $512$ features. Now if we input a sequence of $256$ tokens, the activation matrix will have a shape of $[512, 256]$.

\textit{Poly-semanticity} means that token representations in the semantic space are clustered as a broader set of semantic meanings, i.e., each neuron has a broader set of semantic meanings. For example, in the gray box on the top left of~\Cref{fig:learning-process}, both yellow- and red-backgrounded tokens represent either a number or a capitalized token. This corresponds to multiple high activations in the feature, where the tokens that activated this feature can either be a number or a capitalized token, which is shown in~\Cref{fig:extract-tokens} (where pink squares represent high activations).

In the \textit{compression} phase, the token representations are pushed towards the semantic axes, so that the tokens will activate on fewer features but will gain higher activations on these features, which is essentially an activation condensing process. In the \textit{sparsification} phase, the neurons (axes) are made further from each other, meaning that the features have less semantic overlap with each other. In this case, the results will become the gray box on the top right side of~\Cref{fig:learning-process}, indicating that each neuron has distinct semantic meanings, like ``numbers" or ``capitalized tokens". Note that this is a minimal example. In practice, tokens appear in context.

\jack{This section describes practical settings of how we implement the interpretation evaluation, such as how many activation samples are used, what models are used, what each metric evaluates for, and how we reduce the compute cost and make the automated evaluation more accessible.}

\section{Details on Interpretability Evaluations} \label{app:impl-interp-metric}

This section details the implementation details of the interpretability evaluations. In practice, we adopt three evaluation metrics: two from OpenAI~\citep{bills2023language} and one from Anthropic~\citep{bricken2023towards}. As the Anthropic metric is a closed-source follow-up of OpenAI, we start from the official implementation provided by OpenAI~\citep{wu2023automated} for both metrics. For each layer, we use randomly sampled $8,000$ text excerpts of $1,024$ tokens each, which sums up to 8M tokens in total, from the test split of the uncopyrighted Pile dataset, to evaluate the interpretability scores.

\subsection{Parameters of Evaluation Metrics}

Comprehensive parameter settings are shown in~\Cref{tab:eval-method-settings}. For the OpenAI metrics, each input text excerpt contains 64 tokens. For the \textit{OpenAI top-and-random} metric, we use 5 top activated excerpts for explanation, and a mixture of 5 top activated and 5 randomly activated excerpts for simulation. For the \textit{OpenAI random-only} metric, we only use 5 randomly activated excerpts for simulation. 

For the \textit{Anthropic} metric, each text excerpt contains only 8 tokens. For the explanation model, we input 15 top activated excerpts, 5 randomly activated excerpts, and 22 excerpts from different activation quantiles. To elaborate, we evenly divide the activation range into 11 quantiles, where we pick 2 excerpts from each of them. For the simulation model, we input 10 top activated excerpts, 5 randomly activated excerpts, 22 quantiled excerpts, and 10 top activated out-of-context (OOC) excerpts. Our implementation of the OOC excerpts is to cut the input text excerpt into length of only 3 tokens.

\begin{table}[H]
\def\arraystretch{1.1}
\small
\caption{\small Evaluation parameter settings of the OpenAI and Anthropic approach.}
\centering
\begin{tabular}{cc|c|ccc|cccc}
\hline
& && \multicolumn{3}{c|}{\textbf{Explanation}} & \multicolumn{4}{c}{\textbf{Simulation}} \\
&& \#Token & \#Top & \#Rand & \#Qua & \#Top & \#Rand & \#Qua & \#OOC \\
\hline
\multirow{2}{*}{\textbf{OpenAI}}& TaR & 64 & 5 &  &  & 5 & 5 &  &  \\
& Rand & 64 & 5 &  &  &  & 5 &  &  \\
\hline
\multicolumn{2}{c|}{\textbf{Anthropic}} & 8 & 15 & 5 & $2\cdot 11$ & 10 & 5 & $2\cdot 11$ & 10 \\
\hline
\end{tabular}
\label{tab:eval-method-settings}
\end{table}

\subsection{Discussion on Focus of Different Measures}

The OpenAI random-only metric is the easiest to interpret. As noted by~\citet{bills2023language}, the random-only metric considers an explanation’s ability to capture the neuron’s representation of features in the pre-training distribution, because the simulated tokens are uniformly randomly sampled from the validation set of the pre-train dataset. However, the random-only scoring with small sample size risks failing to capture behavior, due to lacking both tokens with high simulated activations and tokens with high real activations. Top-and-random scoring addresses the latter, but causes us to penalize falsely low simulations more than falsely high simulations, and thus tends to accept overly broad explanations.

The Anthropic metric, on the other hand, puts more focus on the mono-semanticity of the activations, as noted by~\citet{bricken2023towards}. For sparse features, which don't fire on most random samples, evaluating across a wide range of activations effectively tests the model's ability to distinguish a feature's large activations from zero, and the short text excerpts make it easier for the simulation model to identify the sparse activations.

\subsection{More Accessible Evaluation}

To reduce compute cost, we use \texttt{Mistral-7B-instruct} as the explanation model, and \texttt{LLaMA-2-7B} as the simulation model. We empirically prove that these replacements does not affect the conclusions of apple-to-apple comparison between \ours{} and GPT-2 below. 

\textbf{Explanation model.} In the official implementation~\citep{wu2023automated}, the explanation model is \texttt{gpt-4}. According to ablations described in~\citet{bills2023language}, it also makes sense to use the sligtly cheaper model \texttt{gpt-3.5-instruct}. Due to the high compute cost, we use the open-source model \texttt{mistral-7b-instruct} instead. We demonstrate the  performance of \texttt{gpt-3.5-turbo} and \texttt{mistral-7b-instruct} using the OpenAI random-only and top-and-random metrics in~\Cref{tab:abl-1}. Results show that the change of model doesn't significantly change the scores, and doesn't affect conclusions at all.

\begin{table}[!t]
\def\arraystretch{1.1}
    \small
    \caption{\small Interpretability measure of GPT-2, GPT2-SAE and \ours on the Pile dataset based on the OpenAI metrics. Explanation model: \texttt{Mistral-7B-instruct/GPT-3.5-turbo}. Simulation model: \texttt{LLaMA-2-7B}.}
    \centering
    \begin{tabular}{ccc|ccc|ccc}
    \hline
    \multicolumn{3}{c|}{\texttt{mistral-7b-instruct}} & \multicolumn{3}{c|}{\textbf{$\rho$ (Random-only)} ($\%, \uparrow$)} & \multicolumn{3}{c}{\textbf{$\rho$ (Top-and-Random)} ($\%, \uparrow$)} \\
    Model & Size & Loss & Layer 1 & Layer 2 & Layer 3 & Layer 1 & Layer 2 & Layer 3 \\
     \hline
     \ours{}-1L & 6.54M & 4.06 & \gcg{4.8} & - & - & \gch{3.9} & - & - \\
     \ours{}-2L & 6.64M & 3.55 & \gcg{8.0} & \gcg{5.8} & - & \gch{7.8} & \gch{8.3} & - \\
     \ours{}-3L & 6.74M & 3.46 & \gcg{9.0} & \gcg{9.4} & \gca{6.9} & \gch{9.6} & \gch{9.3} & \gch{8.4} \\
     \hline
     GPT-1L & 6.64M & 3.83 & \gcg{8.9} & - & - & \gch{8.8} & - & - \\
     GPT-2L & 6.83M & 3.23 & \gcg{2.3} & \gcg{1.6} & - & \gch{4.3} & \gch{4.1} & - \\
     GPT-3L & 7.03M & 3.11 & \gcg{3.1} & \gcg{-0.3} & \gcg{1.3} & \gch{7.3} & \gch{0.8} & \gch{2.6} \\
     \hline
     \multicolumn{3}{c|}{GPT-1L (16x SAE)} & \gcg{2.9} & - & - & \gch{5.4} & - & - \\
     \multicolumn{3}{c|}{GPT-2L (16x SAE)} & \gcg{3.5} & \gcg{1.8} & - & \gch{7.4} & \gch{4.2} & - \\
     \multicolumn{3}{c|}{GPT-3L (16x SAE)} & \gcg{3.2} & \gcg{2.3} & \gcg{1.1} & \gch{9.6} & \gch{5.0} & \gch{4.5} \\
     \hline
    \end{tabular}

    \vspace{1mm}

    \begin{tabular}{ccc|ccc|ccc}
    \hline
    \multicolumn{3}{c}{\texttt{GPT-3.5-turbo}} & \multicolumn{3}{c|}{\textbf{$\rho$ (Random-only)} ($\%, \uparrow$)} & \multicolumn{3}{c}{\textbf{$\rho$ (Top-and-Random)} ($\%, \uparrow$)} \\
    Model & Size & Loss & Layer 1 & Layer 2 & Layer 3 & Layer 1 & Layer 2 & Layer 3 \\
     \hline
     \ours{}-1L & 6.54M & 4.06 & \gcg{4.8} & - & - & \gch{3.9} & - & - \\
     \ours{}-2L & 6.64M & 3.55 & \gcg{8.2} & \gcg{6.0} & - & \gch{7.5} & \gch{8.0} & - \\
     \ours{}-3L & 6.74M & 3.46 & \gcg{9.1} & \gcg{9.2} & \gca{6.9} & \gch{9.5} & \gch{9.1} & \gch{8.3} \\
     \hline
     GPT-1L & 6.64M & 3.83 & \gcg{9.0} & - & - & \gch{9.0} & - & - \\
     GPT-2L & 6.83M & 3.23 & \gcg{2.2} & \gcg{1.6} & - & \gch{4.3} & \gch{4.4} & - \\
     GPT-3L & 7.03M & 3.11 & \gcg{3.0} & \gcg{-0.3} & \gcg{1.2} & \gch{7.0} & \gch{3.1} & \gch{3.0} \\
     \hline
     \multicolumn{3}{c|}{GPT-1L (16x SAE)} & \gcg{2.6} & - & - & \gch{4.7} & - & - \\
     \multicolumn{3}{c|}{GPT-2L (16x SAE)} & \gcg{3.4} & \gcg{1.6} & - & \gch{5.0} & \gch{2.9} & - \\
     \multicolumn{3}{c|}{GPT-3L (16x SAE)} & \gcg{2.8} & \gcg{1.8} & \gcg{1.2} & \gch{7.4} & \gch{3.8} & \gch{3.2} \\
     \hline
    \end{tabular}
    \label{tab:abl-1}
\end{table}

\textbf{Simulation model.} The official implementation of the simulation model utilizes \texttt{text-davinci-003} (now named \texttt{gpt-3.5-turbo-instruct}), which no longer supports retrieving the logprobs through the API, so we use \texttt{LLaMA-2-70B} as an equally capable replacement~\citep{touvron2023llama}. For more accessible evaluations, we use \texttt{LLaMA-2-7B} instead. We show the difference in interpretability caused by different simulation model size on \texttt{LLaMA-2-7B} and \texttt{LLaMA-2-70B} in~\Cref{tab:abl-2}. Empirical results show that although  \texttt{LLaMA-2-7B} has overall lower scores and higher variance than \texttt{LLaMA-2-70B}, it doesn't affect essential conclusions about the apple-to-apple comparison between \ours{} and GPT-2.

\begin{table}[!t]
\def\arraystretch{1.1}
    \small
    \caption{\small Interpretability measure of GPT-2 and \ours on the Pile dataset based on the OpenAI Top-and-random metric. Explanation model: \texttt{GPT-3.5-turbo}. Simulation model: \texttt{LLaMA-2-7B/LLaMA-2-70B}.
    \yaodong{seems like when using large models for interpretability, the standard GPT-2 model would become better but CRATE model gets worse?}
    \jackadv{I've added data for 6L. Seems like it's the case for $\ell=6$ but okay for other layers. There might be some tricks inside but we simply can't afford 70B.}
    }
    \centering
    \begin{tabular}{c|ccc|ccc}
    \hline
    & \multicolumn{3}{c|}{\textbf{Interpretability (7B)} ($\%, \uparrow$)} & \multicolumn{3}{c}{\textbf{Interpretability (70B)} ($\%, \uparrow$)} \\
    & Layer 1 & Layer 2 & Layer 3 & Layer 1 & Layer 2 & Layer 3\\
     \hline
     \ours{}-1L & \gca{3.9} & - & - & \gcb{6.4} & - & - \\
     \ours{}-2L & \gca{7.5} & \gca{8.0} & - & \gcb{7.4} & \gcb{7.1} &-\\
     \ours{}-3L & \gca{9.5} & \gca{9.1} & \gca{8.3} & \gcb{10.4} & \gcb{7.4} & \gcb{6.5} \\
     \hline
     GPT-1L & \gca{9.0} & - & - & \gcb{13.4} & - & - \\
     GPT-2L & \gca{4.3} & \gca{4.4} & - & \gcb{6.4} & \gcb{7.8}  & - \\
     GPT-3L & \gca{7.0} & \gca{3.1} & \gca{3.0} & \gcb{10.1} & \gcb{3.2} & \gcb{6.3} \\
     \hline
    \end{tabular}

    \vspace{2mm}

    \centering
    \begin{tabular}{c|c|cccccc}
    \hline
    && Layer 1 & Layer 2 & Layer 3 & Layer 4 & Layer 5 & Layer 6 \\
     \hline
     \multirow{2}{*}{\textbf{7B}} & \ours{}-6L & \gcdx{10.5} & \gcdx{7.4} & \gcdx{5.8} & \gcdx{8.0} & \gcdx{8.1} & \gcdx{5.7} \\
     & GPT-6L & \gcdx{13.7} & \gcdx{5.2} & \gcdx{1.9} & \gcdx{0.6} & \gcdx{5.6} & \gcdx{6.9} \\
     \hline
     \multirow{2}{*}{\textbf{70B}} & \ours{}-6L & \gcdx{10.1} & \gcdx{6.5} & \gcdx{7.0} & \gcdx{8.3} & \gcdx{9.4} & \gcdx{0.7} \\
      & GPT-6L & \gcdx{14.5} & \gcdx{6.2} & \gcdx{2.5} & \gcdx{0.7} & \gcdx{3.9} & \gcdx{4.1} \\
    \hline
    \end{tabular}
    \label{tab:abl-2}
\end{table}

\section{Analysis on Activation Sparsity} \label{app:analysis-sparsity}

\begin{figure}[H]
     \centering
     \begin{subfigure}[b]{0.48\textwidth}
         \centering
    \includegraphics[width=\textwidth]{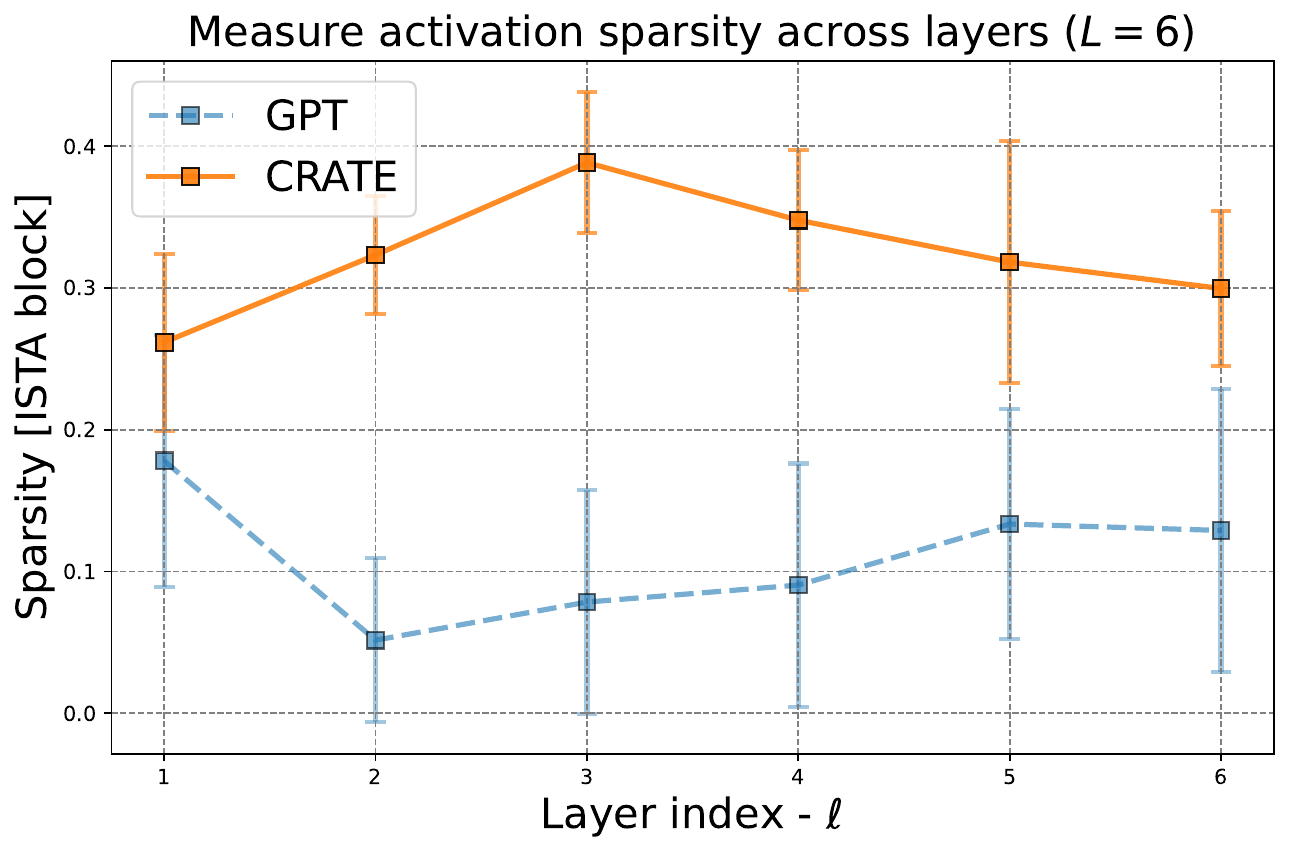}
     \end{subfigure}
     \begin{subfigure}[b]{0.48\textwidth}
         \centering
    \includegraphics[width=\textwidth]{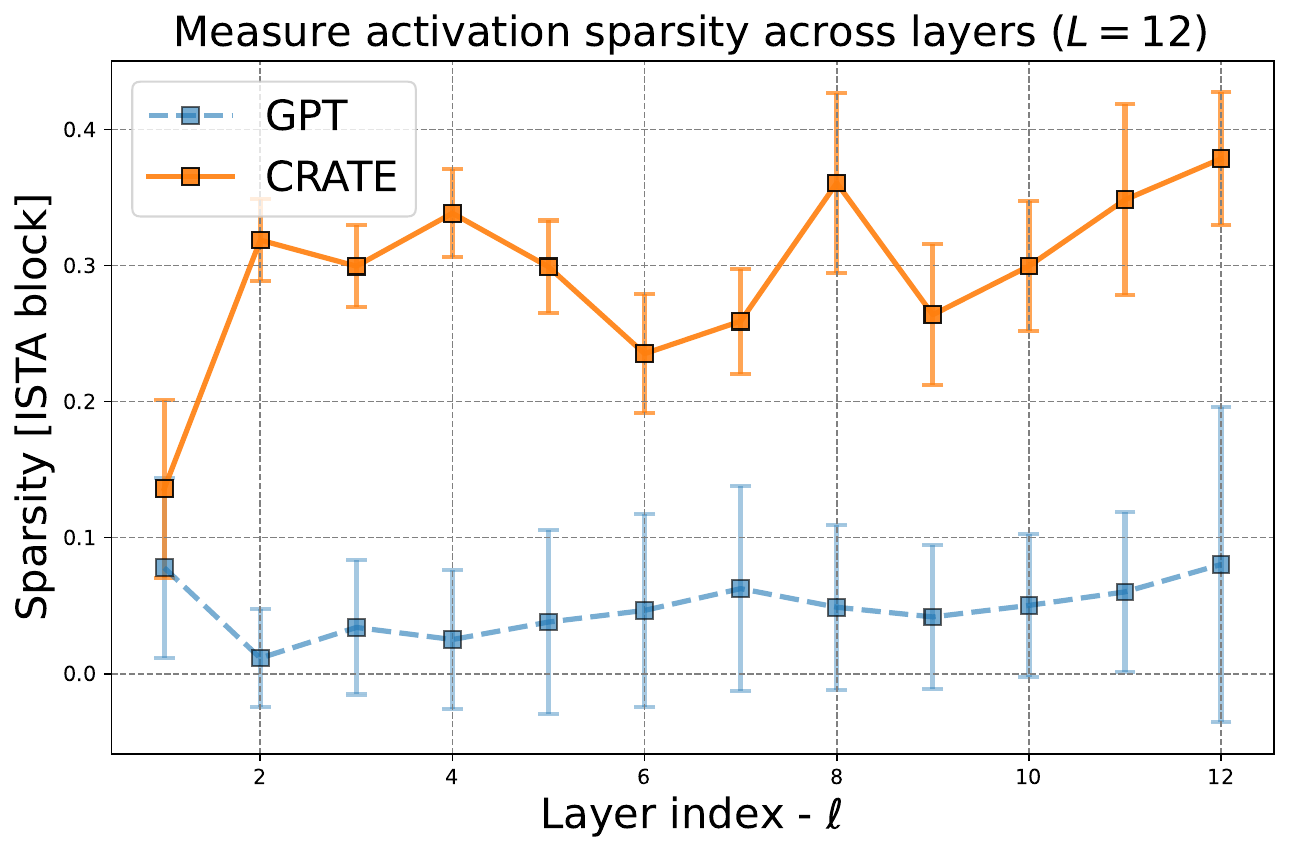}
     \end{subfigure}
        \caption{Layer-wise activation sparsity of \ours{} and GPT. \textit{Left:} 6L models. \textit{Right:} 12L models.}
        \label{fig:crate-gpt-sparsity}
\end{figure}

\begin{figure}[H]
     \centering
     \begin{subfigure}[b]{0.50\textwidth}
         \centering
    \includegraphics[width=\textwidth]{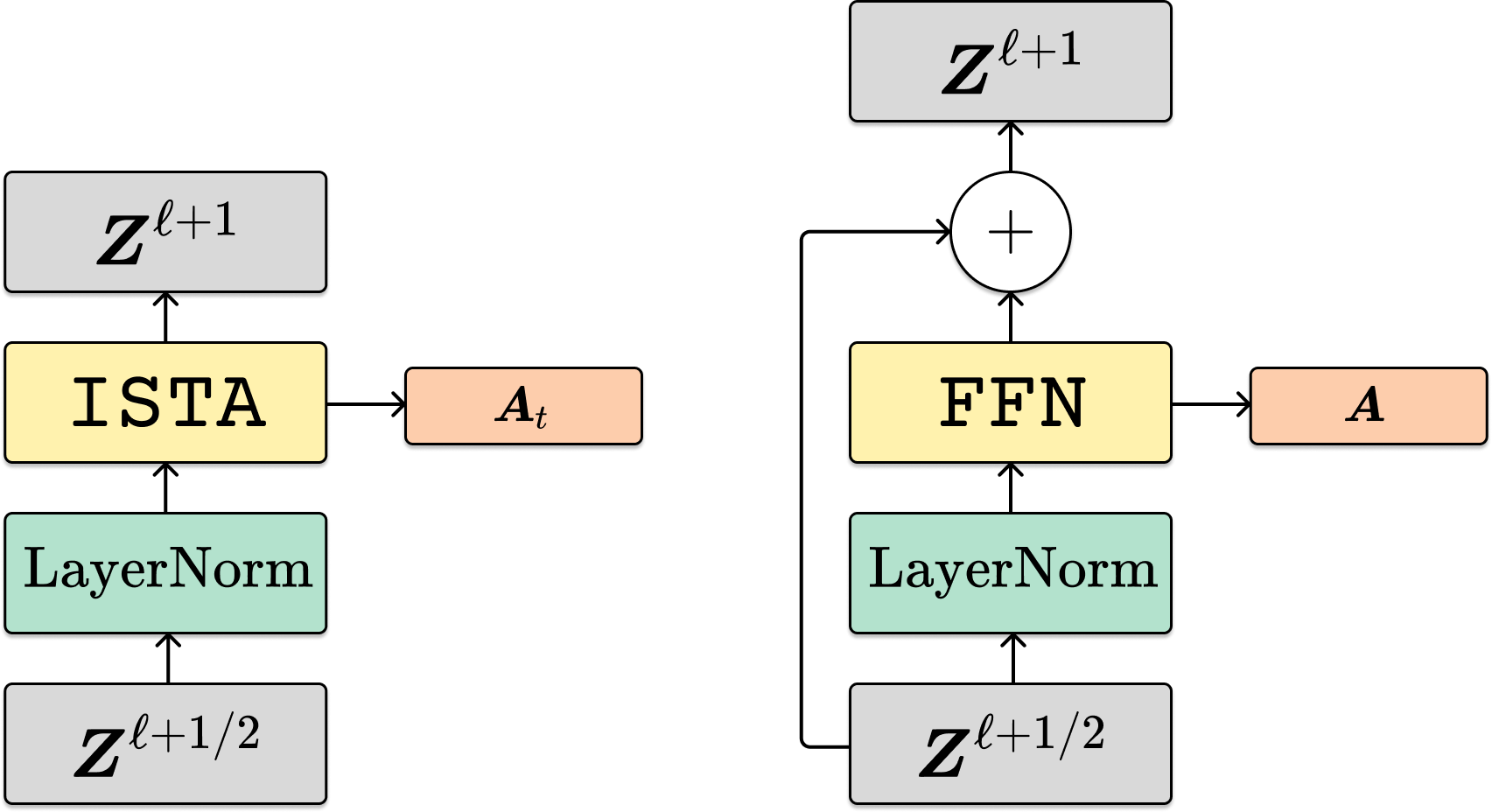}
     \end{subfigure}
        \caption{Extracting sparse code $\vA_t$ from \ours{} and hidden layer output $\vA$ from GPT-2.}
        \label{fig:extract-act}
\end{figure}

We demonstrate the activation sparsity of \ours{} compared to GPT-2 in~\Cref{fig:crate-gpt-sparsity}. One might have the confusion about why \ours{} is designed to be sparse but the activations evaluated is denser than GPT-2. Note that the sparsity evaluated in standard transformer model is output from the hidden layer of the MLP layer, which is the activation matrix $\vA$ \textit{before applying to the residual stream}, as shown in~\Cref{fig:extract-act}. The actual representations in standard transformers, which are after applying the residual stream, are not sparse at all. In contrast, the sparsity evaluated in \ours{} is the actual representations $\vA_t$ (\textit{including the residual stream}).

\begin{figure}[H]
     \centering
     \begin{subfigure}[b]{0.48\textwidth}
         \centering
    \includegraphics[width=\textwidth]{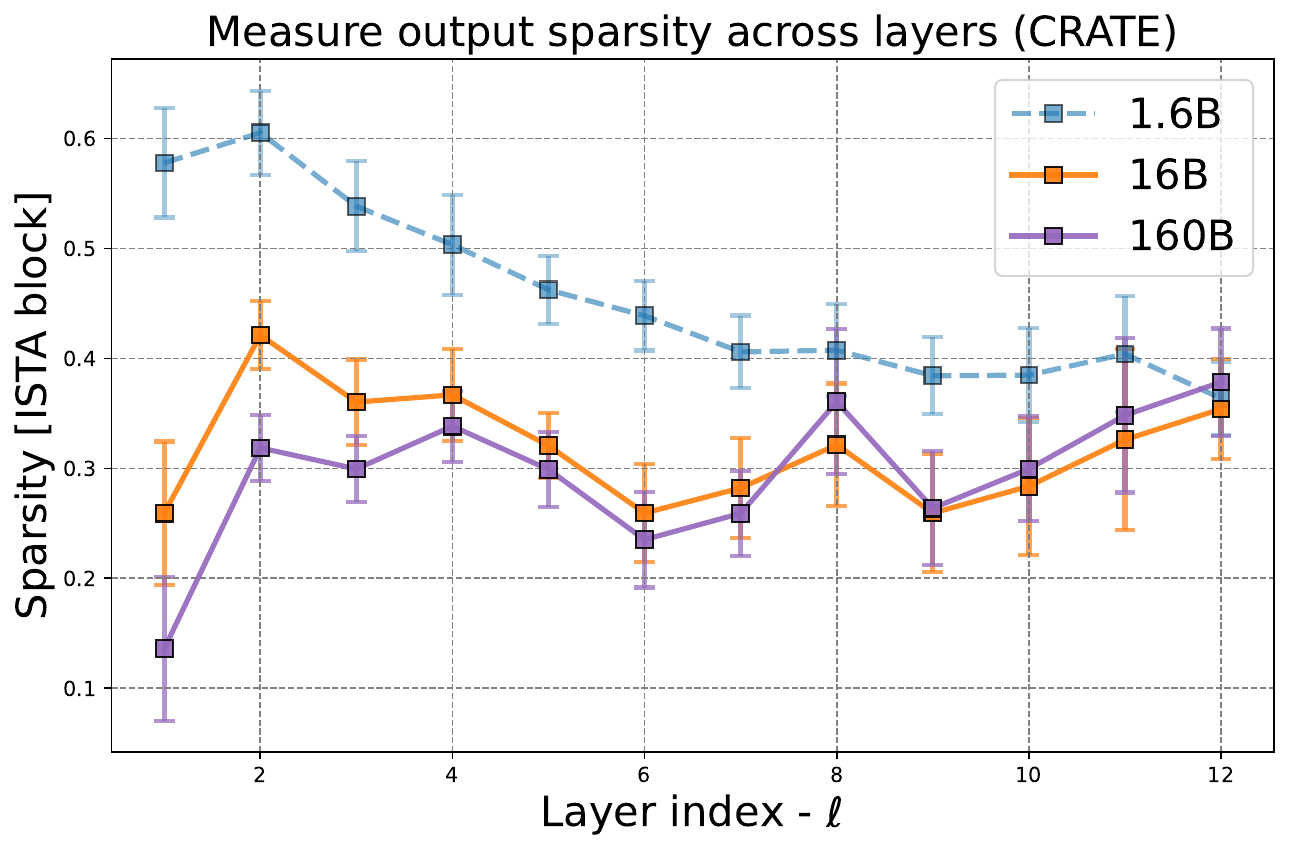}
     \end{subfigure}
     \begin{subfigure}[b]{0.48\textwidth}
         \centering
    \includegraphics[width=\textwidth]{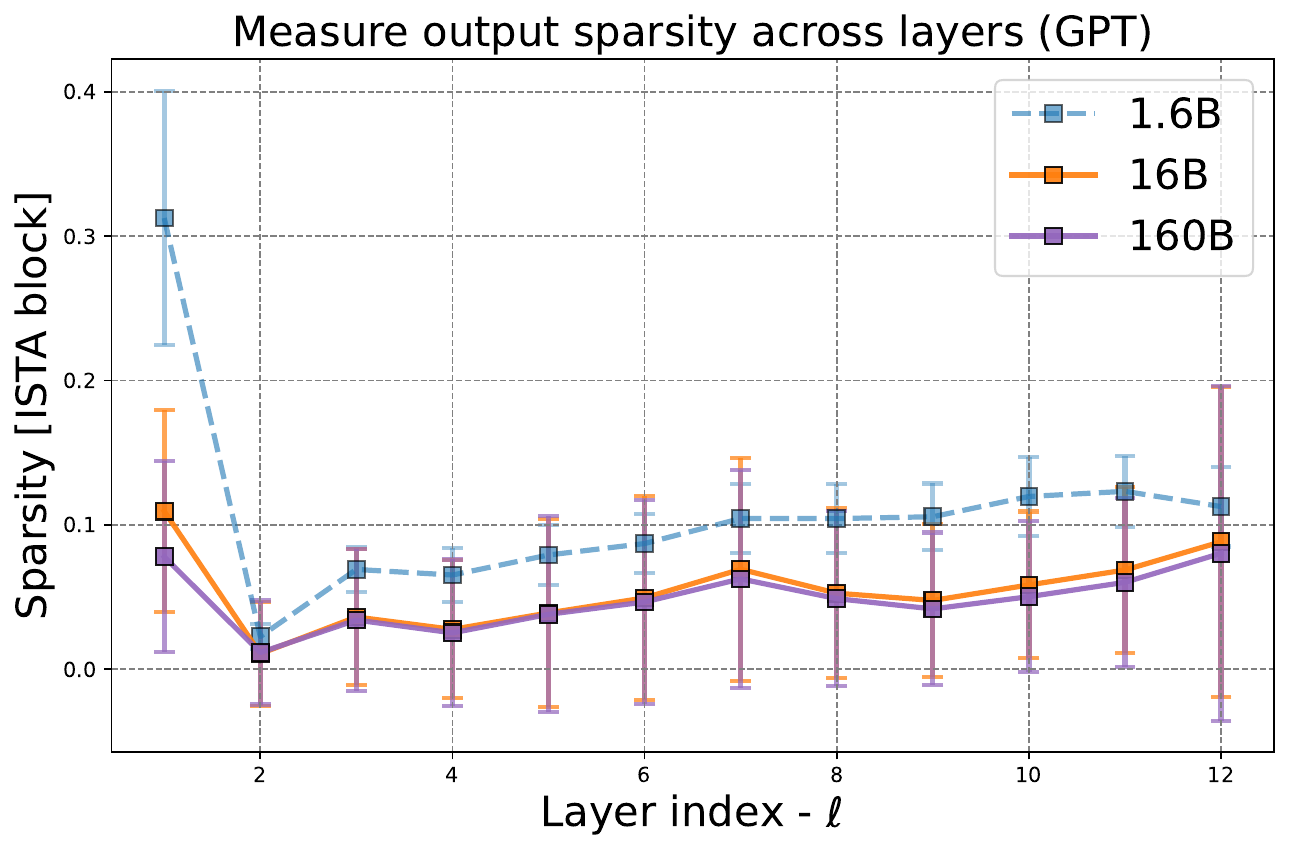}
     \end{subfigure}
        \caption{Layer-wise activation sparsity w.r.t. tokens trained. \textit{Left:} \ours{} language model. \textit{Right:} GPT-2.}
        \label{fig:sparsity-wrt-tokens}
\end{figure}

We also present the activation dynamics of \ours{} and GPT-2 with the progression of the pre-training process in~\Cref{fig:sparsity-wrt-tokens}. We observe a strong trend that the sparsity of \ours{} monotonically decreases in the early stage (trained on 1.6B tokens), which aligns with the design purpose. In the late stage (16B, 160B tokens), the sparsities in the early sites ($L<12$) significantly decreases, which also aligns with the design purpose. On the other hand, GPT-2 never appears to have a decreasing trend of activation sparsity over layers across the whole pre-training stage, indicating a systematic difference between the sparsity dynamics between \ours{} and GPT-2. One counter-intuitive observation is that the decreasing trend fades as the stage moves on. Our hypothesis is that \ours{} overfits on the next token prediction task due to the large amount of tokens trained.

\section{Details on Interpretation Score Distributions} \label{app:interp-distribution}

\begin{figure}[!t]
    \centering
    \centering
     \begin{subfigure}[b]{\textwidth}
         \centering
    \includegraphics[width=\textwidth]{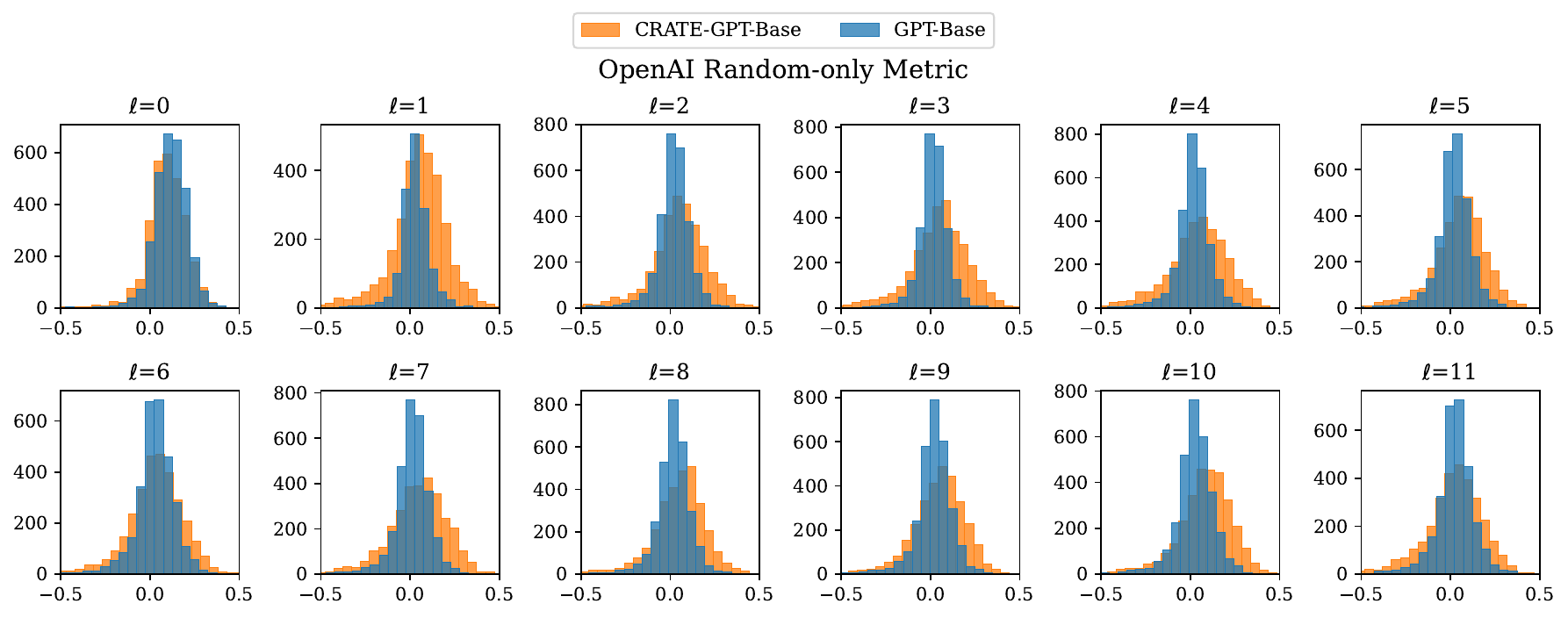}
     \end{subfigure}
     \begin{subfigure}[b]{\textwidth}
         \centering
    \includegraphics[width=\textwidth]{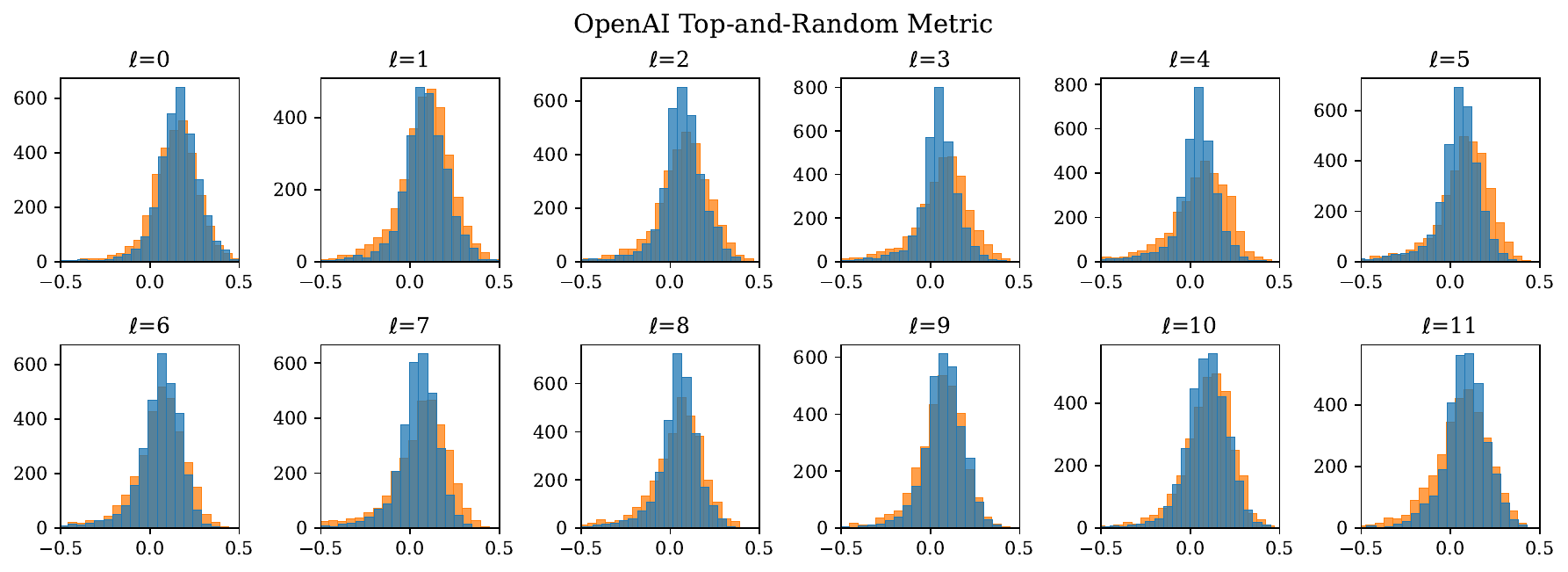}
     \end{subfigure}
      \begin{subfigure}[b]{\textwidth}
         \centering
    \includegraphics[width=\textwidth]{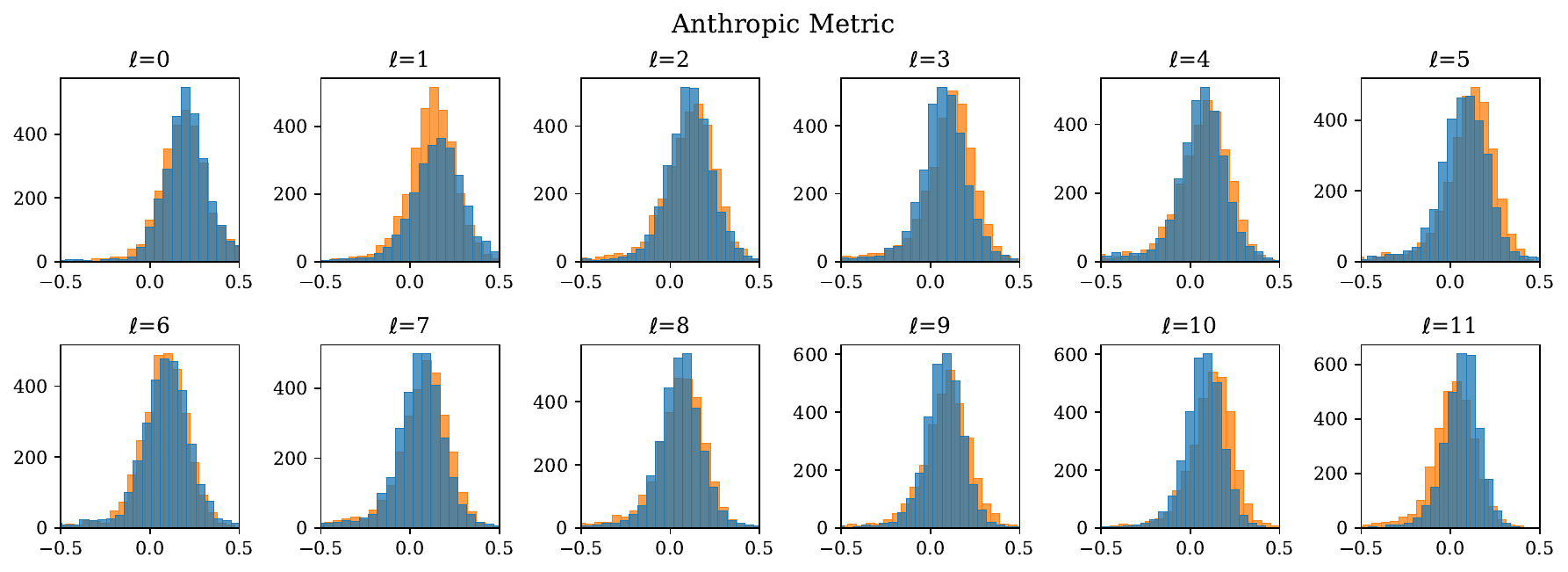}
     \end{subfigure}
        \caption{Distribution of the interpretation scores over \ours{}-12L and GPT2-12L. $x$-axis: interpretation score. $y$-axis: count of neurons falling in the corresponding interval of interpretation score.}
        \label{fig:interp-distr}
\end{figure}

We visualize the distributions of layer-wise interpretation scores of \ours{} and GPT-2 with $L=12$ in~\Cref{fig:interp-distr}. We exclude cases where activations sampled from GPT2-Base (random-only metric) are all zeroes, as in these cases the correlation $\rho$ will be undefined. This results in a smaller number counted in the GPT-2 activations in the first two rows.


\yaodong{why in the first few rows the overall number of samples in CRATE is larger than the one of standard transformer? this seems to be inconsistent. } \jackadv{For cases where activations sampled from GPT2-Base (random-only metric) are all zeroes, we exclude them as the correlation $\rho$ will be undefined. This results in a smaller number of activations counted in the first two rows. (Updated in text)}

\section{Details on Sparse Auto-encoder} \label{app:sae}

\subsection{Sparse Autoencoder and Dictionary Learning}

The dictionary learning model is an MLP model with a single hidden layer. It is trained as an auto-encoder using input weights as the encoder (which maps the input activations to a higher dimension), and output weights as the decoder. 
Formally, given activation $\va\in \R^{h}$ sampled from $\vA\in\R^{h\times N}$, the encoder $\vW_1, \vb_1$ with dimension multiplicator $\mu$ maps the activations to a hidden representation $\vh \in\R^{\mu h}$, whereas the decoder $\vW_2, \vb_2$ maps the representation back to the original dimension $\hat{\va}\in\R^{h}$. The dictionary learning objective can thus be expressed as

\vspace{-4mm}
\begin{align}
    \bar \va&=\va-\vb_2 \\
    \vh&=\text{ReLU}(\vW_1\bar x+\vb_1)\\
    \hat \va&=\vW_2 \vh+\vb_2\\
    \mathcal L&=\frac{1}{|\vA|}\sum_{\va\in\vA} \|\va-\hat \va\|_2^2+\lambda \|\vh\|_1
\end{align}

\subsection{Detailed Setup} \label{app:sae-setup}

We train the sparse auto-encoders on the train split of the uncopyrighted Pile dataset until convergence. Following~\citet{bricken2023towards} and~\citet{arthur2023saearticle}, we adopt the resampling strategy to re-train the dead features, and the learning rate scheduling strategy to improve recovery rate. For implementation, we mainly follow~\citet{arthur2024saerepo}, with $\lambda_{\ell_1}=1.6\times 10^{-4}, \alpha=1.2\times 10^{-3}$ for all sizes of models. We evaluate using the average loss of randomly sampled batches on the validation split of the uncopyrighted Pile dataset.

\subsection{Loss Curve} \label{app:sae-lr-curve}

\begin{figure}[H]
     \centering
     \begin{subfigure}[b]{0.48\textwidth}
         \centering
    \includegraphics[width=\textwidth]{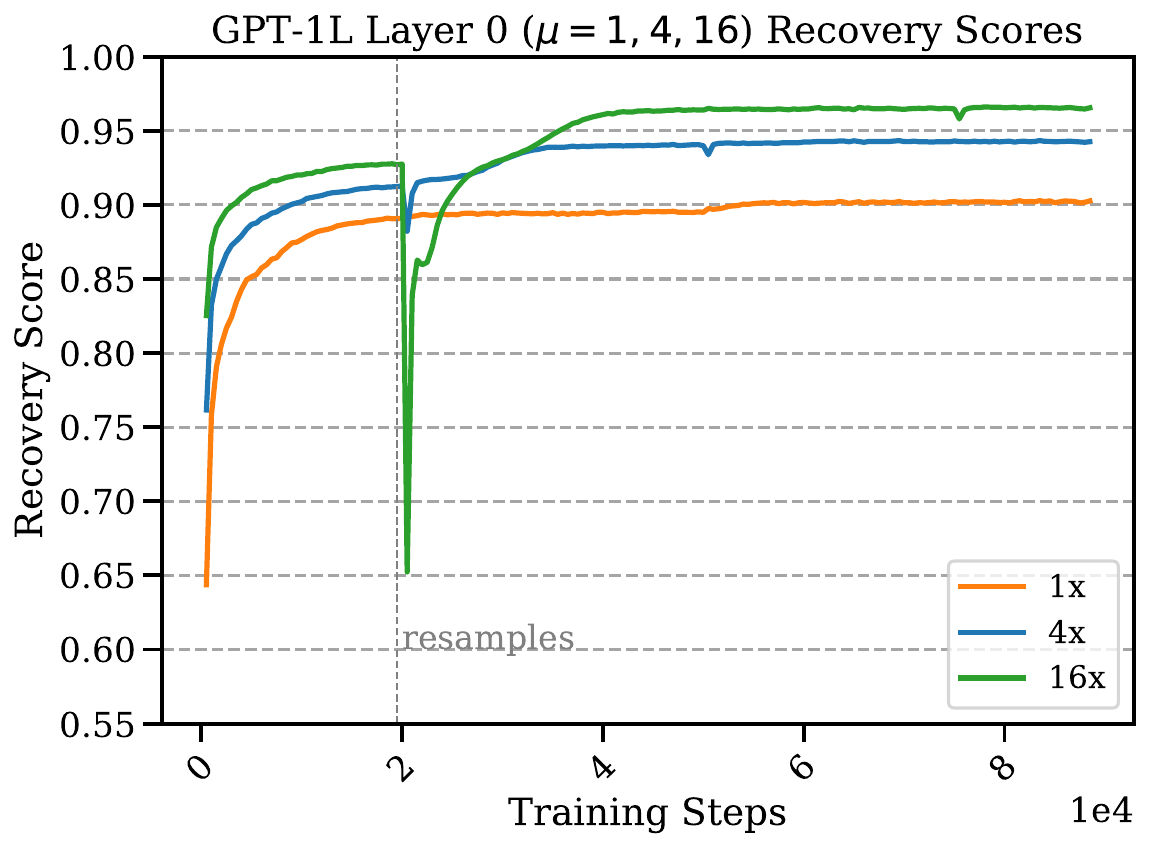}
     \end{subfigure}
     \begin{subfigure}[b]{0.48\textwidth}
         \centering
    \includegraphics[width=\textwidth]{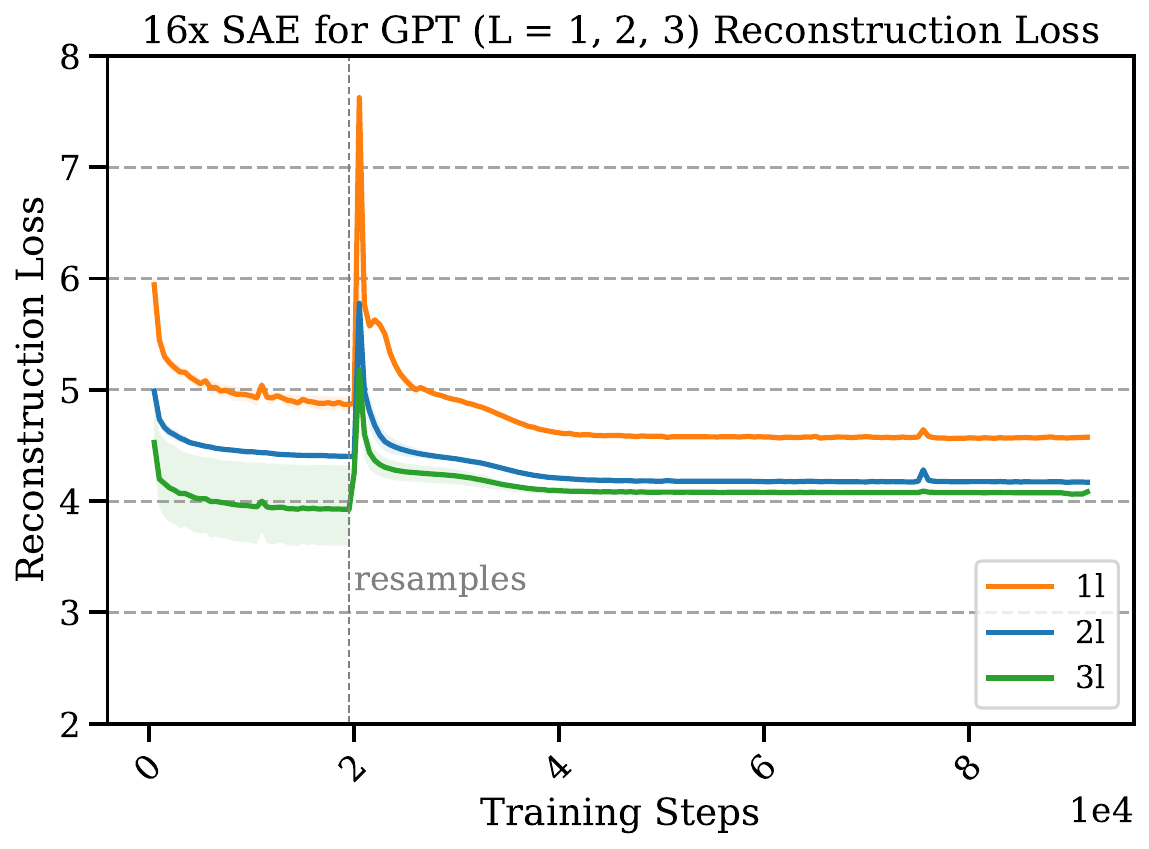}
     \end{subfigure}
        \caption{\textit{Left:} The recovery scores of GPT-1L ($\ell=0$) with SAE multiplication factors $\mu=\in\{1,4,16\}$. 
        \textit{Right:} The reconstruction loss of SAE with $\mu=16$ on different sizes of GPT-2 models $L\in\{1,2,3\}$, averaged across all layers.}
        \label{fig:gpt-sae-curves}
\end{figure}

The loss curves of training the sparse auto-encoders are shown in~\Cref{fig:gpt-sae-curves}. Generally, resampling boosts the performance of the reconvery score, which aligns with the conclusions shown in~\citet{bricken2023towards} and~\cite{arthur2023saearticle}. We also observe an increasing trend of performance with the increases of the SAE multiplication factor $\mu$ and model size $L$.

\subsection{Performance} \label{app:sae-performance}

The performance of sparse auto-encoders of \ours{} and GPT-2 under a variety of settings (model $L, \ell$ and sparse autoencoder width multiplication factor $\mu$) are shown in~\Cref{tab:crate-gpt-sae-loss-1x}. The percentages of dead neurons for all layers of $L\in\{1,2,3\}$ are less than 1\%.

\begin{table}[H]
\def\arraystretch{1.1}
    \small
    \caption{\small Reconstruction loss and recovery score of the sparse autoencoders on \ours{} and GPT-2.
    }
    \centering
    \begin{tabular}{ccc|ccc|ccc}
    \hline
    \multicolumn{3}{c|}{$\mu=16$} & \multicolumn{3}{c|}{\textbf{Reconstruction Loss} ($\downarrow$)} & \multicolumn{3}{c}{\textbf{Recovery Score} ($\%, \uparrow$)} \\
    & Size & Loss & Layer 0 & Layer 1 & Layer 2 & Layer 0 & Layer 1 & Layer 2\\
     \hline
     GPT-1L & 6.64M & \gcl{3.83} & \gcl{4.35} & - & - & 95.0 & - & - \\
     GPT-2L & 6.83M & \gcl{3.23} & \gcl{3.50} & \gcl{3.45} & - & 95.2 & 92.2 & - \\
     GPT-3L & 7.03M & \gcl{3.11} & \gcl{3.38} & \gcl{3.39} & \gcl{3.29} & 94.6 & 94.8 & 92.4 \\
     \hline
     \ours{}-1L & 6.54M & \gcl{4.06} & \gcl{4.33} & - & - & 93.6 & - & - \\
     \ours{}-2L & 6.64M & \gcl{3.55} & \gcl{4.12} & \gcl{3.80} & - & 95.7 & 95.7 & - \\
     \ours{}-3L & 6.74M & \gcl{3.46} & \gcl{4.05} & \gcl{3.99} & \gcl{3.77} & 93.0 & 93.9 & 95.0 \\
     \hline
    \end{tabular}

    \vspace{1mm}

    \begin{tabular}{ccc|ccc|ccc}
    \hline
    \multicolumn{3}{c|}{$\mu=4$} & \multicolumn{3}{c|}{\textbf{Reconstruction Loss} ($\downarrow$)} & \multicolumn{3}{c}{\textbf{Recovery Score} ($\%, \uparrow$)} \\
    & Size & Loss & Layer 0 & Layer 1 & Layer 2 & Layer 0 & Layer 1 & Layer 2\\
     \hline
     GPT-1L & 6.64M & \gcl{3.83} & \gcl{4.34} & - & - & 93.7 & - & - \\
     GPT-2L & 6.83M & \gcl{3.23} & \gcl{3.59} & \gcl{3.56} & - & 92.7 & 88.6 & - \\
     GPT-3L & 7.03M & \gcl{3.11} & \gcl{3.45} & \gcl{3.50} & \gcl{3.34} & 92.2 & 94.9 & 89.9 \\
     \hline
     \ours{}-1L & 6.54M & \gcl{4.06} & \gcl{4.39} & - & - & 92.1 & - & - \\
     \ours{}-2L & 6.64M & \gcl{3.55} & \gcl{4.37} & \gcl{3.93} & - & 93.7 & 93.8 & - \\
     \ours{}-3L & 6.74M & \gcl{3.46} & \gcl{4.03} & \gcl{4.11} & \gcl{3.81} & 92.6 & 92.6 & 92.6 \\
     \hline
    \end{tabular}

    \vspace{1mm}

    \begin{tabular}{ccc|ccc|ccc}
    \hline
    \multicolumn{3}{c|}{$\mu=1$} & \multicolumn{3}{c|}{\textbf{Reconstruction Loss} ($\downarrow$)} & \multicolumn{3}{c}{\textbf{Recovery Score} ($\%, \uparrow$)} \\
    & Size & Loss & Layer 0 & Layer 1 & Layer 2 & Layer 0 & Layer 1 & Layer 2\\
     \hline
     GPT-1L & 6.64M & \gcl{3.83} & \gcl{4.93} & - & - & 95.0 & - & - \\
     GPT-2L & 6.83M & \gcl{3.23} & \gcl{3.89} & \gcl{3.75} & - & 89.0 & 82.2 & - \\
     GPT-3L & 7.03M & \gcl{3.11} & \gcl{3.63} & \gcl{3.61} & \gcl{3.58} & 86.9 & 91.0 & 84.9 \\
     \hline
     \ours{}-1L & 6.54M & \gcl{4.06} & \gcl{4.69} & - & - & 86.2 & - & - \\
     \ours{}-2L & 6.64M & \gcl{3.55} & \gcl{4.68} & \gcl{4.29} & - & 90.4 & 88.9 & - \\
     \ours{}-3L & 6.74M & \gcl{3.46} & \gcl{4.39} & \gcl{4.38} & \gcl{4.16} & 97.0 & 89.2 & 88.1 \\
     \hline
    \end{tabular}
    \label{tab:crate-gpt-sae-loss-1x}
\end{table}

\subsection{Interpretability} \label{app:interp-crate-sae}

In~\Cref{sec:ablations}, we've demonstrated that \ours{} has more optimal representation than GPT-2 in terms of interpretability. Here we show the details of this experiment and layer-wise decomposition of the interpretability comparison shown in~\Cref{tab:interp-comparison-gpt-crate-sae}.

As it's hard to decide how much interpretability gain it is from \ours{} to \ours{}-SAE directly (as explained in~\Cref{sec:interpretability}), we compare the interpretability \textit{improvement} of \ours{}-SAE over \ours{} to the interpretability improvement of GPT2-SAE over GPT-2. The interpretability of GPT2-SAE is already included in~\Cref{fig:gpt-crate-interp-comp}. 
The interpretability of \ours{}-SAE under the OpenAI TaR and Anthropic metrics are shown in~\Cref{tab:interp-crate-sae}.

\begin{table}[h]
    \caption{Interpretability of \ours{}-SAE under the OpenAI TaR and Anthropic metics.}
    \centering
    \begin{tabular}{@{}lccc|ccc@{}}
        \toprule
        & \multicolumn{3}{c|}{\textbf{OpenAI TaR} ($\uparrow$)} & \multicolumn{3}{c}{\textbf{Anthropic} ($\uparrow$)} \\
        & Layer 1 & Layer 2 & Layer 3 & Layer 1 & Layer 2 & Layer 3 \\ \midrule
        1L & $6.0$ & - & - & $17.9$ & - & - \\
        2L & $7.7$ & $5.2$ & - & $21.7$ & $12.4$ & - \\
        3L & $7.2$ & $6.4$ & $4.6$ & $19.3$ & $18.4$ & $11.6$ \\ \bottomrule
    \end{tabular}
    \label{tab:interp-crate-sae}
\end{table}

\subsection{Steering the LM or SAE} \label{sec:steer-lm-or-sae}

\begin{figure}[H]
     \centering
     \begin{subfigure}[b]{0.95\textwidth}
         \centering
    \includegraphics[width=\textwidth]{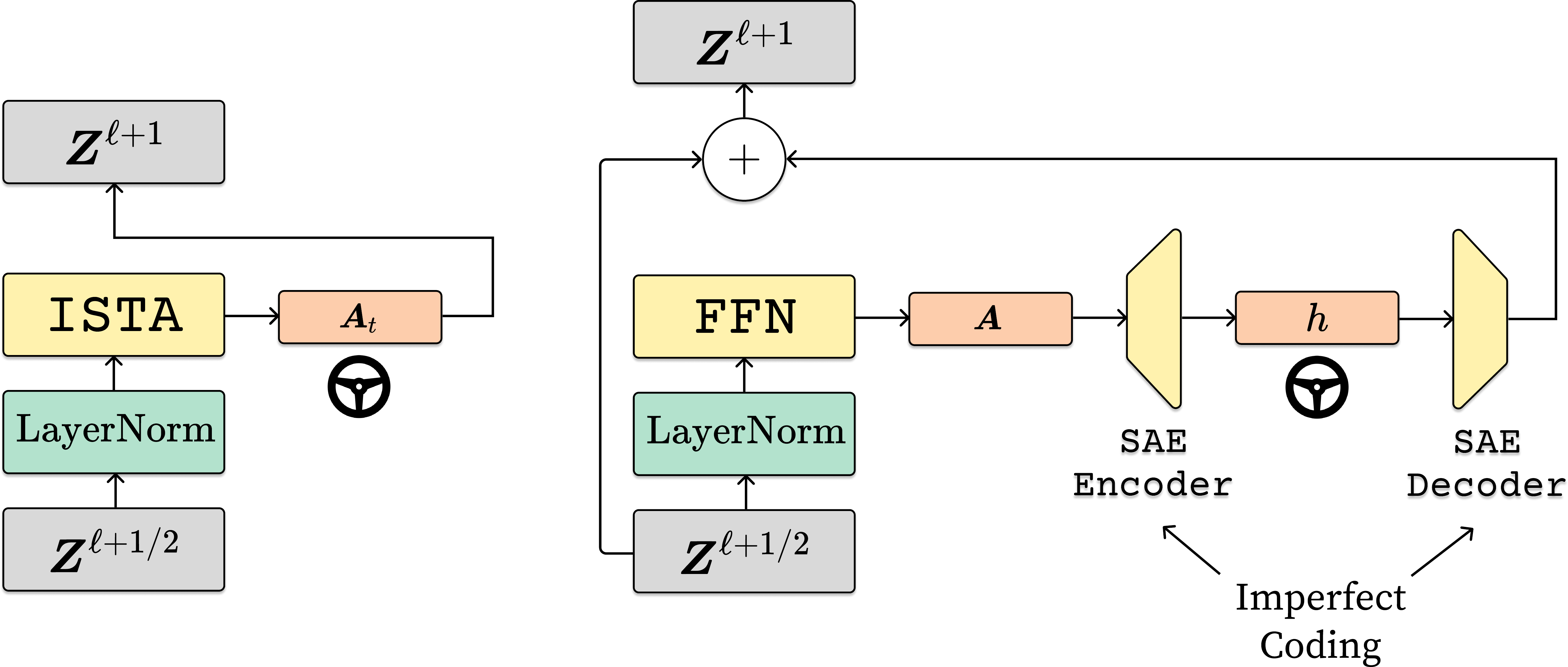}
     \end{subfigure}
        \caption{Illustration of steering a language model directly or using SAE.}
        \label{fig:steering}
\end{figure}

In comparison to post-hoc trained SAEs, built-in sparsification processes, such as the one we proposed in this paper, have the potential to be steered with perfect fidelity. As visualized in~\Cref{fig:steering}, post-hoc approaches like SAE require steering the model with the decomposed hidden states $h$, whose encoding and decoding processes are both \textit{lossy}. An imperfect reconstruction systematically leads to \textit{distortions} of the steering signal upon the hidden states, and thus affects downstream applications of the GPT2-SAE model. In contrast, \ours{} doesn't include any approximation that distorts the steering signal, so the signal can be propagated without loss of fidelity. This conclusion does not change whether the performance of GPT2-SAE outperforms \ours{} or not.

\section{Further Qualitative Results on Interpretable Neurons} \label{app:qual-neuron-interps}

\setlength{\fboxsep}{0pt}
\setlength{\fboxrule}{0.5pt}

This section lists some further qualitative examples of tokens and their activations when $L=3$, including examples of the GPT2-SAE activations. Specifically, we demonstrate two neurons in each model. Tokens with a deeper \ct{b5b8ff}{blue} background have a higher activation. Explanations $k_i$ and scores $s_i$ are obtained by~\Cref{algo:interp-scoring}.

\subsection{\ourscaps{}-3L, Layer 0, Neuron 288}

\paragraph{OpenAI Evaluation} \textbf{Score:} 0.44478400135318646\\\textbf{Explanation:}  information related to the regulation of mRNA expression and its role in carbohydrate metabolism, with a focus on CRC cells and gene signaling in the context of cancer development.\\ \textbf{Top Activations}\\\fbox{\begin{minipage}{\textwidth}\ct{f6f6ff}{ animal} \ct{eaebff}{ models} \ct{eaebff}{ of} \ct{dadbff}{ cart} \ct{dfe1ff}{ilage} \ct{cfd1ff}{ degradation} \ct{f8f8ff}{ ([} \ct{ffffff}{@} \ct{fcfcff}{b} \ct{f9faff}{41} \ct{dfe0ff}{-} \ct{ffffff}{mm} \ct{fdfdff}{r} \ct{e4e5ff}{-} \ct{ffffff}{16} \ct{e9eaff}{-} \ct{ffffff}{04} \ct{ededff}{-} \ct{fbfbff}{38} \ct{fefeff}{41} \ct{dadcff}{]).} \ct{f6f7ff}{ Among} \ct{e6e7ff}{ these} \ct{acafff}{ cytok} \ct{c8caff}{ines} \ct{e1e2ff}{,} \ct{b6b9ff}{ IL} \ct{babdff}{-} \ct{ced0ff}{1} \ct{acafff}{$\beta$} \ct{bdc0ff}{ is} \ct{9fa3ff}{ highly} \ct{8085ff}{ overe} \ct{a2a5ff}{xp} \ct{9599ff}{ressed} \ct{bfc2ff}{ in} \ct{c1c3ff}{ the} \ct{bdc0ff}{ cart} \ct{ccceff}{ilage} \ct{e1e2ff}{ and} \ct{d9dbff}{ in} \ct{d6d8ff}{ the} \ct{dcdeff}{ syn} \ct{f4f4ff}{ov} \ct{dddeff}{ial} \ct{c6c9ff}{ tissue} \ct{eaebff}{,} \ct{ffffff}{ while} \ct{eeeeff}{ the} \ct{c4c7ff}{ expression} \ct{dedfff}{ of} \ct{c1c3ff}{ IL} \ct{c1c3ff}{-} \ct{d6d7ff}{1} \ct{acafff}{ receptor} \ct{c0c3ff}{ antagonist} \ct{e7e8ff}{ ([} \ct{ffffff}{@} \ct{f2f3ff}{b} \ct{eaebff}{42} \ct{d1d3ff}{-} \ct{ffffff}{mm} \ct{f5f5ff}{r} \ct{d8d9ff}{-} \end{minipage}}\\ \fbox{\begin{minipage}{\textwidth}\ct{dedfff}{raft} \ct{e3e4ff}{s} \ct{ffffff}{.} \ct{e7e8ff}{ Moreover} \ct{f5f6ff}{,} \ct{eeefff}{ AC} \ct{dddeff}{OX} \ct{d5d7ff}{1} \ct{8a8fff}{ overe} \ct{aeb1ff}{xp} \ct{a5a9ff}{ression} \ct{c0c3ff}{ atten} \ct{c3c5ff}{uated} \ct{c5c7ff}{ the} \ct{e5e6ff}{ aug} \ct{d9daff}{mentation} \ct{cbcdff}{ of} \ct{c3c5ff}{ migration} \ct{d6d7ff}{ and} \ct{cacdff}{ invasion} \ct{cfd1ff}{ of} \ct{c5c7ff}{ CRC} \ct{aeb1ff}{ cells} \ct{d7d9ff}{ by} \ct{b1b4ff}{ mi} \ct{d2d4ff}{R} \ct{bbbeff}{-} \ct{e9eaff}{15} \ct{e6e7ff}{b} \ct{caccff}{-} \ct{dedfff}{5} \ct{dbddff}{p} \ct{8085ff}{ overe} \ct{aaadff}{xp} \ct{a0a4ff}{ression} \ct{e4e5ff}{.} \ct{e5e6ff}{ In} \ct{dadbff}{ conclusion} \ct{eff0ff}{,} \ct{e4e5ff}{ our} \ct{c5c8ff}{ study} \ct{aeb1ff}{ demonstrated} \ct{dbdcff}{ a} \ct{bfc2ff}{ functional} \ct{a5a9ff}{ role} \ct{c3c6ff}{ of} \ct{c8caff}{ the} \ct{dcddff}{ S} \ct{cfd1ff}{IRT} \ct{cfd1ff}{1} \ct{f7f7ff}{/} \ct{d5d7ff}{mi} \ct{e6e7ff}{R} \ct{c9cbff}{-} \ct{f4f5ff}{15} \ct{f0f0ff}{b} \ct{d3d5ff}{-} \ct{e5e6ff}{5} \ct{e2e3ff}{p} \ct{fafaff}{/} \ct{e7e8ff}{AC} \ct{e3e4ff}{OX} \ct{e2e4ff}{1} \ct{e3e4ff}{ axis} \end{minipage}}\\ \textbf{Random Activations}\\\fbox{\begin{minipage}{\textwidth}\ct{868bff}{ed} \ct{ecedff}{ with} \ct{ffffff}{ Peter} \ct{ffffff}{ Braun} \ct{ffffff}{,} \ct{ffffff}{ the} \ct{ffffff}{ Mor} \ct{ffffff}{av} \ct{ffffff}{ian} \ct{ffffff}{\textbackslash{}n} \ct{ffffff}{mission} \ct{b7baff}{ary} \ct{fcfcff}{ in} \ct{d7d8ff}{ Ant} \ct{f5f6ff}{ig} \ct{fbfbff}{ua} \ct{ffffff}{;} \ct{ffffff}{ and} \ct{ffffff}{ to} \ct{fefeff}{ that} \ct{ffffff}{ correspondence} \ct{b6b9ff}{ he} \ct{ffffff}{ owed} \ct{ffffff}{ in} \ct{ffffff}{ part} \ct{ffffff}{ his} \ct{ffffff}{\textbackslash{}n} \ct{ffffff}{interest} \ct{ffffff}{ in} \ct{ffffff}{ missionary} \ct{ffffff}{ work} \ct{ffffff}{.} \ct{ffffff}{ But} \ct{ffffff}{ that} \ct{ffffff}{ was} \ct{ffffff}{ not} \ct{ffffff}{ the} \ct{ffffff}{ end} \ct{ffffff}{ of} \ct{f6f7ff}{ the} \ct{ffffff}{ Bre} \ct{ffffff}{thren} \ct{ffffff}{'s} \ct{ffffff}{\textbackslash{}n} \ct{ffffff}{inf} \ct{ffffff}{luence} \ct{ffffff}{.} \ct{ffffff}{ At} \ct{ffffff}{ all} \ct{ffffff}{ meetings} \ct{ffffff}{ addressed} \ct{ffffff}{ by} \ct{ffffff}{ the} \ct{ffffff}{ founders} \ct{ffffff}{ of} \ct{ffffff}{ the} \ct{ffffff}{ proposed} \ct{ffffff}{\textbackslash{}n} \ct{ffffff}{Soc} \ct{ffffff}{iety} \ct{ffffff}{,} \ct{ffffff}{ the} \ct{ffffff}{ speaker} \ct{ffffff}{ repeatedly} \end{minipage}}\\ \fbox{\begin{minipage}{\textwidth}\ct{ffffff}{ESA} \ct{ffffff}{)} \ct{ffffff}{ and} \ct{ffffff}{ the} \ct{ffffff}{ American} \ct{ffffff}{ Association} \ct{ffffff}{ for} \ct{ffffff}{ the} \ct{ffffff}{ Advance} \ct{ffffff}{ment} \ct{ffffff}{ of} \ct{ffffff}{ Science} \ct{ffffff}{ (} \ct{ffffff}{AA} \ct{ffffff}{AS} \ct{ffffff}{)} \ct{d4d6ff}{ have} \ct{bec0ff}{ well} \ct{dadbff}{-} \ct{ffffff}{developed} \ct{ffffff}{ and} \ct{f9f9ff}{ successful} \ct{ffffff}{ science} \ct{ffffff}{ policy} \ct{ffffff}{ fellows} \ct{ffffff}{hips} \ct{ffffff}{.} \ct{ffffff}{ These} \ct{ffffff}{ programs} \ct{ffffff}{ acknowledge} \ct{fcfcff}{ that} \ct{ffffff}{ scientists} \ct{ffffff}{ can} \ct{e4e5ff}{ play} \ct{b4b7ff}{ important} \ct{8186ff}{ roles} \ct{c9cbff}{ in} \ct{b8bbff}{ directing} \ct{babcff}{ new} \ct{fefeff}{ laws} \ct{ffffff}{ and} \ct{ffffff}{ policies} \ct{fafaff}{ in} \ct{ffffff}{ their} \ct{ffffff}{ field} \ct{ffffff}{,} \ct{ffffff}{ and} \ct{ffffff}{ that} \ct{ffffff}{ their} \ct{ffffff}{ expertise} \ct{fefeff}{ is} \ct{ffffff}{ needed} \ct{ffffff}{ for} \ct{ebecff}{ effective} \ct{ffffff}{ decision} \ct{e5e6ff}{-} \ct{ffffff}{making} \ct{fefeff}{ \textbackslash{}} \ct{efefff}{[[} \ct{ffffff}{@} \ct{fbfbff}{B} \ct{ffffff}{81} \ct{bdc0ff}{-} \ct{d8daff}{in} \end{minipage}}\\ \textbf{Anthropic Evaluation} \textbf{Score:} 0.2813215497394413\\\textbf{Explanation:}  Phrases related to molecular biology and gene expression, specifically in the context of mRNA transcription and its activation or inhibition. Additionally, there are mentions of certain proteins (IPOTENT, cytokine, IRT), cellular processes (proliferation,\\ \textbf{Top Activations}\\\fbox{\begin{minipage}{\textwidth}\ct{e6e7ff}{b} \ct{caccff}{-} \ct{dedfff}{5} \ct{dbddff}{p} \ct{8085ff}{ overe} \ct{aaadff}{xp} \ct{a0a4ff}{ression} \ct{e4e5ff}{.} \end{minipage}}\\ \fbox{\begin{minipage}{\textwidth}\ct{f3f4ff}{ cases} \ct{c4c6ff}{ showed} \ct{bbbeff}{ EG} \ct{b0b3ff}{FR} \ct{8085ff}{ overe} \ct{a5a8ff}{xp} \ct{a0a4ff}{ression} \ct{e6e7ff}{.} \end{minipage}}\\ \fbox{\begin{minipage}{\textwidth}\ct{fdfdff}{,} \ct{d9dbff}{ TL} \ct{e1e2ff}{R} \ct{d3d5ff}{2} \ct{8085ff}{ overe} \ct{a5a9ff}{xp} \ct{9fa3ff}{ression} \ct{bcbeff}{ in} \end{minipage}}\\ \fbox{\begin{minipage}{\textwidth}\ct{b6b9ff}{ IL} \ct{b4b7ff}{-} \ct{caccff}{1} \ct{a6a9ff}{$\beta$} \ct{8085ff}{ mRNA} \ct{8a8fff}{ expression} \ct{b0b3ff}{ in} \ct{b9bbff}{ the} \end{minipage}}\\ \textbf{Random Activations}\\\fbox{\begin{minipage}{\textwidth}\ct{ffffff}{G} \ct{ffffff}{.} \ct{ffffff}{ al} \ct{ffffff}{bid} \ct{e9eaff}{us} \ct{d6d7ff}{*} \ct{858aff}{ T} \ct{ffffff}{MW} \end{minipage}}\\ \fbox{\begin{minipage}{\textwidth}\ct{bdc0ff}{ had} \ct{ffffff}{ suffered} \ct{ffffff}{ from} \ct{ffffff}{ heat} \ct{ffffff}{ contact} \ct{ffffff}{ ur} \ct{afb2ff}{tic} \ct{ffffff}{aria} \end{minipage}}\\ \fbox{\begin{minipage}{\textwidth}\ct{ffffff}{ } \ct{ffffff}{ "} \ct{ffffff}{al} \ct{ffffff}{ive} \ct{ffffff}{":} \ct{ffffff}{ true} \ct{ffffff}{,} \ct{ffffff}{\textbackslash{}n} \end{minipage}}\\ \fbox{\begin{minipage}{\textwidth}\ct{ffffff}{ =} \ct{ffffff}{ \_} \ct{ffffff}{mm} \ct{ffffff}{\_} \ct{ffffff}{pack} \ct{ffffff}{us} \ct{ffffff}{\_} \ct{ffffff}{ep} \end{minipage}}\\ 

\subsection{GPT-3L, Layer 2, Neuron 289}

\paragraph{OpenAI Evaluation} \textbf{Score:} -0.3236237946813878\\\textbf{Explanation:}  The provided text contains multiple sections, but the activations given for Neuron 4 seem to be related to genetic and statistical data (e.g., population, CI, percent, risk association, and recessive models). Given this, the main thing this neuron does is identify\\ \textbf{Top Activations}\\\fbox{\begin{minipage}{\textwidth}\ct{ffffff}{ in} \ct{ffffff}{ Asian} \ct{ffffff}{ population} \ct{ffffff}{.} \ct{ffffff}{ Similarly} \ct{ffffff}{,} \ct{ffffff}{ in} \ct{ffffff}{ Caucasian} \ct{ffffff}{ population} \ct{ffffff}{,} \ct{ffffff}{ the} \ct{cdcfff}{ rs} \ct{d8daff}{499} \ct{eaebff}{776} \ct{f3f3ff}{ polymorph} \ct{ffffff}{ism} \ct{ffffff}{ attributes} \ct{ffffff}{ risk} \ct{ffffff}{ association} \ct{ffffff}{ in} \ct{ffffff}{ hom} \ct{ffffff}{ozyg} \ct{ffffff}{ote} \ct{ffffff}{ OR} \ct{ccceff}{ 0} \ct{fcfcff}{.} \ct{ffffff}{70} \ct{d5d7ff}{ (} \ct{c2c4ff}{95} \ct{ecedff}{\%} \ct{e4e5ff}{ CI} \ct{d6d7ff}{ [} \ct{7f84ff}{0} \ct{c5c8ff}{.} \ct{b5b7ff}{50} \ct{c5c7ff}{-} \ct{9599ff}{0} \ct{d4d6ff}{.} \ct{e0e1ff}{98} \ct{ffffff}{]),} \ct{ffffff}{ dominant} \ct{ffffff}{ OR} \ct{cdcfff}{ 3} \ct{eeefff}{.} \ct{ffffff}{57} \ct{dadbff}{ (} \ct{c8caff}{95} \ct{edeeff}{\%} \ct{e7e8ff}{ CI} \ct{cdcfff}{ [} \ct{8489ff}{2} \ct{c4c6ff}{.} \ct{b4b7ff}{34} \ct{c3c5ff}{-} \ct{a0a3ff}{5} \ct{d0d2ff}{.} \ct{e4e5ff}{27} \ct{ffffff}{]),} \ct{ffffff}{ and} \ct{f4f4ff}{ recess} \ct{ffffff}{ive} \ct{ffffff}{ models} \ct{ffffff}{ OR} \ct{cbcdff}{ 0} \end{minipage}}\\ \fbox{\begin{minipage}{\textwidth}\ct{e0e1ff}{ *} \ct{f7f7ff}{SE} \ct{f3f4ff}{*} \ct{e8e9ff}{ =} \ct{bec1ff}{ 0} \ct{f8f8ff}{.} \ct{ffffff}{04} \ct{ffffff}{40} \ct{ffffff}{,} \ct{e0e1ff}{ *} \ct{f2f3ff}{t} \ct{f6f6ff}{*} \ct{f0f1ff}{ =} \ct{d1d3ff}{ -} \ct{c2c4ff}{1} \ct{f2f2ff}{.} \ct{eeefff}{07} \ct{ffffff}{75} \ct{ffffff}{,} \ct{e3e4ff}{ *} \ct{eff0ff}{p} \ct{ffffff}{*} \ct{e8e9ff}{ \textbackslash{}} \ct{f2f2ff}{\textgreater{}} \ct{c2c4ff}{ 0} \ct{fafaff}{.} \ct{ffffff}{05} \ct{ffffff}{,} \ct{f2f3ff}{ 95} \ct{ffffff}{\%} \ct{f6f6ff}{ CI} \ct{ccceff}{ (-} \ct{7f84ff}{0} \ct{c4c6ff}{.} \ct{b6b9ff}{13} \ct{ced0ff}{38} \ct{e3e4ff}{,} \ct{8d91ff}{ 0} \ct{e1e2ff}{.} \ct{c1c4ff}{0} \ct{e5e6ff}{390} \ct{ffffff}{)} \ct{ffffff}{ for} \ct{ffffff}{ the} \ct{ffffff}{ Slov} \ct{ffffff}{ak} \ct{ffffff}{ian} \ct{ffffff}{ villagers} \ct{ffffff}{ story} \ct{fcfcff}{\textbackslash{}} \ct{ffffff}{],} \ct{ffffff}{ therefore} \ct{ffffff}{ indicating} \ct{ffffff}{ full} \ct{ffffff}{ mediation} \ct{ffffff}{ by} \ct{ffffff}{ exoner} \ct{ffffff}{ations} \ct{ffffff}{ and} \ct{ffffff}{ out} \ct{ffffff}{-} \ct{ffffff}{group} \ct{ffffff}{ focused} \ct{ffffff}{ emotions} \end{minipage}}\\ \textbf{Random Activations}\\\fbox{\begin{minipage}{\textwidth}\ct{ffffff}{long} \ct{ffffff}{er} \ct{ffffff}{ the} \ct{ffffff}{ vortex} \ct{ffffff}{,} \ct{ffffff}{ it} \ct{ffffff}{'s} \ct{ffffff}{ the} \ct{ffffff}{ smooth} \ct{ffffff}{ current} \ct{ffffff}{ of} \ct{ffffff}{ rotating} \ct{ffffff}{ air} \ct{ffffff}{ which} \ct{ffffff}{ is} \ct{ffffff}{ next} \ct{ffffff}{ to} \ct{d6d8ff}{\textbackslash{}n} \ct{ffffff}{the} \ct{ffffff}{ vortex} \ct{ffffff}{,} \ct{ffffff}{ and} \ct{ffffff}{ we} \ct{ffffff}{ use} \ct{ffffff}{ the} \ct{ffffff}{ upd} \ct{ffffff}{raft} \ct{ffffff}{ of} \ct{ffffff}{ this} \ct{ffffff}{ air} \ct{ffffff}{."} \ct{ffffff}{ Taking} \ct{ffffff}{ advantage} \ct{ffffff}{ of} \ct{ffffff}{ the} \ct{ffffff}{ free} \ct{7f84ff}{\textbackslash{}n} \ct{ffffff}{lift} \ct{ffffff}{ in} \ct{ffffff}{ this} \ct{ffffff}{ upd} \ct{ffffff}{raft} \ct{ffffff}{ of} \ct{ffffff}{ air} \ct{ffffff}{ is} \ct{ffffff}{ called} \ct{ededff}{ "} \ct{ffffff}{wake} \ct{ffffff}{-} \ct{ffffff}{energy} \ct{ffffff}{ retrieval} \ct{ffffff}{."} \ct{ffffff}{ ...} \ct{ffffff}{ on} \ct{ffffff}{ long} \ct{ffffff}{-} \ct{a4a7ff}{\textbackslash{}n} \ct{ffffff}{haul} \ct{ffffff}{ flights} \ct{ffffff}{,} \ct{ffffff}{ fuel} \ct{ffffff}{ savings} \ct{ffffff}{ of} \ct{ffffff}{ between} \end{minipage}}\\ \fbox{\begin{minipage}{\textwidth}\ct{ffffff}{ushing} \ct{ffffff}{.} \ct{ffffff}{\textbackslash{}n\textbackslash{}n} \ct{ffffff}{ } \ct{ffffff}{ } \ct{ffffff}{ } \ct{ffffff}{ } \ct{ffffff}{ } \ct{ffffff}{ } \ct{7f84ff}{ S} \ct{ffffff}{igh} \ct{ffffff}{ called} \ct{ffffff}{ her} \ct{ffffff}{ supervisor} \ct{ffffff}{.} \ct{fbfbff}{ Sergeant} \ct{ffffff}{ Sweeney} \ct{ffffff}{ and} \ct{ffffff}{ Deputy} \ct{ffffff}{ Ray} \ct{ffffff}{ responded} \ct{ffffff}{\textbackslash{}n} \ct{ffffff}{\textbackslash{}n} \ct{ffffff}{and} \ct{ffffff}{ moved} \ct{ffffff}{ Don} \ct{ffffff}{ery} \ct{ffffff}{ so} \ct{ffffff}{ that} \ct{999dff}{ S} \ct{ffffff}{igh} \ct{ffffff}{ could} \ct{ffffff}{ search} \ct{ffffff}{ his} \ct{ffffff}{ cell} \ct{ffffff}{.} \ct{ffffff}{ Don} \ct{ffffff}{ery} \ct{ffffff}{ had} \ct{ffffff}{ been} \ct{ffffff}{ in} \ct{ffffff}{ his} \ct{ffffff}{ new} \ct{bcbfff}{\textbackslash{}n} \ct{f3f3ff}{\textbackslash{}n} \ct{ffffff}{cell} \ct{ffffff}{ for} \ct{ffffff}{ less} \ct{ffffff}{ than} \ct{ffffff}{ five} \ct{ffffff}{ minutes} \ct{ffffff}{ when} \ct{ffffff}{ the} \ct{ffffff}{ toilet} \ct{ffffff}{ overfl} \ct{ffffff}{owed} \ct{ffffff}{ and} \ct{ffffff}{ water} \ct{ffffff}{ began} \ct{ffffff}{ flowing} \ct{ffffff}{ out} \ct{ffffff}{\textbackslash{}n} \ct{ffffff}{\textbackslash{}n} \ct{ffffff}{of} \end{minipage}}\\ \paragraph{Anthropic Evaluation} \textbf{Score:} 0.06532542667915378\\\textbf{Explanation:}  strings containing specific numbers and alphanumeric characters, such as "CI-50-.", "e-44-", "87\"-", and "f\"-". Additionally, it activates slightly for certain words like "cost", "weeks", "disability", and\\ \textbf{Top Activations}\\\fbox{\begin{minipage}{\textwidth}\ct{c2c4ff}{95} \ct{ecedff}{\%} \ct{e4e5ff}{ CI} \ct{d6d7ff}{ [} \ct{7f84ff}{0} \ct{c5c8ff}{.} \ct{b5b7ff}{50} \ct{c5c7ff}{-} \end{minipage}}\\ \fbox{\begin{minipage}{\textwidth}\ct{c3c5ff}{95} \ct{e7e8ff}{\%} \ct{dcdeff}{ CI} \ct{caccff}{,} \ct{7f84ff}{ 0} \ct{d4d6ff}{.} \ct{bec0ff}{13} \ct{c1c3ff}{-} \end{minipage}}\\ \fbox{\begin{minipage}{\textwidth}\ct{e5e6ff}{f} \ct{eeefff}{\textbackslash{}"} \ct{eaebff}{],} \ct{d4d6ff}{ [} \ct{7f84ff}{0} \ct{c4c7ff}{.} \ct{babdff}{22} \ct{c4c6ff}{22} \end{minipage}}\\ \fbox{\begin{minipage}{\textwidth}\ct{c9cbff}{95} \ct{eff0ff}{\%} \ct{eaeaff}{ CI} \ct{ccceff}{ [} \ct{7f84ff}{0} \ct{c4c6ff}{.} \ct{bbbeff}{50} \ct{c3c6ff}{-} \end{minipage}}\\ \textbf{Random Activations}\\\fbox{\begin{minipage}{\textwidth}\ct{a4a8ff}{\{} \ct{f9f9ff}{8} \ct{7f84ff}{\}\{} \ct{ffffff}{45} \ct{fbfbff}{\}\textbackslash{}} \ct{ffffff}{pi} \ct{ffffff}{\textbackslash{}\textbackslash{}} \ct{f1f2ff}{\textbackslash{}n} \end{minipage}}\\ \fbox{\begin{minipage}{\textwidth}\ct{ffffff}{ instead} \ct{ffffff}{ of} \ct{ffffff}{ that} \ct{ffffff}{ silly} \ct{ffffff}{ website} \ct{ffffff}{.} \ct{7f84ff}{\textless{}|endoftext|\textgreater{}} \ct{e2e3ff}{How} \end{minipage}}\\ \fbox{\begin{minipage}{\textwidth}\ct{ffffff}{ is} \ct{ffffff}{ free} \ct{ffffff}{ software} \ct{ffffff}{;} \ct{8085ff}{ you} \ct{b0b3ff}{ can} \ct{ffffff}{ redist} \ct{ffffff}{ribute} \end{minipage}}\\

\subsection{GPT-3L-SAE-16x, Layer 2, Neuron 57}

\paragraph{OpenAI Evaluation} \textbf{Score:} 0.1427145260798203\\\textbf{Explanation:}  months or the word "Bank" followed by a year.\\ \textbf{Top Activations}\\\fbox{\begin{minipage}{\textwidth}\ct{ffffff}{ } \ct{ffffff}{ } \ct{ffffff}{ } \ct{ffffff}{ } \ct{ffffff}{ No} \ct{ffffff}{.} \ct{ffffff}{ 18} \ct{ffffff}{-} \ct{ffffff}{20} \ct{ffffff}{609} \ct{ffffff}{ } \ct{ffffff}{ } \ct{ffffff}{ } \ct{ffffff}{ } \ct{ffffff}{ } \ct{ffffff}{ } \ct{ffffff}{ } \ct{ffffff}{ } \ct{ffffff}{ } \ct{ffffff}{ } \ct{ffffff}{ } \ct{ffffff}{ } \ct{ffffff}{ } \ct{ffffff}{ } \ct{ffffff}{ } \ct{ffffff}{ } \ct{ffffff}{ } \ct{ffffff}{ } \ct{ffffff}{ } \ct{ffffff}{ } \ct{ffffff}{ } \ct{ffffff}{ } \ct{7f84ff}{ February} \ct{edeeff}{ 21} \ct{ffffff}{,} \ct{ffffff}{ 2020} \ct{ffffff}{\textbackslash{}n} \ct{ffffff}{ } \ct{ffffff}{ } \ct{ffffff}{ } \ct{ffffff}{ } \ct{ffffff}{ } \ct{ffffff}{ } \ct{ffffff}{ } \ct{ffffff}{ } \ct{ffffff}{ } \ct{ffffff}{ } \ct{ffffff}{ } \ct{ffffff}{ } \ct{ffffff}{ } \ct{ffffff}{ } \ct{ffffff}{ } \ct{ffffff}{ } \ct{ffffff}{ } \ct{ffffff}{ } \ct{ffffff}{ } \ct{ffffff}{ } \ct{ffffff}{ } \ct{ffffff}{ } \ct{ffffff}{ } \ct{ffffff}{ } \ct{ffffff}{ } \ct{ffffff}{ } \ct{ffffff}{ } \end{minipage}}\\ \fbox{\begin{minipage}{\textwidth}\ct{ffffff}{ with} \ct{ffffff}{ the} \ct{ffffff}{ compact} \ct{ffffff}{-} \ct{ffffff}{open} \ct{ffffff}{ top} \ct{ffffff}{ology} \ct{ffffff}{,} \ct{ffffff}{ is} \ct{ffffff}{ a} \ct{ffffff}{ locally} \ct{ffffff}{ compact} \ct{ffffff}{ group} \ct{ffffff}{.'} \ct{ffffff}{\textbackslash{}n} \ct{ffffff}{author} \ct{ffffff}{:} \ct{ffffff}{\textbackslash{}n} \ct{ffffff}{-} \ct{ffffff}{ '} \ct{ffffff}{Nic} \ct{ffffff}{olas} \ct{ffffff}{ Rad} \ct{ffffff}{u} \ct{ffffff}{[} \ct{ffffff}{\^{}} \ct{ffffff}{1} \ct{ffffff}{]'} \ct{ffffff}{\textbackslash{}n} \ct{f8f8ff}{date} \ct{ffffff}{:} \ct{ffffff}{ '} \ct{7f84ff}{July} \ct{ffffff}{ 15} \ct{ffffff}{,} \ct{ffffff}{ 2016} \ct{ffffff}{'} \ct{ffffff}{\textbackslash{}n} \ct{ffffff}{title} \ct{ffffff}{:} \ct{ffffff}{ |} \ct{ffffff}{\textbackslash{}n} \ct{ffffff}{ } \ct{ffffff}{ } \ct{ffffff}{ } \ct{ffffff}{ A} \ct{ffffff}{ top} \ct{ffffff}{ological} \ct{ffffff}{ characterization} \ct{ffffff}{ of} \ct{ffffff}{ the} \ct{ffffff}{ M} \ct{ffffff}{ouf} \ct{ffffff}{ang} \ct{ffffff}{\textbackslash{}} \ct{ffffff}{\textbackslash{}n} \ct{ffffff}{ } \ct{ffffff}{ } \ct{ffffff}{ } \ct{ffffff}{ property} \ct{ffffff}{ for} \ct{ffffff}{ compact} \ct{ffffff}{ polyg} \ct{ffffff}{ons} \end{minipage}}\\ \textbf{Random Activations}\\\fbox{\begin{minipage}{\textwidth}\ct{ffffff}{ of} \ct{ffffff}{ DN} \ct{ffffff}{ *} \ct{ffffff}{db} \ct{ffffff}{/} \ct{ffffff}{db} \ct{ffffff}{*} \ct{ffffff}{ mice} \ct{ffffff}{.} \ct{ffffff}{\textbackslash{}} \ct{ffffff}{\textbackslash{}n} \ct{ffffff}{(} \ct{ffffff}{**} \ct{ffffff}{A} \ct{ffffff}{**} \ct{ffffff}{)} \ct{ffffff}{ Ur} \ct{ffffff}{inary} \ct{ffffff}{ album} \ct{ffffff}{in} \ct{ffffff}{ to} \ct{ffffff}{ creat} \ct{ffffff}{in} \ct{ffffff}{ine} \ct{ffffff}{ ratio} \ct{ffffff}{.} \ct{ffffff}{ (} \ct{ffffff}{**} \ct{ffffff}{B} \ct{ffffff}{**} \ct{ffffff}{)} \ct{ffffff}{ Ser} \ct{ffffff}{um} \ct{ffffff}{ u} \ct{ffffff}{rea} \ct{ffffff}{ nitrogen} \ct{ffffff}{.} \ct{ffffff}{ (} \ct{ffffff}{**} \ct{ffffff}{C} \ct{ffffff}{**} \ct{ffffff}{)} \ct{ffffff}{ Left} \ct{ffffff}{ kidney} \ct{ffffff}{ weight} \ct{ffffff}{ to} \ct{ffffff}{ body} \ct{ffffff}{ weight} \ct{ffffff}{ ratio} \ct{ffffff}{.} \ct{ffffff}{ (} \ct{ffffff}{**} \ct{ffffff}{D} \ct{ffffff}{**} \ct{ffffff}{)} \ct{ffffff}{ HE} \ct{ffffff}{ st} \ct{ffffff}{aining} \ct{ffffff}{.} \ct{ffffff}{ Bar} \ct{ffffff}{ } \ct{ffffff}{ } \ct{ffffff}{=} \ct{ffffff}{ } \end{minipage}}\\ \fbox{\begin{minipage}{\textwidth}\ct{ffffff}{ this} \ct{ffffff}{ email} \ct{ffffff}{:} \ct{ffffff}{ ot} \ct{ffffff}{isd} \ct{ffffff}{ark} \ct{ffffff}{o} \ct{ffffff}{60} \ct{ffffff}{@} \ct{ffffff}{yahoo} \ct{ffffff}{.} \ct{ffffff}{com} \ct{ffffff}{\textbackslash{}n} \ct{ffffff}{\textbackslash{}n} \ct{ffffff}{HE} \ct{ffffff}{ FIX} \ct{ffffff}{ THE} \ct{ffffff}{ FO} \ct{ffffff}{LLOW} \ct{ffffff}{ING} \ct{ffffff}{ PR} \ct{ffffff}{OB} \ct{ffffff}{LE} \ct{ffffff}{MS} \ct{ffffff}{ TO} \ct{ffffff}{ ALL} \ct{ffffff}{\textbackslash{}n} \ct{ffffff}{\textbackslash{}n} \ct{ffffff}{AC} \ct{ffffff}{R} \ct{ffffff}{OSS} \ct{ffffff}{ THE} \ct{ffffff}{ GL} \ct{ffffff}{OB} \ct{ffffff}{E} \ct{ffffff}{ ON} \ct{ffffff}{:} \ct{ffffff}{\textbackslash{}n} \ct{ffffff}{\textbackslash{}n} \ct{ffffff}{1} \ct{ffffff}{.} \ct{ffffff}{ Getting} \ct{ffffff}{ your} \ct{ffffff}{ lover} \ct{ffffff}{ or} \ct{ffffff}{ husband} \ct{ffffff}{ back} \ct{ffffff}{\textbackslash{}n} \ct{ffffff}{\textbackslash{}n} \ct{ffffff}{2} \ct{ffffff}{.} \ct{ffffff}{ Spiritual} \ct{ffffff}{ bullet} \ct{ffffff}{proof} \ct{ffffff}{\textbackslash{}n} \ct{ffffff}{\textbackslash{}n} \ct{ffffff}{3} \ct{ffffff}{.} \ct{ffffff}{ Training} \ct{ffffff}{\textbackslash{}n} \ct{ffffff}{\textbackslash{}n} \ct{ffffff}{4} \ct{ffffff}{.} \ct{ffffff}{ Money} \end{minipage}}\\ \paragraph{Anthropic Evaluation} \textbf{Score:} 0.2540425061001668\\\textbf{Explanation:}  dates and specifically, the month and day for a given year. The neuron is not activated by the year alone, and it requires both the month and day for a complete activation.\\ \textbf{Top Activations}\\\fbox{\begin{minipage}{\textwidth}\ct{ffffff}{ } \ct{ffffff}{ } \ct{ffffff}{ } \ct{ffffff}{ } \ct{7f84ff}{ February} \ct{edeeff}{ 21} \ct{ffffff}{,} \ct{ffffff}{ 2020} \end{minipage}}\\ \fbox{\begin{minipage}{\textwidth}\ct{ffffff}{\textbackslash{}n} \ct{f8f8ff}{date} \ct{ffffff}{:} \ct{ffffff}{ '} \ct{7f84ff}{July} \ct{ffffff}{ 15} \ct{ffffff}{,} \ct{ffffff}{ 2016} \end{minipage}}\\ \fbox{\begin{minipage}{\textwidth}\ct{ffffff}{field} \ct{ffffff}{,} \ct{ffffff}{ Missouri} \ct{ffffff}{ (} \ct{7f84ff}{December} \ct{ffffff}{ 15} \ct{ffffff}{,} \ct{ffffff}{ 2014} \end{minipage}}\\ \fbox{\begin{minipage}{\textwidth}\ct{ffffff}{ } \ct{ffffff}{ } \ct{ffffff}{ } \ct{ffffff}{ } \ct{7f84ff}{ February} \ct{ecedff}{ 5} \ct{ffffff}{,} \ct{ffffff}{ 1998} \end{minipage}}\\ \textbf{Random Activations}\\\fbox{\begin{minipage}{\textwidth}\ct{ffffff}{\textbackslash{}n} \ct{ffffff}{ } \ct{ffffff}{ Can} \ct{ffffff}{ola} \ct{ffffff}{ oil} \ct{ffffff}{ } \ct{ffffff}{ } \ct{ffffff}{ } \end{minipage}}\\ \fbox{\begin{minipage}{\textwidth}\ct{ffffff}{\}} \ct{ffffff}{
} \ct{ffffff}{\textbackslash{}n} \ct{ffffff}{
} \ct{ffffff}{\textbackslash{}n} \ct{ffffff}{.} \ct{ffffff}{c} \ct{ffffff}{ke} \end{minipage}}\\ \fbox{\begin{minipage}{\textwidth}\ct{ffffff}{uana} \ct{ffffff}{ when} \ct{ffffff}{ a} \ct{ffffff}{ draw} \ct{ffffff}{ would} \ct{ffffff}{ have} \ct{ffffff}{ clin} \ct{ffffff}{ched} \end{minipage}}\\ \fbox{\begin{minipage}{\textwidth}\ct{ffffff}{ } \ct{ffffff}{ } \ct{ffffff}{.} \ct{ffffff}{\textless{}/} \ct{ffffff}{p} \ct{ffffff}{\textgreater{}} \ct{ffffff}{\textbackslash{}n} \ct{ffffff}{\textbackslash{}t} \end{minipage}}\\ 

\end{document}